\PassOptionsToPackage{table}{xcolor}
\documentclass{fairmeta}
\pdfinfoomitdate=1 \pdftrailerid{} \pdfsuppressptexinfo=-1 \hypersetup{pdfcreator={},pdfproducer={}}%

\usepackage{adjustbox}
\usepackage{amsmath,amssymb}
\usepackage{array}
\usepackage{colortbl}
\usepackage{csquotes}
\usepackage{float}
\usepackage{fontawesome}
\usepackage{multirow}
\usepackage{longtable}
\usepackage{needspace}
\usepackage{paralist}
\usepackage{pifont}
\usepackage{tabularx}
\usepackage{threeparttable}
\usepackage{tikz}
\usepackage{wasysym}
\usepackage{wrapfig}
\usepackage{bm}

\let\surveyLightbulbIcon\faLightbulbO
\let\surveyExchangeIcon\faExchange
\let\surveyQuestionIcon\faQuestionCircle
\DeclareRobustCommand{\insighticon}{\textmd{\surveyLightbulbIcon}}
\DeclareRobustCommand{\differenceicon}{\textmd{\surveyExchangeIcon}}
\DeclareRobustCommand{\frontiericon}{\textmd{\surveyQuestionIcon}}

\usepackage[backend=biber,style=numeric-comp,natbib=true,sorting=nyt,
            giveninits=false,maxbibnames=99,minbibnames=99,
            url=true,doi=false,isbn=false,eprint=true]{biblatex}
\AtEveryBibitem{%
  \clearfield{eprintclass}%
  \iffieldundef{eprint}{}{\clearfield{url}}%
  \iffieldundef{doi}{}{\clearfield{url}}%
  \iffieldundef{booktitle}{}{\clearfield{url}}%
  \iffieldundef{journaltitle}{}{\clearfield{url}}%
}
\AtBeginBibliography{\ifPDFTeX\setlength{\emergencystretch}{1em}\fi}

\graphicspath{{figures/}}
\setdefaultleftmargin{1.2em}{}{}{}{}{}

\newcommand{\cmark}{\ding{51}}
\newcommand{\xmark}{\ding{55}}
\newcommand{\pmark}{\LEFTcircle}
\newcommand{\nna}{--}
\newcommand{\correspond}{\spadesuit}
\newcolumntype{Y}{>{\raggedright\arraybackslash}X}

\definecolor{surveyInk}{RGB}{24,35,42}
\definecolor{surveyNeutral}{RGB}{65,83,99}
\definecolor{surveyRead}{HTML}{FFBF00}
\definecolor{surveyWrite}{HTML}{06B0F0}
\definecolor{surveyShare}{HTML}{82F5B1}
\definecolor{surveyInteract}{HTML}{CD93E3}
\definecolor{surveyReadInk}{HTML}{C08000}
\definecolor{surveyWriteInk}{HTML}{0097CF}
\definecolor{surveyShareInk}{HTML}{00A344}
\definecolor{surveyInteractInk}{HTML}{C363E9}

\newif\ifopicons\opiconstrue

\DeclareRobustCommand{\opsep}{\ifopicons\ \fi}
\DeclareRobustCommand{\readicon}{\opbadge{surveyReadInk}{\faEye}}
\DeclareRobustCommand{\writeicon}{\opbadge{surveyWriteInk}{\faPencil}}
\DeclareRobustCommand{\shareicon}{\opbadge{surveyShareInk}{\faExchange}}
\DeclareRobustCommand{\interacticon}{\opbadge{surveyInteractInk}{\faGamepad}}

\DeclareRobustCommand{\readop}[1]{\opkeyword{\readicon}{surveyReadInk}{#1}}
\DeclareRobustCommand{\writeop}[1]{\opkeyword{\writeicon}{surveyWriteInk}{#1}}
\DeclareRobustCommand{\shareop}[1]{\opkeyword{\shareicon}{surveyShareInk}{#1}}
\DeclareRobustCommand{\interactop}[1]{\opkeyword{\interacticon}{surveyInteractInk}{#1}}
\definecolor{surveyLink}{RGB}{23,104,127}
\definecolor{surveyMuted}{RGB}{91,103,109}
\definecolor{surveyLeadText}{RGB}{51,65,71}
\definecolor{surveyCaptionText}{RGB}{55,66,71}
\colorlet{surveyFuture}{surveyNeutral}
\colorlet{surveyTableGap}{surveyNeutral!7}
\colorlet{surveyCurrentAccent}{surveyNeutral}
\colorlet{surveyCurrentSoft}{surveyNeutral!12}
\colorlet{surveyCurrentFaint}{surveyNeutral!5}
\colorlet{surveyCurrentStripe}{surveyCurrentFaint}
\colorlet{surveyCurrentHead}{surveyCurrentSoft}
\colorlet{surveyCurrentGroupBack}{surveyCurrentSoft}
\colorlet{surveyCurrentBadgeBack}{surveyCurrentSoft}
\colorlet{surveyCurrentChapterBack}{surveyCurrentFaint}
\colorlet{surveyCurrentChapterTop}{surveyNeutral}
\colorlet{surveyCurrentTableText}{surveyInk}
\colorlet{surveyCurrentGroupText}{surveyInk}
\colorlet{surveyCurrentHeadingText}{surveyInk}
\colorlet{surveyCurrentLeadLabel}{surveyInk}
\colorlet{surveyBoxAccent}{surveyNeutral}
\colorlet{surveyBoxBack}{surveyNeutral!5}
\colorlet{surveyBoxTitleText}{white}
\newcommand{\surveytheme}[1]{%
  \colorlet{surveyCurrentAccent}{#1}%
  \colorlet{surveyCurrentSoft}{#1!12}%
  \colorlet{surveyCurrentFaint}{#1!5}%
  \colorlet{surveyCurrentStripe}{surveyCurrentFaint}%
  \colorlet{surveyCurrentHead}{surveyCurrentSoft}%
  \colorlet{surveyCurrentGroupBack}{surveyCurrentSoft}%
  \colorlet{surveyCurrentBadgeBack}{surveyCurrentSoft}%
  \colorlet{surveyCurrentChapterBack}{surveyCurrentFaint}%
  \colorlet{surveyCurrentChapterTop}{#1}%
  \colorlet{surveyCurrentTableText}{surveyInk}%
  \colorlet{surveyCurrentGroupText}{surveyInk}%
  \colorlet{surveyCurrentHeadingText}{surveyInk}%
  \colorlet{surveyCurrentLeadLabel}{surveyInk}%
  \colorlet{surveyBoxAccent}{#1}%
  \colorlet{surveyBoxBack}{#1!4}%
  \colorlet{surveyBoxTitleText}{white}}

\newcommand{\surveytablecolors}{\rowcolors{2}{white}{surveyCurrentStripe}}
\newcommand{\surveytablehead}{\rowcolor{surveyCurrentHead}}
\newcommand{\gaphl}{\rowcolor{surveyTableGap}}
\newcommand{\tableheadcell}[1]{%
  \textcolor{surveyCurrentTableText}{\sffamily\bfseries #1}}
\newcommand{\surveygroup}[2]{%
  \rowcolor{surveyCurrentGroupBack}%
  \multicolumn{#1}{@{}l}{%
    \strut\textcolor{surveyCurrentGroupText}{\sffamily\bfseries #2}}\\[-0.25ex]}
\newcommand{\surveylonggroup}[2]{%
  \rowcolor{surveyCurrentGroupBack}%
  \multicolumn{#1}{@{}l}{%
    \strut\textcolor{surveyCurrentGroupText}{\sffamily\bfseries #2}}\\*[-0.25ex]}
\newlength{\surveytableunit}
\newcommand{\surveybadge}[1]{%
  \begingroup\setlength{\fboxsep}{2.2pt}%
  \colorbox{surveyCurrentBadgeBack}{%
    \textcolor{surveyCurrentGroupText}{\scriptsize\sffamily\bfseries\strut #1}}%
  \endgroup}
\newcommand{\surveybadgecolor}[2]{%
  \begingroup\setlength{\fboxsep}{2.2pt}%
  \colorbox{#1!12}{\textcolor{surveyInk}{\scriptsize\sffamily\bfseries\strut #2}}%
  \endgroup}
\DeclareRobustCommand{\tableyes}{\textcolor{surveyLink}{\cmark}}
\DeclareRobustCommand{\tableno}{\textcolor{surveyMuted}{\nna}}
\DeclareRobustCommand{\tableoptional}{\textcolor{surveyLink}{\ensuremath{\circ}}}
\DeclareRobustCommand{\tablepartial}{\textcolor{surveyNeutral}{\textmd{\pmark}}}
\DeclareRobustCommand{\trainedmark}{\tableyes}
\DeclareRobustCommand{\frozenmark}{\textcolor{surveyMuted}{\xmark}}
\DeclareRobustCommand{\spacecode}[1]{%
  \textcolor{surveyLeadText}{\sffamily\bfseries #1}}
\newcommand{\trainingcell}[2]{%
  \begingroup\hangindent=1.25em\hangafter=1%
  \noindent\makebox[1.25em][l]{#1}#2\par\endgroup}
\newcommand{\modeslot}[1]{\makebox[.40cm][c]{#1}}
\newcommand{\modeflag}[1]{%
  \modeslot{\ifnum#1=1\textcolor{surveyLink}{\textbullet}\else\tableno\fi}}
\newcommand{\generationmodes}[5]{%
  \mbox{\modeflag{#1}\modeflag{#2}\modeflag{#3}\modeflag{#4}\modeflag{#5}}}
\newcommand{\generationmodehead}{%
  \mbox{\scriptsize\sffamily\bfseries
    \modeslot{T}\modeslot{I}\modeslot{V}\modeslot{AV}\modeslot{L}}}
\newcommand{\videocaps}[3]{%
  \mbox{\modeslot{#1}\modeslot{#2}\modeslot{#3}}}
\newcommand{\videocaphead}{%
  \videocaps{\tableheadcell{U}}{\tableheadcell{G}}{\tableheadcell{E}}}
\newcommand{\generationtablekey}{%
  \par\vspace{3pt}\noindent\begin{minipage}{\textwidth}\footnotesize\raggedright
  \textbf{Space:} \spacecode{Px} pixels; \spacecode{VAE} VAE latents;
  \spacecode{Tok} discrete tokens; \spacecode{BD} backend-dependent;
  \spacecode{NR} not reported.\par
  \textbf{Online:} \tableyes{} active MLLM; \tableoptional{} optional MLLM;
  \tableno{} no online MLLM; \spacecode{NR} not reported. LLM only: text-only backbone.\par
  \textbf{Modes:} \textcolor{surveyLink}{\textbullet} listed; \tableno{} not listed.
  T text-to-video; I image-to-video; V video-conditioned generation/editing;
  AV joint audio-video output; L streaming or minute-scale output. AR autoregressive.
  \end{minipage}}
\newcommand{\unifiedtablekey}{%
  \par\vspace{3pt}\noindent\begin{minipage}{\textwidth}\footnotesize\raggedright
  \textbf{Video tasks:} U understanding; G generation; E editing.
  \tableyes{} demonstrated; \tablepartial{} inherited/task-specific evidence;
  \tableno{} not established.\par
  \textbf{Training:} AR autoregressive; FM flow matching; NR not reported.
  \textbf{Predictor attention:} C causal; B bidirectional; H hybrid.
  \end{minipage}}

\titleformat{\section}[hang]
  {\Large\sffamily\bfseries\color{surveyInk}}
  {\textcolor{surveyCurrentAccent}{\thesection}}{0.72em}{}
\titleformat{\subsection}[hang]
  {\large\sffamily\bfseries\color{surveyInk}}
  {\textcolor{surveyCurrentAccent}{\thesubsection}}{0.68em}{}
\titleformat{\subsubsection}[hang]
  {\normalsize\sffamily\bfseries\color{surveyInk}}
  {\textcolor{surveyCurrentAccent}{\thesubsubsection}}{0.62em}{}
\titleformat*{\paragraph}{\normalsize\sffamily\bfseries\color{surveyCurrentHeadingText}}
\hypersetup{
  linkcolor=surveyLink,
  citecolor=surveyLink,
  urlcolor=surveyLink
}
\DeclareCaptionFormat{surveycustom}{%
  {\sffamily\bfseries\color{surveyCurrentGroupText}#1#2}\enspace
  {\color{surveyCaptionText}#3}}

\tcbset{
  survey insight common/.style={
    enhanced,
    frame hidden,
    borderline north={0.55pt}{0pt}{surveyBoxAccent!42},
    borderline south={0.55pt}{0pt}{surveyBoxAccent!20},
    colback=surveyBoxBack,
    arc=2pt,
    left=10pt,
    right=10pt,
    top=8pt,
    bottom=7pt,
    boxsep=0pt,
    fonttitle=\small\sffamily\bfseries,
    fontupper=\small,
    coltitle=surveyInk,
    attach boxed title to top left={xshift=8pt,yshift=-2.4mm},
    varwidth boxed title=0.90\linewidth,
    boxed title style={
      colback=surveyBoxAccent!13,
      colframe=surveyBoxAccent!13,
      boxrule=0pt,
      arc=2pt,
      left=7pt,
      right=8pt,
      top=3pt,
      bottom=3pt
    },
    before={\par\Needspace{0.16\textheight}},
    after={\par\medskip}
  }
}
\newcommand{\surveyinsighttitle}[2]{%
  \textcolor{surveyBoxAccent!72!black}{#1}\enspace
  \textcolor{surveyBoxAccent!72!black}{\scriptsize\sffamily\bfseries INSIGHT}\enspace
  #2}
\newtcolorbox{insightbox}[1]{
  survey insight common,
  colback=surveyBoxBack,
  colframe=surveyBoxAccent,
  colbacktitle=surveyBoxAccent!13,
  title={\surveyinsighttitle{\insighticon}{#1}}
}
\newtcolorbox{differencebox}[1]{
  survey insight common,
  colback=surveyBoxBack,
  colframe=surveyBoxAccent,
  colbacktitle=surveyBoxAccent!13,
  title={\surveyinsighttitle{\differenceicon}{#1}}
}
\newtcolorbox{frontierbox}[1]{
  survey insight common,
  colback=surveyBoxBack,
  colframe=surveyBoxAccent,
  colbacktitle=surveyBoxAccent!13,
  title={\surveyinsighttitle{\frontiericon}{#1}}
}

\tcbuselibrary{theorems}
\tcbset{
  survey formal/.style={
    enhanced,
    frame hidden,
    colback=surveyNeutral!7,
    arc=2pt,
    left=9pt,
    right=9pt,
    top=6pt,
    bottom=6pt,
    before skip=9pt,
    after skip=9pt,
    detach title,
    before upper={\tcbtitle\enspace},
    fonttitle=\sffamily\bfseries\small,
    coltitle=surveyInk,
    description font=\mdseries,
    separator sign none,
    description delimiters parenthesis
  }
}
\newtcbtheorem[number within=section]{definition}{Definition}{survey formal}{def}
\newtcbtheorem[use counter from=definition]{observation}{Observation}%
  {survey formal,fontupper=\itshape}{obs}

\title{Reading, Writing, Sharing, and Interacting\\ with the World through Video}
\subtitle{A Survey Toward Unif\kern0pt ied Video World Models}%
\project{\url{https://vid-gen.github.io/video-world-model-survey}}
\github{\url{https://github.com/vid-gen/awesome-video-world-model}}

\author[1,*]{Shengqiong Wu}
\author[2,*]{Meng Luo}
\author[2]{Bobo Li}
\author[3]{Ziwei Liu}
\author[2]{Shuicheng Yan}
\author[4]{Ming-Hsuan Yang}
\author[1]{Philip Torr}
\author[1,\correspond]{Hao Fei}

\affiliation[1]{University of Oxford}
\affiliation[2]{NUS}
\affiliation[3]{NTU}
\affiliation[4]{University of California, Merced}
\contribution[*]{Equal contribution}\contribution[\correspond]{Project lead, correspondence (haofei7419@gmail.com)}

\abstract{
Video foundation models have advanced rapidly in understanding and generating visual sequences, while video world models seek to capture the underlying dynamics of the world they depict. 
Despite their different objectives, both concern the relationship between video observations and an evolving world. 
This survey adopts world modeling as a unifying lens for examining these developments. 
Viewing an observed video as a partial, lossy, and temporally ordered projection of a latent temporal world, we organize the literature around four complementary operations: \readop{reading} infers world properties from observed video; \writeop{writing} synthesizes video from specified or learned world conditions; \shareop{sharing} integrates understanding and generation through common modeling mechanisms; and \interactop{interacting} predicts and updates world states under actions and evolving observations. 
This perspective distinguishes temporal coherence in observed or generated sequences from consistency in the underlying world state. 
Across tasks, representations, models, datasets, and benchmarks, we trace a shift in emphasis from clip-level recognition and perceptual fidelity toward persistent state, controllable evolution, physical and causal validity, and evaluation grounded in interaction and downstream decisions, and we distill the open challenges that follow. 
}

\begin{document}
\definecolor{coverSky}{RGB}{233,246,254}
\newlength{\coverfigw}
\setlength{\coverfigw}{\dimexpr\textwidth*1354/1330\relax}
\AddToHookNext{shipout/background}{\put(0,0){%
  \begin{tikzpicture}[remember picture,overlay]
    \pgfmathsetmacro{\coverscale}{\coverfigw/\textwidth*1330/1354}
    \begin{scope}
      \clip ([xshift={-0.5*1330/1354*\coverfigw},yshift=-0.8cm]current page.north) rectangle ([xshift={0.5*1330/1354*\coverfigw},yshift=2.5cm]current page.south);
      \fill[coverSky] (current page.north west) rectangle (current page.south east);
      \node[anchor=north,inner sep=0pt,outer sep=0pt,
            yshift={1081/1054*\coverscale*\paperwidth-160.25/1354*\coverfigw}] (coverbg)
        at (current page.north |- teaserfig.north)
        {\includegraphics[width=\coverscale\paperwidth]{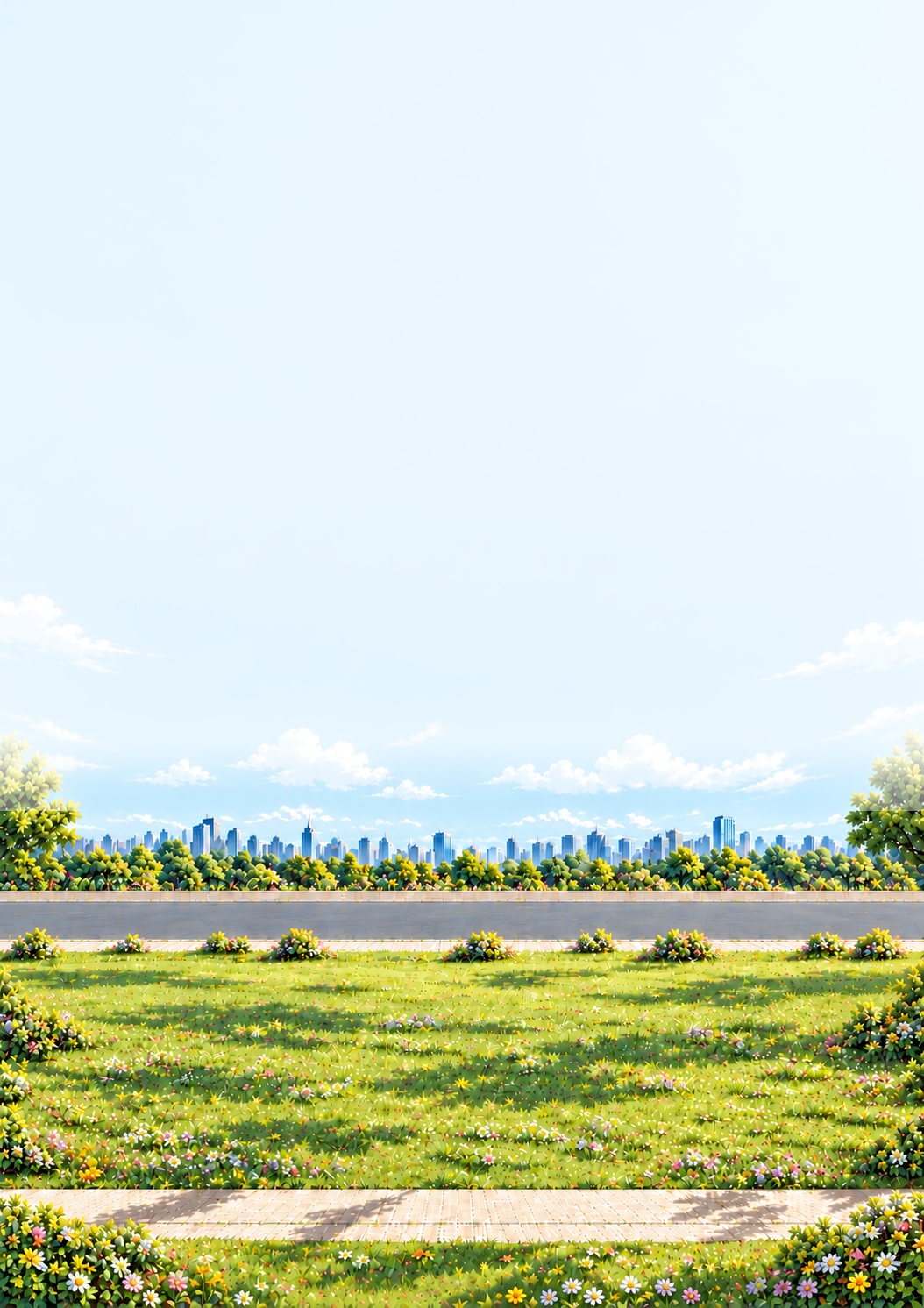}};
    \end{scope}
  \end{tikzpicture}}}
\thispagestyle{firstheader}
\maketitle

\newcommand{\coverfigcaption}{Four operations between video and world state: reading, writing, sharing, and interacting.}
\newsavebox{\coverbox}
\newlength{\coverfigh}
\par
\sbox{\coverbox}{\parbox{\linewidth}{\small\textbf{Figure 1:} \coverfigcaption}}%
\setlength{\coverfigh}{\dimexpr\pagegoal-\pagetotal-\ht\coverbox-\dp\coverbox-5pt-\intextsep-0.6cm\relax}%
\ifdim\coverfigh<5cm\setlength{\coverfigh}{5cm}\fi
\vfill
\begin{figure}[H]
  \centering
  \sbox{\coverbox}{\includegraphics[width=\dimexpr\textwidth*1354/1330\relax,height=\coverfigh,keepaspectratio]{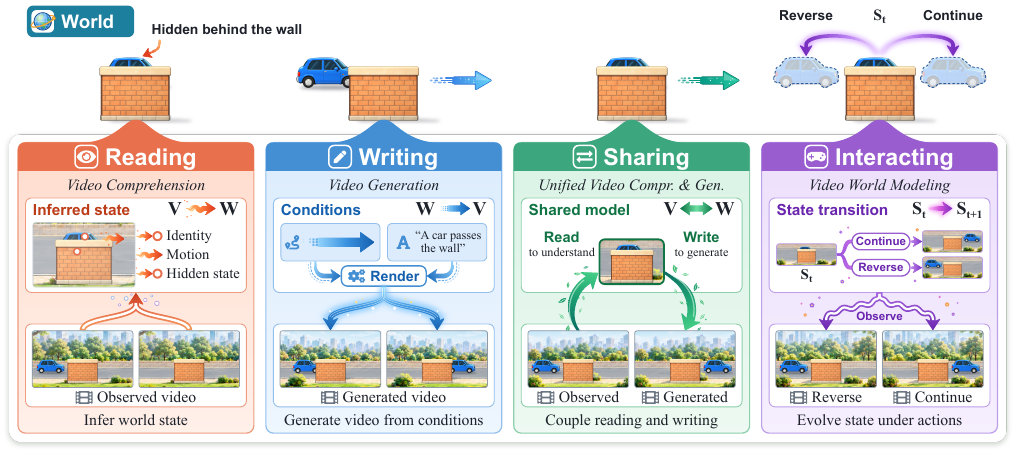}}%
  \global\coverfigw=\wd\coverbox
  \makebox[\linewidth][c]{%
  \begin{tikzpicture}[remember picture]
    \node[inner sep=0pt,outer sep=0pt] (teaserfig) {\usebox{\coverbox}};
  \end{tikzpicture}}
  \vspace{-4pt}%
  \label{fig:trace-operations}
\end{figure}
\vspace*{-1.8cm}
\clearpage

\pagestyle{fancy}
\fancyhf{}
\fancyhead[L]{\footnotesize\sffamily\bfseries\color{surveyNeutral}Video Foundation Models}
\fancyhead[R]{\footnotesize\sffamily\color{surveyMuted}World-Modeling Lens}
\fancyfoot[C]{\small\sffamily\color{surveyMuted}\thepage}
\renewcommand{\headrulewidth}{0.45pt}
\renewcommand{\headrule}{%
  \hbox to\headwidth{%
    \textcolor{surveyRead}{\rule{0.25\headwidth}{\headrulewidth}}%
    \textcolor{surveyWrite}{\rule{0.25\headwidth}{\headrulewidth}}%
    \textcolor{surveyShare}{\rule{0.25\headwidth}{\headrulewidth}}%
    \textcolor{surveyInteract}{\rule{0.25\headwidth}{\headrulewidth}}\hss}}
\setlength{\headheight}{41pt}

\hypersetup{linktoc=page}
\begingroup
\edef\surveySavedTocDepth{\number\value{tocdepth}}
\setcounter{tocdepth}{3}
\makeatletter
\let\surveyTocSection\l@section
\let\surveyTocSubsection\l@subsection
\let\surveyTocSubsubsection\l@subsubsection
\renewcommand{\l@section}[2]{\vspace{3pt}\surveyTocSection{#1}{#2}}
\renewcommand{\l@subsection}[2]{%
  \vspace{1.5pt}\surveyTocSubsection{#1}{#2}\vspace{1.5pt}}
\renewcommand{\l@subsubsection}[2]{%
  \vspace{1.5pt}\surveyTocSubsubsection{#1}{#2}\vspace{1.5pt}}
\makeatother
\tableofcontents
\setcounter{tocdepth}{\surveySavedTocDepth}
\endgroup
\clearpage

\begin{figure}[p]
  \centering
    \includegraphics[width=0.96\textwidth]{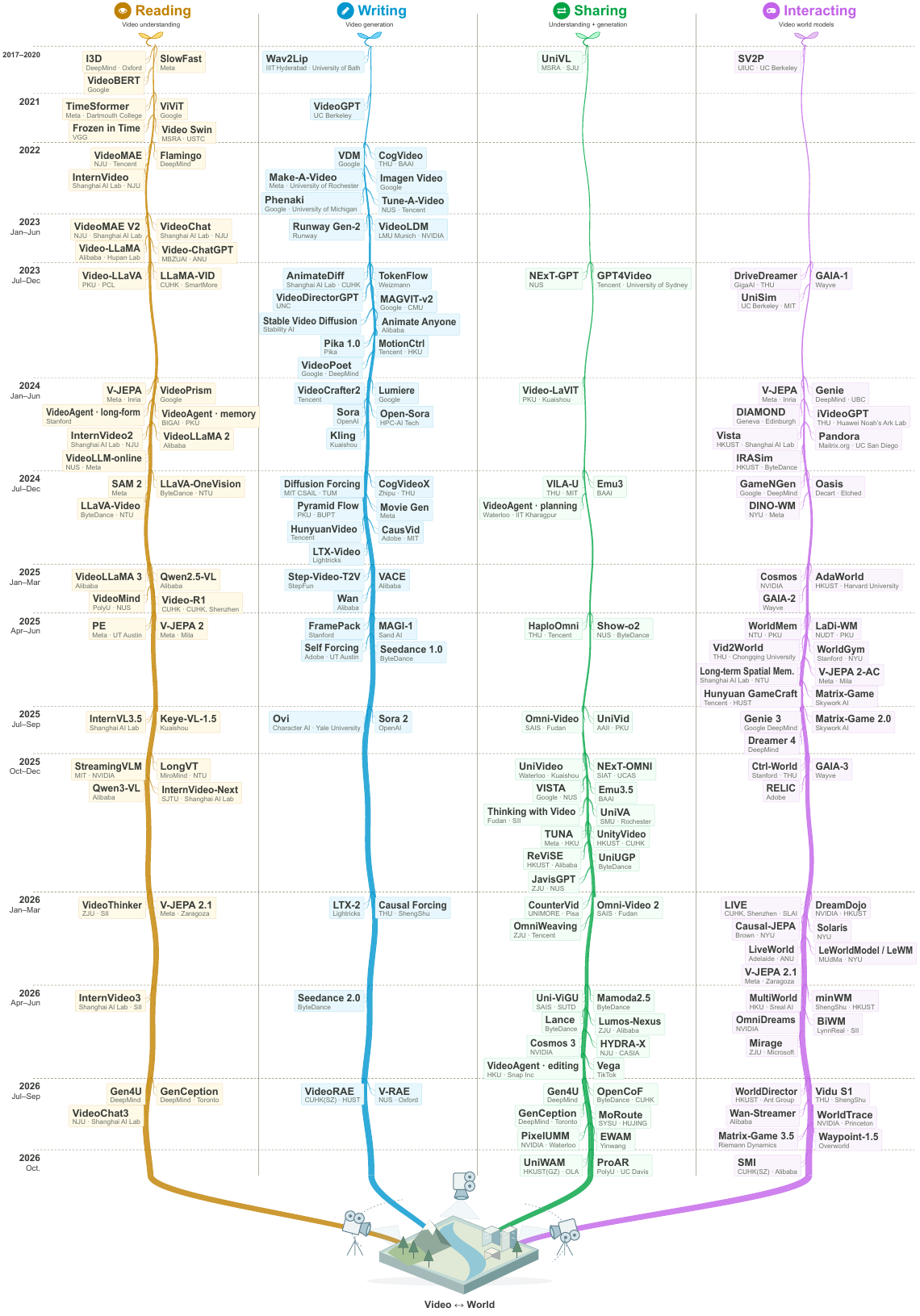}
  \caption{Selected milestones in the parallel evolution of video understanding, generation, shared representations and workflows, and world prediction and control.}
  \label{fig:field-timeline}
\end{figure}
\clearpage

\newcommand{\chapterlead}[1]{%
  \begin{tcolorbox}[
    enhanced,
    frame hidden,
    boxrule=0pt,
    colback=surveyCurrentAccent!4,
    borderline north={0.55pt}{0pt}{surveyCurrentAccent!32},
    sharp corners,
    left=8pt,
    right=7pt,
    top=5pt,
    bottom=5pt,
    before skip=2pt,
    after skip=7pt
  ]
  \small\itshape\color{surveyLeadText}#1
  \end{tcolorbox}
}

\newcommand{\surveyquote}[2]{%
  \begin{tcolorbox}[
    enhanced,
    colback=surveyLink!5,
    colframe=surveyLink!35,
    boxrule=0.5pt,
    arc=2.5mm,
    left=10pt,
    right=10pt,
    top=6pt,
    bottom=6pt,
    before skip=4pt,
    after skip=10pt
  ]
  \itshape\color{surveyInk}``#1''\par\smallskip
  {\raggedleft #2\par}
  \end{tcolorbox}
}

\newcommand{\surveybullet}{%
  \textcolor{surveyCurrentAccent}{\raisebox{0.18ex}{\rule{3.2pt}{3.2pt}}}}
\newcommand{\maplabel}[1]{%
  \textcolor{surveyCurrentAccent!65!black}{\sffamily\bfseries #1}}
\newcommand{\leadlabel}[1]{%
  \noindent{\sffamily\bfseries\color{surveyCurrentAccent!67!black}#1}}

\newcommand{\sectionmap}[1]{%
  \begin{compactitem}[\surveybullet]
  \raggedright
  \setlength{\itemsep}{1.2pt}
  \setlength{\parsep}{0pt}
  #1
  \end{compactitem}
}

\newcommand{\topic}[2]{\item \maplabel{#1.}\enspace #2}
\newcommand{\methodgroup}[1]{%
  \par\Needspace{3\baselineskip}\smallskip
  \noindent\textcolor{surveyCurrentAccent}{\rule[0.35ex]{1.15em}{1.4pt}}\hspace{0.45em}%
  {\sffamily\bfseries\color{surveyCurrentAccent!67!black}#1}\par\vspace{-0.7mm}}

\surveytheme{surveyNeutral}
\section{Introduction}
\surveyquote{Vision is the process of discovering from images what is present in the world, and where it is.}{David Marr, \emph{Vision}, 1982~\cite{marr1982vision}}
Video occupies a distinctive position among the modalities available to machine intelligence.
Unlike static images that capture individual moments, text that conveys selectively described information, or audio that primarily records acoustic dynamics, video directly captures how visible entities, their relations, and actions evolve across space and time, often together with synchronized audio, speech, and text.
This rich spatiotemporal information makes video essential to a wide range of applications, including communication, entertainment, autonomous driving, robotics, healthcare, and education.
Consequently, video has long been a central subject of computer vision and multimedia research, progressing from spatiotemporal architectures for action recognition~\cite{kay2017kinetics,carreira2017quo,feichtenhofer2019slowfast} to self-supervised pretraining for transferable video representations~\cite{tong2022videomae,wang2024internvideo2}.
More recently, advances in large language models (LLMs) and large-scale generative modeling have driven a shift from task-specific video models toward general-purpose video foundation models, expanding their capabilities along three complementary directions.
\textbf{For video understanding}, Video-LLMs such as Video-LLaVA~\cite{lin2024videollava} enable language-based interaction with visual sequences, while Qwen3-VL~\cite{bai2025qwen3} extends multimodal context to 256K tokens to accommodate longer video inputs. 
\textbf{For video generation}, models such as Sora~\cite{brooks2024video}, Wan~\cite{wan2025open}, HunyuanVideo~\cite{hunyuanvideo2024}, Movie Gen~\cite{polyak2024moviegen}, and Seedance 2.0~\cite{seedance2_2026} demonstrate increasingly capable synthesis under diverse conditions, with Sora and Loong~\cite{wang2024loong} further exploring minute-scale generation. 
\textbf{Toward the unification of understanding and generation}, emerging models~\cite{wang2024emu3,xie2025show,univideo2025} seek to integrate these capabilities through shared token representations, latent spaces, or modeling interfaces, bringing video perception and synthesis into a common foundation-model framework.

Beyond video understanding and generation, a broader research direction is emerging around \textbf{modeling how the physical world evolves}, rather than merely interpreting or synthesizing its visual observations.
Although language models acquire extensive knowledge through text, their representations are only indirectly grounded in physical experience, motivating the pursuit of \textit{world models} that can predict how an environment changes over time and support reasoning and planning~\cite{lecun2022path,worldmodelroadmap2026}.
This idea has roots in model-based reinforcement learning, where agents learn predictive dynamics to simulate possible futures and evaluate candidate actions before acting~\cite{ha2018worldmodels,hafner2023dreamerv3}.
Video provides a natural and scalable foundation for extending this paradigm beyond task-specific environments: by capturing continuous changes in objects, relations, and actions, large-scale video data offers rich supervision for learning world dynamics without requiring manually annotated labels, although actions and underlying physical states often remain unobserved.
Building on this opportunity, recent research has explored several complementary approaches to video-based world modeling. 
Generative models such as Sora~\cite{brooks2024video} have motivated the prospect of learning world dynamics through video synthesis, while Genie~\cite{bruce2024genie} demonstrates how interactive environments can be learned from unlabeled videos.
Action- and condition-aware models such as GAIA-1~\cite{hu2023gaia1} and Cosmos~\cite{nvidia2025cosmos} extend predictive generation toward driving and physical AI, whereas V-JEPA 2~\cite{assran2025vjepa2} and Dreamer 4~\cite{hafner2025dreamer4} explore predictive representations and imagined trajectories for planning and control.
Together with progress in interactive simulation~\cite{yang2024unisim}, egocentric and robotic learning~\cite{grauman2022ego4d,wu2023gr1}, and long-horizon prediction~\cite{xiao2025worldmem}, these developments are expanding the role of video from a medium for perception and generation toward a foundation for predicting and interacting with dynamic environments.
Figure~\ref{fig:field-timeline} traces milestones of these lines of work.

These two lines of progress are increasingly converging around a common challenge: \textit{connecting video observations with the dynamics of the world they depict}.
Video world models must infer underlying states from observed sequences before predicting how those states evolve, and often translate predicted states back into visual observations for simulation, evaluation, or planning.
Conversely, video generators are increasingly incorporating action- and state-dependent controls, while unified video models explicitly couple visual understanding with synthesis.
Despite their different objectives, these approaches therefore share a fundamental question: \textit{what relationship does a model learn between an observed video and the evolving world that gives rise to it?}

To examine this question systematically, we adopt \textit{world modeling as a unifying perspective} on video foundation models.
Specifically, we view an \textbf{observed video} $V$ as a partial, lossy, and temporally ordered observation of an underlying \textbf{latent temporal world} $W$.
This distinction provides a common basis for organizing four complementary research paradigms: \readop{reading}, which infers world properties from observed video; \writeop{writing}, which synthesizes video observations from specified or learned world conditions; \shareop{sharing}, which integrates both directions within a unified modeling framework; and \interactop{interacting}, which updates or predicts world states in response to actions and evolving observations. 
These paradigms characterize different ways in which models connect video observations to the underlying world, rather than merely categorizing models by their input and output modalities. 
This perspective enables a systematic comparison of their objectives, capabilities, and evaluation criteria while preserving the distinctions between visual plausibility, temporal coherence, and consistency of the modeled world.
Section~\ref{sec:lens} formalizes this framework and clarifies how these properties differ across the four paradigms.

Existing relevant surveys have examined related developments from different perspectives, including general world models organized around predictive state representations across modalities~\cite{ding2024worldmodelsurvey}, interactive video world models centered on online prediction and control~\cite{liu2026interactivevideoworld,luo2026ai}, 3D and 4D world modeling grounded in explicit geometric representations~\cite{kong2025world3d4d}, multimodal language models emphasizing the integration of visual encoders and language reasoning~\cite{yin2023mllmsurvey}, long-form video generation focusing on temporal extension and consistency~\cite{li2024survey}, and audio-visual foundation models spanning complementary modalities~\cite{qin2026audiovisual}.
While these surveys provide valuable insights into individual research directions, they do not collectively establish a unified account of how video models connect observations to evolving world states across understanding, generation, unification, and interaction.
\textit{Our survey addresses this gap by taking the relation between video observations and the underlying dynamic world as its central organizing principle}, not any single architecture, task, or modality.

Under this perspective, we systematically examine the four paradigms of reading, writing, sharing, and interacting through a common analytical framework.
For each paradigm, we review the tasks, datasets, and evaluation protocols that define its objectives; the representations and model architectures that support its capabilities; the reasoning and control mechanisms that extend its functionality; and the applications that demonstrate its practical utility.
Beyond reviewing individual paradigms, we identify several cross-cutting questions: how models represent and preserve world states beyond immediate observations, how actions and external conditions influence state evolution, and how predictive capabilities translate into effective reasoning and decision-making. 
These questions also motivate a closer examination of evaluation practices, particularly the distinction between conventional metrics of recognition accuracy or perceptual fidelity and emerging criteria concerning state persistence, controllability, physical and causal consistency, and downstream utility. 
Within the interacting paradigm, we further distinguish video world models along two complementary dimensions: \textbf{prediction space}, ranging from observation-space to latent-state prediction, and \textbf{transition interface}, ranging from passive prediction to action-conditioned rollout and closed-loop interaction.
This taxonomy separates what a model predicts from how its predictions are conditioned, updated, and used.

Our review covers publicly available research that contributes relevant architectural, methodological, or evaluation evidence. 
We retain foundational studies where necessary to establish conceptual and technical developments, while interpreting recent advances in the context of their specific tasks, training data, and evaluation protocols. 
Throughout the survey, we distinguish between the capabilities supported by a model's design, the supervision provided by its training data, and the empirical evidence demonstrated by its evaluation, avoiding assumptions that strong performance on individual video tasks necessarily implies consistent or generalizable world modeling.

Specifically, this survey makes three main contributions:
\sectionmap{
\topic{A world-modeling lens}{The four-operation framework of reading, writing, sharing, and interacting relates video understanding, video generation, unified video models, and video world models in one account, plus a two-axis classification of world models by prediction space and transition interface.}
\topic{A comprehensive scope with insights}{Across tasks, representations, models, systems, datasets, and benchmarks, we compare how the four operations represent state, learn from video, accept control, preserve memory, and are evaluated.
The synthesis distills recurring design patterns and failure modes into explicit insights, and identifies open directions, including long-horizon consistency, physical and causal grounding, measurable synergy between understanding and generation, and standardized evaluation of world properties.}
\topic{Open resources}{A public reading list, taxonomy, codes, data, and resources are all categorized and available at \href{https://vid-gen.github.io/video-world-model-survey}{project page}.}
}

\begin{figure}[p]
  \centering
  \resizebox{\textwidth}{!}{\input{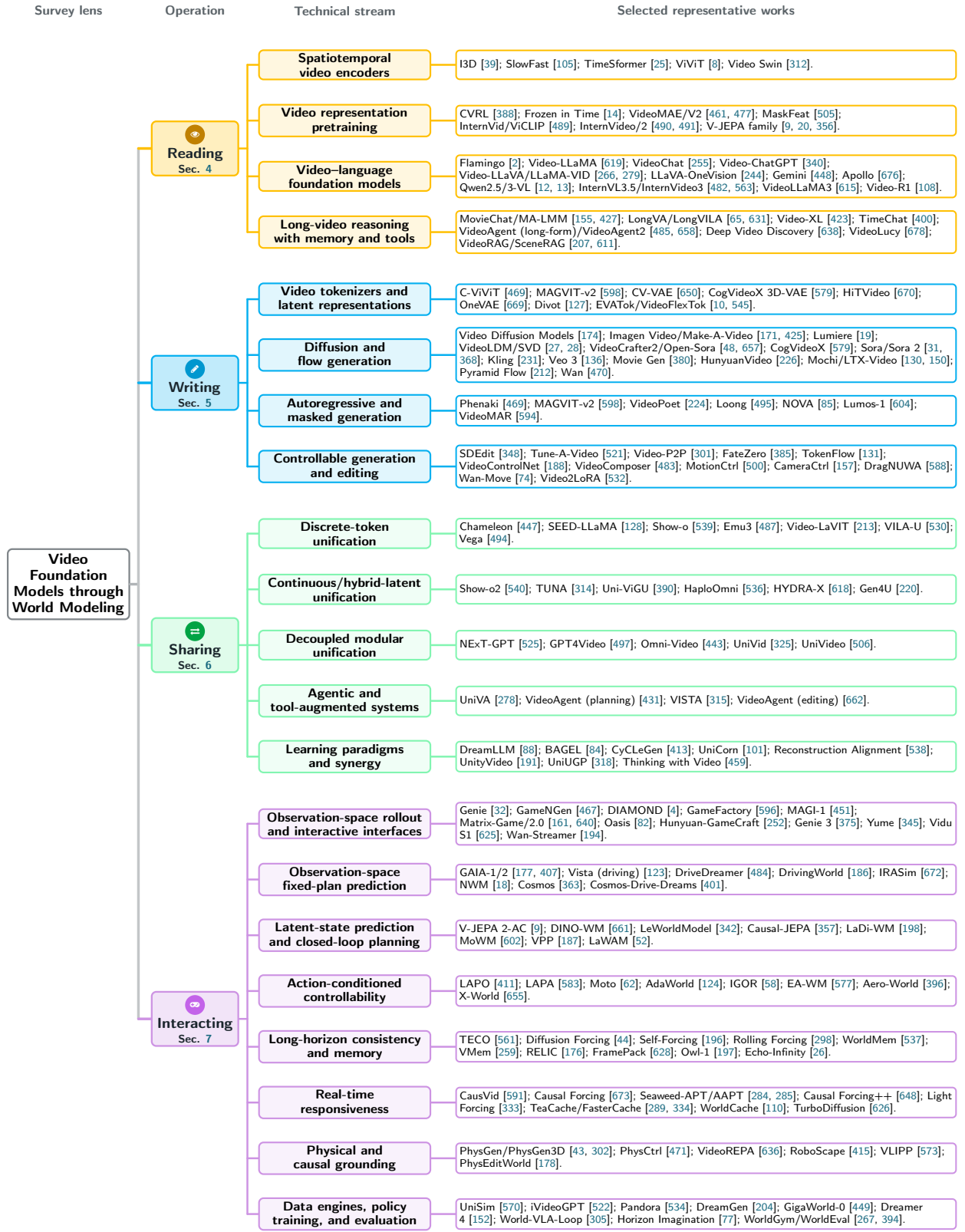}}
  \caption{Knowledge graph of the four research fields.
  Reading, writing, sharing, and interacting branch into technical streams and representative works.}
  \label{fig:survey-knowledge-graph}
\end{figure}

The remainder of this survey proceeds as follows.
Section~\ref{sec:lens} formalizes the video--world relation, the two forms of consistency, and the four operations.
Section~\ref{sec:preliminaries} assembles the shared background: recurring video scenarios, language-model and generative-modeling foundations, and the definition and taxonomy of video world models.
Sections~\ref{sec:reading} through~\ref{sec:interacting} then follow the operations from observing a trace to evolving a world state.
Section~\ref{sec:reading} studies how models read semantic, spatial, temporal, physical, and causal properties from video;
Section~\ref{sec:writing} examines the representations, generative paradigms, and controls used to write traces;
Section~\ref{sec:sharing} turns to the architectures and training regimes that connect the two directions; and
Section~\ref{sec:interacting} completes the progression with open-loop and online models that evolve video-derived state under actions or conditions.
Figure~\ref{fig:survey-knowledge-graph} condenses the resulting landscape: each operation branches into parallel research streams and then into the representative works examined in those chapters.
Section~\ref{sec:challenges} distills the challenges that cut across the four operations into directions for future work, and a brief conclusion closes the survey.

\surveytheme{surveyNeutral}

\begin{table}[!t]
\centering
\small
\caption{Notation used throughout the survey.}
\label{tab:notation}
\setlength{\tabcolsep}{5pt}
\renewcommand{\arraystretch}{1.12}
\surveytablecolors
\begin{tabularx}{\textwidth}{@{}>{\raggedright\arraybackslash}p{3.6cm} Y >{\raggedright\arraybackslash}p{2.6cm}@{}}
\toprule
\surveytablehead
\tableheadcell{Symbol} & \tableheadcell{Meaning} & \tableheadcell{Introduced in} \\
\midrule
\surveygroup{3}{World and video}
$t\in\{1,\dots,T\}$ & video time; $T$ is the number of frames & Section~\ref{sec:lens-objects} \\
$W=(w_t)_{t\ge1}$, $\mathcal{W}$ & latent temporal world and its state space & Definition~\ref{def:world} \\
$\mathcal{T}(\,\cdot\mid w_t,a_t)$ & world dynamics & Definition~\ref{def:world} \\
$a_t$ & action or external intervention; may be absent & Definition~\ref{def:world} \\
$V=(x_1,\dots,x_T)$ & observed video and its frames & Definition~\ref{def:video} \\
$\mathcal{O}(\,\cdot\mid w_t,\pi_t)$, $\pi_t$ & sensing process and camera configuration & Definition~\ref{def:video} \\
$x_{\le t}$, $x_{<i}$, $x_{1:T}$ & elements up to $t$, before $i$, and from $1$ to $T$ of any sequence & Definition~\ref{def:state} \\
\addlinespace
\surveygroup{3}{Models and the four operations}
$\theta$; $\theta_{\mathrm{R}}$, $\theta_{\mathrm{W}}$ & model parameters; reading- and writing-map parameters in sharing & Definitions~\ref{def:state}, \ref{def:operations} \\
$s_t\sim q_\theta(\cdot\mid x_{\le t},a_{<t})$ & represented state and the map that infers it (deterministic $f_\theta$ as a special case); in general $s_t\neq w_t$ & Definition~\ref{def:state} \\
$\hat V=(\hat x_1,\dots,\hat x_T)$ & trace generated or predicted by a model & Definition~\ref{def:consistency} \\
$d$, $\delta$ & local discrepancy between neighboring frames and its tolerance & Definition~\ref{def:consistency} \\
$\mathcal{C}$ & admissible world trajectories (identity, relations, dynamics, physical and causal constraints) & Definition~\ref{def:consistency} \\
$\phi(w_t)$, $\hat\phi_t$, $r_\theta$ & target world property, its estimate, and the reading distribution & Definition~\ref{def:operations} \\
$c$, $c_t$ & condition (text, image or video prefix, audio, camera motion, trajectory, or layout); $c_t$ is the condition at step $t$ & Definition~\ref{def:operations} \\
$p_\theta$ & distribution over traces (writing) or next states (interacting) & Definition~\ref{def:operations} \\
$g_\theta$ & decoder from represented state to observation & Definition~\ref{def:operations} \\
$H$ & prediction horizon of a world model & Figure~\ref{fig:worldmodel-taxonomy} \\
\addlinespace
\surveygroup{3}{Video-LLMs and generative modeling}
$E_v$, $P$ & video encoder and connector to the LLM & Section~\ref{sec:prelim-videollm} \\
$z$, $z_{1:L}$ & video-derived features or latents; $L$ embeddings passed to the LLM & Section~\ref{sec:prelim-videollm} \\
$z$, $z_0$ & generative target (latent, video tokens, or frames) and its clean sample & Section~\ref{sec:prelim-generative} \\
$z_k$, $k$, $K$ & noisy latent at denoising step $k\in\{1,\dots,K\}$ (continuous $k\in[0,1]$ for flow matching), distinct from video time $t$ & Section~\ref{sec:prelim-generative} \\
$\epsilon$, $\epsilon_\theta$, $v_\theta$ & Gaussian noise, noise-prediction network, and velocity field & Section~\ref{sec:prelim-generative} \\
\bottomrule
\end{tabularx}
\end{table}

\section{The World-Modeling Lens}
\label{sec:lens}
This section defines the world and its video (Section~\ref{sec:lens-objects}), two forms of consistency used later for evaluation (Section~\ref{sec:lens-consistency}), and the four operations that organize Sections~\ref{sec:reading}--\ref{sec:interacting} (Section~\ref{sec:lens-operations}).
Table~\ref{tab:notation} lists the notation.

\subsection{World, Video, and Represented State}
\label{sec:lens-objects}
We distinguish three objects: the evolving world, the video through which it is observed, and the representations that a model constructs from available information.
The following definitions provide a conceptual abstraction:

\begin{definition}{Latent temporal world}{world}
The latent temporal world is a stochastic process $W=(w_t)_{t\ge1}$ over a state space $\mathcal{W}$.
Each state $w_t\in\mathcal{W}$ encompasses entities, their attributes and relations, and relevant physical and dynamic variables.
Assuming that $w_t$ contains sufficient information for predicting subsequent evolution, its dynamics can be expressed as
\[
  w_{t+1}\sim\mathcal{T}(\,\cdot\mid w_t,a_t),
\]
where $a_t$ denotes an optional intervention or control input, with uncontrolled evolution recovered when no such input is supplied.
\end{definition}

\begin{definition}{Observed video}{video}
An observed video $V=(x_1,\dots,x_T)$ is a temporally ordered sequence of visual observations generated by a sensing process, $x_t\sim\mathcal{O}(\,\cdot\mid w_t,\pi_t)$, where $\pi_t$ is the camera configuration.
We refer to an observed or generated video $V$ as a
\emph{trace}.
This trace is
\emph{partial}: $x_t$ depends only on the part of $w_t$ visible under $\pi_t$, which changes with camera motion, shot changes, or multiple cameras;
\emph{lossy}: sensing, resolution, compression, and editing remove information, so distinct world states may produce indistinguishable observations; and
\emph{temporally ordered}: frame order and duration carry evidence about motion and state transition, and thus about $\mathcal{T}$.
\end{definition}

\begin{definition}{Represented state}{state}
A model with parameters $\theta$ may construct an internal representation $s_t\sim q_\theta(\cdot\mid x_{\le t},a_{<t})$, inferred from the observations and actions available up to time $t$; a deterministic encoder $s_t=f_\theta(x_{\le t},a_{<t})$ is the special case of a point mass, and a transition model may instead sample $s_{t+1}$ from its own predictions (Definition~\ref{def:operations}).
In general, $s_t\neq w_t$, and its information about
the underlying world may be incomplete or inaccurate.
\end{definition}

\leadlabel{Remark.}
Definitions~\ref{def:world}--\ref{def:state} describe a partially observable dynamical system, closely related to the environment model underlying a partially observable Markov decision process.
Such formulations are widely used in model-based reinforcement learning~\cite{ha2018worldmodels,hafner2023dreamerv3}, where agents infer predictive states from observations and simulate possible futures under candidate actions.
For general video foundation models, however,
the represented state may be implicit or constructed
from an entire sequence rather than updated online.

\subsection{Two Forms of Consistency}
\label{sec:lens-consistency}
Because video observes the world only partially (Definition~\ref{def:video}), a trace can be consistent at two levels.

\begin{definition}{Temporal coherence and world consistency}{consistency}
Let $\hat V=(\hat x_1,\dots,\hat x_T)$ be a trace that a model generates or predicts.
$\hat V$ is \emph{temporally coherent} if neighboring frames agree locally, $d(\hat x_t,\hat x_{t+1})\le\delta$ for all $t$, where $d$ penalizes implausible local motion, appearance change, or flicker.
$\hat V$ is \emph{world-consistent} relative to
specified dynamics and constraints, if its
observations are compatible with at least one
admissible world trajectory:
\[
  \exists\,w_{1:T}\in\mathcal{C},\ \pi_{1:T}
  \quad\text{s.t.}\quad
  \hat x_t\in
  \operatorname{supp}
  \mathcal{O}(\cdot\mid w_t,\pi_t),
  \quad \forall t,
\]
where $\mathcal{C}$ contains the trajectories that satisfy the assumed dynamics, identity persistence,
spatial relations, and relevant physical and causal
constraints under the specified interventions.
When the camera path is specified as a condition, $\pi_{1:T}$ is fixed to it instead of quantified.
\end{definition}

\begin{observation}{Coherence-consistency}{consistency}
Neither property implies the other.
An object that changes identity after leaving the frame, a room that alters its layout between visits, or an action whose effect later disappears can yield locally smooth frames that no trajectory in $\mathcal{C}$ explains; conversely, a hard cut violates local smoothness while remaining world-consistent.
\end{observation}

Temporal coherence is a property of the trace; world consistency is a property of the world the trace implies.
A recorded video is world-consistent by construction, because a real trajectory produced it, so the distinction matters for traces that a model writes or predicts.
World-modeling evidence must therefore test world consistency, not only temporal coherence.
As $\mathcal{O}$ and $\mathcal{C}$ are unobserved, Definition~\ref{def:consistency} is an evaluation target, not a metric; benchmarks in Sections~\ref{sec:writing-eval} and~\ref{sec:wm-benchmarks} probe it via revisits, occlusion, and interventions.

\subsection{The Four Operations}
\label{sec:lens-operations}

\begin{definition}{Four operations}{operations}
\sectionmap{
\topic{\readop{Reading}\enspace$V\rightarrow W$\enspace\textmd{(video comprehension)}}{
Reading infers selected properties of the world
from observed video, $\hat\phi_t\sim r_\theta(\cdot\mid s_t)$, an estimate of $\phi(w_t)$, where $s_t$ is constructed from the available observations and $\phi$ ranges over semantic, spatial, temporal, physical, and causal properties. It is evaluated by accuracy against $\phi(w_t)$ and by calibrated uncertainty for occluded or off-screen content.}
\topic{\writeop{Writing}\enspace$W\rightarrow V$\enspace\textmd{(video generation)}}{Writing samples a trace from a condition, $\hat x_{1:T}\sim p_\theta(x_{1:T}\mid c)$, where $c$ (text, an image or video prefix, audio, camera motion, trajectories, or layouts) specifies the world only in part and the generator resolves the rest.
It is evaluated by perceptual quality, adherence to $c$, and world consistency of $\hat V$ (Definition~\ref{def:consistency}).}
\topic{\shareop{Sharing}\enspace$V\leftrightarrow W$\enspace\textmd{(unified comprehension and generation)}}{Sharing couples a reading map with parameters $\theta_{\mathrm{R}}$ and a writing map with parameters $\theta_{\mathrm{W}}$ through at least one shared component: parameters ($\theta_{\mathrm{R}}\cap\theta_{\mathrm{W}}\neq\varnothing$), tokens or representations, an interface, or a training objective.
It is evaluated at three levels: capability coverage, architectural sharing, and functional synergy, that is, measurable improvement of reading, writing, or both under matched resources.}
\topic{\interactop{Interacting}\enspace$W\circlearrowleft$\enspace\textmd{(video world modeling)}}{Interacting rolls out a transition model, $s_{t+1}\sim p_\theta(s_{t+1}\mid s_{\le t},a_t,c_t)$, with $\hat x_{t+1}=g_\theta(s_{t+1})$ when visible output is needed.
The state $s_t$ may be frames, video tokens, or latent features, and the actions $a_t$ and conditions $c_t$ are fixed before an open-loop rollout or supplied online.
It is evaluated by whether predicted changes follow these inputs and stay world-consistent (Definition~\ref{def:consistency}), beyond the quality of rendered frames.}
}
\end{definition}

\begin{table}[!t]
\centering
\small
\caption{Evidence ladder for claims about the world. Each operation can be supported at four increasing levels; stronger world-modeling claims require the higher levels. Benchmarks are representative, not exhaustive.}
\label{tab:evidence-ladder}
\surveytablecolors
\setlength{\tabcolsep}{3.5pt}
\renewcommand{\arraystretch}{1.12}
\begin{tabularx}{\textwidth}{@{}>{\raggedright\arraybackslash}p{2.15cm} Y Y Y Y@{}}
\toprule
\surveytablehead
\tableheadcell{Operation} & \tableheadcell{1. Plausible output} & \tableheadcell{2. Faithful to evidence or conditions} & \tableheadcell{3. Consistent world state} & \tableheadcell{4. Decision utility} \\
\midrule
\readop{Reading} &
correct labels and answers on clips~\cite{kay2017kinetics,fu2025video} &
answers depend on the video, with shortcuts controlled~\cite{ok2025tempcore,song2026videoindex,krojer2025mvp} &
state kept across occlusion, revisits, and streams~\cite{ma2026beyond3d,li2026ovosbench,liu2026svcbench} &
evidence supports agent or driving decisions~\cite{an2026agentvidbench,sima2024drivelm} \\
\addlinespace
\writeop{Writing} &
perceptual quality and temporal coherence~\cite{huang2023vbenchcomprehensivebenchmarksuite} &
adherence to text, motion, and reference conditions~\cite{sun2025t2vcompbenchcomprehensivebenchmarkcompositional,chu2025wanmove} &
world consistency (Definition~\ref{def:consistency}): identity, layout, physics~\cite{bansal2025videophy2,motamed2025physicsiq} &
generated data or plans improve downstream models~\cite{jang2025dreamgen,shang2026worldarena} \\
\addlinespace
\shareop{Sharing} &
both directions work in one system~\cite{li2024mvbench,zheng2025vbench20advancingvideogeneration} &
integrated, source-faithful transformations~\cite{wei2026univbenchunifiedevaluationvideo,liu2025revise} &
measured transfer under matched data, capacity, and compute~\cite{fu2026lance,poppi2026countervid} &
shared prediction improves planning or action~\cite{lu2025uniugp,wang2026ewam} \\
\addlinespace
\interactop{Interacting} &
plausible predicted frames or features &
predicted changes follow fixed or online controls~\cite{arai2024actbench,fang2026iworldbench} &
persistent state and valid responses to interventions~\cite{ye2026mind,memobench2026,li2025worldmodelbench,whatifworld2026} &
better planning, policy training, or policy ranking~\cite{quevedo2025worldgym,zhang2025worldinworld,shang2026worldarena} \\
\bottomrule
\end{tabularx}
\end{table}

The arrows here are conceptual shorthand: writing reaches $W$ only through $c$, and reading and writing are not exact inverses.
Stronger claims about the represented state need stronger evidence; Table~\ref{tab:evidence-ladder} gives four levels for each operation, from plausible output to decision utility.
One system can perform several operations: a unified editor reads a source clip before writing its revision, and an interactive world model reads its history, writes each next frame, and updates its state in one backbone.
Figure~\ref{fig:field-timeline} places milestones of the four lines of work on one timeline, from video encoders~\cite{feichtenhofer2019slowfast,arnab2021vivit} and generators~\cite{ho2022video,henschel2025streamingt2v} to unified models~\cite{wang2024emu3,xie2025show} and action-conditioned simulators~\cite{hu2023gaia1,bruce2024genie}.

\surveytheme{surveyNeutral}
\section{Preliminaries}
\label{sec:preliminaries}
This section covers the background shared by the four operations: video scenarios (Section~\ref{sec:video-scenarios}), video large language models (Section~\ref{sec:prelim-videollm}), generative modeling (Section~\ref{sec:prelim-generative}), and the scope and taxonomy of video world models (Section~\ref{sec:wm-taxonomy}).

\subsection{Video Scenarios}
\label{sec:video-scenarios}
Video modeling spans diverse scenarios that differ in temporal extent, observation perspective, modality composition, information availability, interaction, and domain-specific constraints. Rather than forming mutually exclusive categories, these characteristics impose complementary demands on what a model must observe, retain, predict, and control.
Table~\ref{tab:scenarios} lists the evidence, modeling demands, and representative resources of each scenario.
\begin{table}[t]
\centering
\small
\begin{adjustbox}{max totalsize={\textwidth}{0.86\textheight},center}
\begin{threeparttable}
\caption{Video scenarios, their defining evidence conditions, modeling demands, and representative resources.}
\label{tab:scenarios}
\surveytablecolors
\setlength{\tabcolsep}{4.2pt}
\renewcommand{\arraystretch}{1.04}
\begin{tabularx}{\textwidth}{@{}>{\hsize=.70\hsize}Y >{\hsize=1.05\hsize}Y >{\hsize=1.05\hsize}Y >{\hsize=1.20\hsize}Y@{}}
\toprule
\surveytablehead
\tableheadcell{Scenario} & \tableheadcell{Evidence} &
\tableheadcell{Modeling demands} & \tableheadcell{Datasets / benchmarks} \\
\midrule
\textbf{Regular short video} &
short, passive clips with bounded actions or events &
clip-level semantics, cross-modal matching, and local temporal structure &
Kinetics~\cite{kay2017kinetics}, Something-Something~\cite{goyal2017something}, MSR-VTT~\cite{Xu_2016_CVPR} \\
\textbf{Sounding video} &
synchronized audio and visual evidence &
cross-modal binding, temporal localization, and modality-balanced inference &
AudioSet~\cite{audioset2017}, VGGSound~\cite{vggsound2020}, AVE~\cite{ave2018}, OmniVideoBench~\cite{li2025omnivideobench} \\
\textbf{Egocentric video} &
first-person activity with body-coupled camera motion and hand--object interaction &
object-state persistence, procedural memory, and ego--exo correspondence &
Ego4D~\cite{grauman2022ego4d}, EPIC-KITCHENS~\cite{damen2022rescaling}, Ego-Exo4D~\cite{grauman2024egoexo4d}, EgoSchema~\cite{mangalam2023egoschema}, HD-EPIC~\cite{perrett2025hdepic}, EgoLife~\cite{yang2025egolife} \\
\textbf{Streaming and online video} &
frames arrive incrementally; outputs are required before the stream ends &
causal state updates, bounded rolling memory, and timely or proactive response &
OVO-Bench~\cite{niu2025ovo}, SVBench~\cite{yang2025svbench} \\
\textbf{Long-form video} &
minutes-to-hours records with sparse, temporally distant evidence &
evidence retrieval, event linkage, compression, and long-range memory &
MLVU~\cite{zhou2025mlvu}, LVBench~\cite{wang2025lvbench}, Video-MME (long)~\cite{fu2025video}, LongVideoBench~\cite{wu2024longvideobench}, HourVideo~\cite{chandrasegaran2024hourvideo}, VideoOdyssey~\cite{videoodyssey2026} \\
\textbf{Interactive video} &
actions or instructions change future observations during rollout &
action fidelity, persistent state, closed-loop latency, and decision utility &
ACT-Bench~\cite{arai2024actbench}, MIND~\cite{ye2026mind}, WorldArena~\cite{shang2026worldarena} \\
\textbf{Domain-specific video} &
domain-specific semantics, workflows, sensors, and costs of error &
specialist knowledge, fine-grained supervision, and domain-calibrated evaluation &
nuScenes~\cite{caesar2020nuscenes}, DriveLM~\cite{sima2024drivelm}; Cholec80~\cite{twinanda2016endonet}; SoccerNet~\cite{giancola2018soccernet}; HowTo100M~\cite{Miech_2019_ICCV}, COIN~\cite{tang2019coin} \\
\bottomrule
\end{tabularx}
\end{threeparttable}
\end{adjustbox}
\end{table}

Three labels describe the recorded content: its default form, its modalities, and its viewpoint.
\textbf{Regular short video}, the standard setting for representation learning and clip-level understanding, consists of passive third-person clips of seconds to minutes, whose audio is usually ignored.
Its datasets annotate action classes~\cite{kay2017kinetics}, actions defined by temporal change~\cite{goyal2017something}, or captions~\cite{Xu_2016_CVPR}, but do not test memory, online response, or control.
\textbf{Sounding video} treats synchronized audio as evidence, with tasks that range from audio-event recognition and source localization to generating events whose sound and appearance must agree~\cite{audioset2017,vggsound2020,ave2018}.
Its benchmarks must rule out visual or language shortcuts.
\textbf{Egocentric video} is recorded from a wearer's or agent's viewpoint, so hands, manipulated objects, gaze, and body-coupled camera motion enter the stream~\cite{grauman2022ego4d,damen2022rescaling,grauman2024egoexo4d,mangalam2023egoschema}.
The task shifts from recognizing actions to tracking object state and intent across long activities with frequent occlusion.

Three further labels describe how the model accesses observations and whether it can change them.
In \textbf{streaming and online video}, frames arrive incrementally, so a model must update its state and respond within bounded latency, without access to future frames.
Typical tasks are online question answering and streaming captioning, which may require timestamped or proactive responses~\cite{niu2025ovo,yang2025svbench}.
The same recording can be processed offline in full or presented as a stream.
\textbf{Long-form video} spans minutes to hours: the full recording is usually available, but not every frame can be processed at full resolution, and the evidence for a query is sparse and spread across distant segments~\cite{zhou2025mlvu,wang2025lvbench,fu2025video,wu2024longvideobench}.
A model must therefore decide which evidence to keep and link events that lie far apart in time.
\textbf{Interactive video} lets a human, policy, or agent supply actions that change what is observed next~\cite{arai2024actbench,ye2026mind,shang2026worldarena}.
Unlike in a passive stream, the future then depends on the interaction history; interactive world models (Section~\ref{sec:wm-taxonomy}) target this setting.

\textbf{Domain-specific video} cuts across all of these labels: driving~\cite{caesar2020nuscenes,sima2024drivelm}, surgery~\cite{twinanda2016endonet}, sports~\cite{giancola2018soccernet}, and instructional video~\cite{Miech_2019_ICCV,tang2019coin} impose their own event structure, terminology, and costs of error.
These domains test whether general video representations transfer to rare events and specialized outputs.

\subsection{Video Large Language Models}
\label{sec:prelim-videollm}

\leadlabel{Connecting video to an LLM.}
A Video-LLM feeds video into a pretrained large language model (LLM), usually an autoregressive Transformer~\cite{vaswani2017attention,Touvron2023LLaMAOA} trained by next-token prediction (Equation~\eqref{eq:autoregressive}).
The LLM learns to follow instructions through supervised instruction tuning~\cite{wei2022finetuned}, which visual instruction tuning extends to images~\cite{liu2023visual}, and is further aligned with human or model feedback~\cite{ouyang2022training} and verifiable rewards~\cite{lambert2025tulu3pushingfrontiers,guo2025deepseek}.
Inside a video system, it understands and generates language, answers questions, and sometimes acts as a controller that decomposes tasks, selects tools, and reasons over several steps.
In a common division of labor, a video encoder supplies perceptual features, the LLM interprets or organizes them, and a decoder, generator, or world model produces pixels or state transitions.
To connect video, a video encoder $E_v$ maps frames or clips $x_{1:T}$ to visual features, and a connector $P$ converts these features into a language-compatible sequence:
\begin{equation}
  z_{1:L}=P\!\left(E_v(x_{1:T})\right).
\end{equation}
Connectors range from linear projectors~\cite{liu2023visual} to learned-query resamplers~\cite{jaegle2021perceiver,pmlr-v202-li23q} and gated cross-attention~\cite{alayrac2022flamingo}; Section~\ref{sec:reading-videollm} compares them and the video-specific training recipes.

\leadlabel{Video tokens and context.}
In the Video-LLM literature, a \emph{video token} usually denotes one of the continuous embeddings in $z_{1:L}$ passed to the LLM.
Autoregressive video generators often use the same term for discrete codebook indices, a different representation (Section~\ref{sec:writing}).
Frame sampling, temporal pooling, and the connector set the number $L$ of embeddings kept from a video.
A larger $L$ preserves finer spatial and temporal evidence but costs more context, memory, and attention; stronger compression is cheaper but can remove brief events, small objects, or long-range relations.
Because the context window is fixed and also holds the instructions, earlier dialogue, and the output, long-video models must split the remaining budget between temporal coverage and visual detail.

\subsection{Generative Modeling Foundations}
\label{sec:prelim-generative}
\leadlabel{Conditional generation.}
Whereas the LLM mainly produces language, generative models produce the visual output of a video system.
Writing samples a trace $\hat x_{1:T}\sim p_\theta(x_{1:T}\mid c)$ (Definition~\ref{def:operations}).
Video generators usually work in the latent space of a variational autoencoder (VAE)~\cite{kingma2013auto}, whose decoder may be trained with an adversarial loss~\cite{goodfellow2020generative}: an encoder maps the video to a latent $z$, the generator models $p_\theta(z\mid c)$, and the decoder maps a sampled latent back to frames.
We write $z$ for whichever target is modeled, whether continuous latents, discrete video tokens, or, for pixel-space models, the frames themselves.
In general, a condition may specify content, such as a prompt or a reference image, or control how the state evolves, such as actions, trajectories, or camera paths.
The same modality can play different roles: an image may specify appearance, provide the initial observation, or serve as a spatial reference.
Action-conditioned and interactive world models (Section~\ref{sec:wm-taxonomy}) require the second kind.

\leadlabel{Autoregressive and masked modeling.}
When the target is a sequence of ordered units $z_{1:N}$, an autoregressive model factorizes the conditional distribution as
\begin{equation}
  p_\theta(z_{1:N}\mid c)
  =\prod_{i=1}^{N}p_\theta(z_i\mid z_{<i},c)
  \label{eq:autoregressive}
\end{equation}
and generates the units in order, each sampled unit joining the context for later predictions.
The units may be discrete codes or continuous latent blocks, ordered by time alone or by a space--time serialization, and Transformer decoders are the common implementation~\cite{vaswani2017attention}.
Masked modeling instead hides a subset of units and predicts them in parallel from the visible context and the condition.
For discrete targets, with sampled masked positions $M$ and visible units $z_{\bar M}$, a common objective is the token cross-entropy
\begin{equation}
  \mathcal{L}_{\mathrm{mask}}
  =\mathbb{E}_{(z,c),M}\!\left[-\sum_{i\in M}\log p_\theta(z_i\mid z_{\bar M},c)\right].
\end{equation}
MaskGIT~\cite{chang2022maskgit} turns these predictions into samples through iterative filling and refinement.
Continuous pixel or feature targets instead use a reconstruction loss such as squared error~\cite{he2022masked}.

\leadlabel{Diffusion and flow matching.}
A diffusion model learns to reverse a forward process that gradually corrupts data with noise~\cite{sohl2015deep,ho2020denoising}.
Let $(z_0,c)\sim p_{\mathrm{data}}$ be a clean training latent and its condition, and let $k\in\{1,\dots,K\}$ index the denoising step; the subscript $k$ is diffusion time, distinct from video time $t$.
In the noise-prediction form, $z_0$ is corrupted to $z_k=\sqrt{\bar\alpha_k}\,z_0+\sqrt{1-\bar\alpha_k}\,\epsilon$, where $\epsilon\sim\mathcal{N}(0,\mathbf{I})$ and $\bar\alpha_k$ decreases toward $0$ as $k$ increases, and the network learns to recover the noise:
\begin{equation}
  \mathcal{L}_{\mathrm{diff}}
  =\mathbb{E}_{(z_0,c),\epsilon,k}\!\left[\big\lVert\epsilon_\theta(z_k,k,c)-\epsilon\big\rVert^2\right].
\end{equation}
Sampling starts from $z_K\sim\mathcal{N}(0,\mathbf{I})$ and applies denoising updates $z_K\to\dots\to z_k\to z_{k-1}\to\dots\to z_0$, after which $z_0$ is decoded into video; networks may equivalently predict the clean sample or the velocity.
Flow matching instead trains a vector field that transports a simple base distribution to the data along a chosen probability path~\cite{lipman2023flow}.
Here $k$ is continuous: for the linear path $z_k=(1-k)z_0+k\epsilon$ with $k\sim\mathcal{U}[0,1]$, the target velocity is $\epsilon-z_0$:
\begin{equation}
  \mathcal{L}_{\mathrm{fm}}
  =\mathbb{E}_{(z_0,c),\epsilon,k}\!\left[\big\lVert v_\theta(z_k,k,c)-(\epsilon-z_0)\big\rVert^2\right].
\end{equation}
Sampling integrates $dz_k/dk=v_\theta(z_k,k,c)$ from noise at $k=1$ to data at $k=0$, in practice over a discrete grid of steps.
Diffusion and flow matching can share backbones and samplers.
Section~\ref{sec:reading} uses masked prediction to learn representations, Section~\ref{sec:writing} covers generators built on these objectives, and Section~\ref{sec:interacting} adapts them to action-conditioned online rollout.

\subsection{Taxonomy of Video World Models}
\label{sec:wm-taxonomy}
A \emph{video world model} performs the interacting operation of Definition~\ref{def:operations} with a video-centered state.
Unlike a general video generator, which aims at plausible content, it learns how that state evolves.
We exclude models whose primary persistent state is an explicit 3D structure, such as a neural field, 3D Gaussians, or an occupancy grid, as well as purely symbolic or text-based world models; models that keep 3D structure only as auxiliary memory for a video-centered state remain in scope (Section~\ref{sec:wm-memory}).
World models from model-based reinforcement learning are included only when their state is learned from video.

We classify video world models along two independent axes.
On the first axis, the \emph{prediction space}, \emph{observation-space} models predict future frames, video tokens, or generative latents decoded to video, as GAIA~\cite{hu2023gaia1,russell2025gaia2} and EVA~\cite{chi2024eva} do.
Although VAE latents are not pixels, predicting them for decoding to video still counts as observation-space prediction.
\emph{Latent-state} models instead predict future video-derived features that need not be decoded to video: V-JEPA~\cite{bardes2024vjepa} and V-JEPA~2~\cite{assran2025vjepa2} learn such features by masked feature prediction, which serves as pretraining for latent transition models (Section~\ref{sec:interacting}), and DINO-WM~\cite{zhou2024dinowm} and V-JEPA~2-AC add action-conditioned transitions for planning.
The second axis, the \emph{transition interface}, describes how controls reach the model.
\emph{Passive} models predict from observations alone, while \emph{action-conditioned} models receive a control sequence fixed before rollout, such as the ego trajectories of Vista~\cite{gao2024vista} or the robot actions of IRASim~\cite{zhu2025irasim}.
\emph{Interactive} models accept new controls during rollout, so each input can change what follows, as in GameNGen~\cite{valevski2025gamengen}, Genie~2~\cite{genie2_2024}, and related playable simulators~\cite{alonso2024diamond,bruce2024genie,zhang2025matrixgame,he2025matrixgame2}.
Section~\ref{sec:interacting} reviews each cell of the resulting $2\times3$ grid (Figure~\ref{fig:worldmodel-taxonomy}).

\surveytheme{surveyRead}
\section{\readicon\opsep Video Comprehension: Reading the World}
\label{sec:reading}
\chapterlead{Reading means inferring world content, structure, dynamics, physics, and causality from partial video traces.}
Video comprehension performs the reading operation $V\rightarrow W$ of Section~\ref{sec:lens}: given a trace $V$, the model estimates properties of the latent world $W$ that produced it.
From a limited viewpoint and interval, it must recover which entities and events are present, where they are, how they change, which constraints govern that change, and what may have caused it.
Occlusion, camera motion, editing, sparse sampling, and missing modalities leave much of this state unobserved, and the viewing scenarios of Section~\ref{sec:video-scenarios} change how much evidence must be selected and remembered.
Sections~\ref{sec:reading-tasks}--\ref{sec:reading-apps} cover tasks and benchmarks, encoders, Video-LLMs, agents, and applications in turn.
Tang et al.~\cite{tang2023videollmsurvey} survey Video-LLMs in more detail; here we organize the evaluation of reading by the world properties it tests (Section~\ref{sec:reading-tasks}) and review Video-LLMs as one stage of the reading pipeline.

\subsection{Tasks, Data, and Evaluation}
\label{sec:reading-tasks}
We organize video-understanding tasks by the main dimension of world state they require: semantic (entities, actions, events), spatial (location, identity, layout, viewpoint), temporal (order, duration, boundaries, long-range dependencies), physical (contact, gravity, containment, material behavior), and causal (effects, interventions, counterfactuals).
The dimensions interact, as when tracking combines identity with temporal continuity, so Table~\ref{tab:world-state-benchmarks} places each benchmark under the dimension it mainly tests, even when the text introduces it elsewhere.
Output format is a separate axis: a label, span, mask, or sentence can probe any dimension, and what a benchmark measures depends on the evidence needed to produce its output.
\begin{table}[!t]
\centering
\small
\caption{Video-understanding benchmarks organized by the primary world-state property tested. A benchmark may probe more than one dimension.}
\label{tab:world-state-benchmarks}
\surveytablecolors
\setlength{\tabcolsep}{3.5pt}
\renewcommand{\arraystretch}{1.13}
\begin{tabularx}{\textwidth}{@{}p{1.55cm} >{\hsize=.72\hsize}Y >{\hsize=1.28\hsize}Y@{}}
\toprule
\surveytablehead
\tableheadcell{Dimension} & \tableheadcell{State property / readout} & \tableheadcell{Benchmarks} \\
\midrule
\surveybadgecolor{surveyRead}{Semantic} & Actions/events; clip--text matching & Kinetics~\cite{kay2017kinetics}, MSR-VTT~\cite{Xu_2016_CVPR}, VATEX~\cite{wang2019vatex}, ActivityNet-QA~\cite{yu2019activitynet} \\
\addlinespace[0.35ex]
\surveybadgecolor{surveyRead}{Semantic} & VideoQA with temporal and knowledge demands & Video-MME series~\cite{fu2025video,videommev2_2026}, MMBench-Video~\cite{fang2024mmbench}, MMVU~\cite{mmvu2025}, Video-MMMU~\cite{hu2025video} \\
\addlinespace[0.35ex]
\surveybadgecolor{surveyRead}{Semantic} & Evidence support; hallucination/omission & VideoHallucer~\cite{wang2024videohallucer}, EventHallusion~\cite{zhang2024eventhallusion}, ARGUS~\cite{rawal2025argus} \\
\addlinespace
\surveybadgecolor{surveyShare}{Spatial} & Masks/tubes and identity continuity & DAVIS~\cite{davis2017}, YouTube-VOS~\cite{youtubevos2018}, Ref-YouTube-VOS~\cite{seo2020urvos}, MeViS~\cite{mevis2023}, MOTChallenge~\cite{motchallenge2016}, LaSOT~\cite{lasot2018}, VidSTG~\cite{vidstg2020} \\
\addlinespace[0.35ex]
\surveybadgecolor{surveyShare}{Spatial} & Geometry, viewpoint, and observer/object motion & VSI-Bench~\cite{vsibench2024}, MMSI-Video-Bench~\cite{mmsivideobench2026}, STI-Bench~\cite{li2025stibench}, DSI-Bench~\cite{zhang2025dsibench}; NuScenes-SpatialQA~\cite{tian2025nuscenesspatialqa} \\
\addlinespace[0.35ex]
\surveybadgecolor{surveyShare}{Spatial} & Out-of-view states and retained spatial context & SCP-Bench~\cite{zhao2026scp}, Beyond3D~\cite{ma2026beyond3d}, OVO-S-Bench~\cite{li2026ovosbench} \\
\addlinespace
\surveybadgecolor{surveyWrite}{Temporal} & Event boundaries and query intervals & THUMOS~\cite{idrees2017thumos}, Charades-STA~\cite{charadessta2017}, QVHighlights~\cite{lei2021qvhighlights} \\
\addlinespace[0.35ex]
\surveybadgecolor{surveyWrite}{Temporal} & Order, direction, speed, repetition, and state change & Something-Something~\cite{goyal2017something}, TGIF-QA~\cite{jang2017tgif}, TemporalBench~\cite{temporalbench2024}, TempCompass~\cite{liu2024tempcompass}, TimeBlind~\cite{li2026timeblind}, Video-MME-Logical~\cite{kwan2026videommelogical} \\
\addlinespace[0.35ex]
\surveybadgecolor{surveyWrite}{Temporal} & Sparse evidence and frame selection & EgoSchema~\cite{mangalam2023egoschema}, MLVU~\cite{zhou2025mlvu}, LVBench~\cite{wang2025lvbench}, LongVideoBench~\cite{wu2024longvideobench}, VideoOdyssey~\cite{videoodyssey2026}; TempCore~\cite{ok2025tempcore} \\
\addlinespace[0.35ex]
\surveybadgecolor{surveyWrite}{Temporal} & Streaming prefixes, response timing, bounded history & SVCBench~\cite{liu2026svcbench}, TRACE~\cite{ma2026trace}, StreamingEval~\cite{tang2026streamingeval} \\
\addlinespace
\surveybadgecolor{surveyInteract}{Physical} & Persistence, plausibility, and future contact & IntPhys~\cite{intphys2018}, IntPhys~2~\cite{intphys2_2025}, Physion~\cite{physion2021}, ImplausiBench~\cite{motamed2025travl} \\
\addlinespace[0.35ex]
\surveybadgecolor{surveyInteract}{Physical} & Properties, interaction outcomes, and action success & CLEVRER~\cite{clevrer2019}, PhysicsMind~\cite{physicsmind2025}, PhysBench~\cite{chow2025physbench}, PAI-Bench-U~\cite{zhou2025paibench}, MVP~\cite{krojer2025mvp} \\
\addlinespace[0.35ex]
\surveybadgecolor{surveyInteract}{Physical} & Mechanism-linked transitions and next-state order & APT~\cite{wu2026apt}, VisPhyWorld~\cite{liang2026visphyworld} \\
\addlinespace[0.35ex]
\surveybadgecolor{surveyInteract}{Physical} & Physical descriptions and quantitative properties & World Embedding Benchmark~\cite{liu2026worldembedding} \\
\addlinespace
\surveybadgecolor{surveyFuture}{Causal} & Action--effect relations and dispersed evidence & NExT-QA~\cite{xiao2021next}, STAR~\cite{star2024}, Video-Holmes~\cite{videoholmes2025}, Causal2Needles~\cite{causal2needles2025}, CaST-Bench~\cite{zhang2026castbench} \\
\addlinespace[0.35ex]
\surveybadgecolor{surveyFuture}{Causal} & Counterfactual answers and changed conditions & CLEVRER~\cite{clevrer2019}, CausalVQA~\cite{foss2025causalvqa}, DMC-CF~\cite{zhang2026dmccf} \\
\bottomrule
\end{tabularx}
\end{table}

Aggregate accuracy does not show that a model used the intended evidence, since many questions can be answered from the text, a single frame, or static appearance.
Text-only, single-frame, shuffled-order, and restricted-context baselines expose these shortcuts, and TempCore~\cite{ok2025tempcore} and Video-Index~\cite{song2026videoindex} use them to filter existing benchmarks down to vision-dependent, order-sensitive subsets.
Minimal pairs of near-identical videos with different answers block appearance-only solutions (TimeBlind~\cite{li2026timeblind}, MVP~\cite{krojer2025mvp}).

\subsubsection{Semantic Understanding}
Semantic understanding identifies the entities, actions, events, and relations that make a video meaningful.
Action recognition is the most established semantic task: UCF101~\cite{soomro2012ucf101} and HMDB51~\cite{kuehne2011hmdb} test closed-set recognition in trimmed clips, Kinetics~\cite{kay2017kinetics} scales up the number and diversity of classes, and Something-Something~\cite{goyal2017something} makes labels depend on how objects change under manipulation.
Language moves the output beyond fixed categories through retrieval (MSR-VTT~\cite{Xu_2016_CVPR}), captioning (VATEX~\cite{wang2019vatex}, YouCook2~\cite{zhou2018youcook2}), and video question answering (VideoQA: ActivityNet-QA~\cite{yu2019activitynet}, TVQA~\cite{lei-etal-2018-tvqa}, TGIF-QA~\cite{jang2017tgif}).

Recent Video-LLM benchmarks mix semantic questions with temporal, spatial, and knowledge-intensive items (Video-MME and Video-MME-v2~\cite{fu2025video,videommev2_2026}, MMVU~\cite{mmvu2025}, Video-MMMU~\cite{hu2025video,luo2026unveiling}).
Hallucination benchmarks test whether the reported content is supported by the video at all: VideoHallucer~\cite{wang2024videohallucer,luo2026dr} uses adversarially paired questions, EventHallusion~\cite{zhang2024eventhallusion} probes events suggested by language priors, and ARGUS~\cite{rawal2025argus} scores hallucination and omission in free-form captions.

\subsubsection{Spatial Understanding}
Spatial understanding asks where entities are, how their extent changes, and whether their identity persists as viewpoint and visibility vary.
Spatiotemporal grounding links a referring expression to a region tube over an interval (VidSTG~\cite{vidstg2020}).
Video object segmentation tracks pixel masks (DAVIS~\cite{davis2017}, YouTube-VOS~\cite{youtubevos2018}), referring variants select the target by language or motion (Ref-YouTube-VOS~\cite{seo2020urvos}, MeViS~\cite{mevis2023}; addressed by ReferFormer~\cite{wu2022referformer}), and MOTChallenge~\cite{motchallenge2016} and LaSOT~\cite{lasot2018} test identity across occlusion and reappearance.
Memory-based and promptable trackers such as XMem~\cite{cheng2022xmem}, AOT~\cite{yang2021aot}, DEVA~\cite{cheng2023deva}, and SAM~2~\cite{ravi2024sam2} keep earlier evidence as the target changes appearance or disappears.

Spatial reasoning benchmarks go beyond masks and tracks: VSI-Bench~\cite{vsibench2024} and MMSI-Video-Bench~\cite{mmsivideobench2026} test relative position, viewpoint, and layout, STI-Bench~\cite{li2025stibench} asks for quantitative pose and motion, and DSI-Bench~\cite{zhang2025dsibench} separates observer motion from object motion.
A newer group tests state that is out of view: SCP-Bench~\cite{zhao2026scp} infers unseen past or future layouts, Beyond3D~\cite{ma2026beyond3d} asks where relocated objects now are, and OVO-S-Bench~\cite{li2026ovosbench} answers from the video prefix seen so far.
These tasks test the object permanence and revisit consistency that world models must maintain (Section~\ref{sec:wm-memory}).

\subsubsection{Temporal Understanding}
Temporal understanding recovers when an event occurs, how long it lasts, and how evidence at different moments is ordered or connected.
Temporal action detection locates predefined actions in untrimmed video (THUMOS~\cite{idrees2017thumos}, ActivityNet~\cite{caba2015activitynet}), and moment retrieval returns the segment that matches a language query (Charades-STA~\cite{charadessta2017}, QVHighlights~\cite{lei2021qvhighlights}, DiDeMo~\cite{anne2017localizing}).
Dense event captioning localizes and describes possibly overlapping events (ActivityNet Captions~\cite{Krishna_2017_ICCV}, addressed end to end by Vid2Seq~\cite{yang2023vid2seq}), and anticipation predicts an action not yet observed (e.g., Ego4D forecasting~\cite{grauman2022ego4d}).

Fine-grained and long-form benchmarks stress subtle temporal differences and sparse evidence.
TemporalBench~\cite{temporalbench2024} and TempCompass~\cite{liu2024tempcompass} test ordering, speed, and direction, TimeBlind~\cite{li2026timeblind} pairs videos that differ only in temporal structure, and Video-MME-Logical~\cite{kwan2026videommelogical} separates logical complexity from scene recognition.
MLVU~\cite{zhou2025mlvu}, LVBench~\cite{wang2025lvbench}, and LongVideoBench~\cite{wu2024longvideobench} pose heterogeneous questions over long videos, and VideoOdyssey~\cite{videoodyssey2026} extends this to ultra-long, omni-modal video, measuring how much context stays usable as evidence length grows.

Streaming benchmarks make state maintenance directly observable.
SVCBench~\cite{liu2026svcbench} asks for running object and event counts at many points in a video, TRACE~\cite{ma2026trace} scores answer quality together with latency, false alarms, and missed responses under causal input, and StreamingEval~\cite{tang2026streamingeval} adds encoding throughput, time to first token, and a fixed memory budget.

\subsubsection{Physical Understanding}
Physical understanding concerns how a visible state can change under constraints that are often only partly observed.
Plausibility benchmarks ask whether an observed transition is possible: IntPhys~\cite{intphys2018} and IntPhys~2~\cite{intphys2_2025} test object permanence and intuitive physics, ImplausiBench~\cite{motamed2025travl} contains videos that violate expected physical behavior, and Physion~\cite{physion2021} instead asks whether objects will make contact.
Other benchmarks ask for next states, interaction outcomes, or the mechanism behind a change.
CLEVRER~\cite{clevrer2019} asks descriptive, predictive, and explanatory questions about collisions, while PhysicsMind~\cite{physicsmind2025}, PhysBench~\cite{chow2025physbench}, and PAI-Bench-U~\cite{zhou2025paibench} broaden coverage to object properties, commonsense constraints, task completion, and next actions.
APT~\cite{wu2026apt} further labels timestamped transitions by mechanism (contact, gravity, or friction).

Two benchmarks check that answers rest on physical evidence rather than priors.
MVP~\cite{krojer2025mvp} pairs near-identical videos that require opposite answers, and VisPhyWorld~\cite{liang2026visphyworld} asks the model to rebuild the scene as executable simulator code so that its physical hypothesis can be checked.
Physics benchmarks for generated video, such as Physics-IQ~\cite{motamed2025physicsiq} and VideoPhy~\cite{bansal2024videophy,bansal2025videophy2}, are covered in Sections~\ref{sec:writing-eval} and~\ref{sec:wm-benchmarks}.

\subsubsection{Causal Understanding}
Causal understanding asks why an observed event occurred and how the outcome would change if an action, object, or condition were different.
On synthetic collisions, CLEVRER~\cite{clevrer2019} adds counterfactual questions to its explanatory ones.
Real-world and procedural benchmarks link actions to effects that may appear much later: NExT-QA~\cite{xiao2021next} and STAR~\cite{star2024} ask about event relations and reasons, Video-Holmes~\cite{videoholmes2025} and Causal2Needles~\cite{causal2needles2025} spread the clues far apart, and CaST-Bench~\cite{zhang2026castbench} covers more varied settings and causal structures.
Counterfactual benchmarks test altered outcomes directly: CausalVQA~\cite{foss2025causalvqa} asks counterfactual, hypothetical, and planning questions about real egocentric video, and DMC-CF~\cite{zhang2026dmccf} pairs such questions about real videos with causal-graph interventions, whose regenerated variants also limit data contamination.

\subsection{Video Representation Learning and Encoding}
\label{sec:reading-representation}
\begin{figure}[t]
  \centering
  \includegraphics[width=0.99\textwidth]{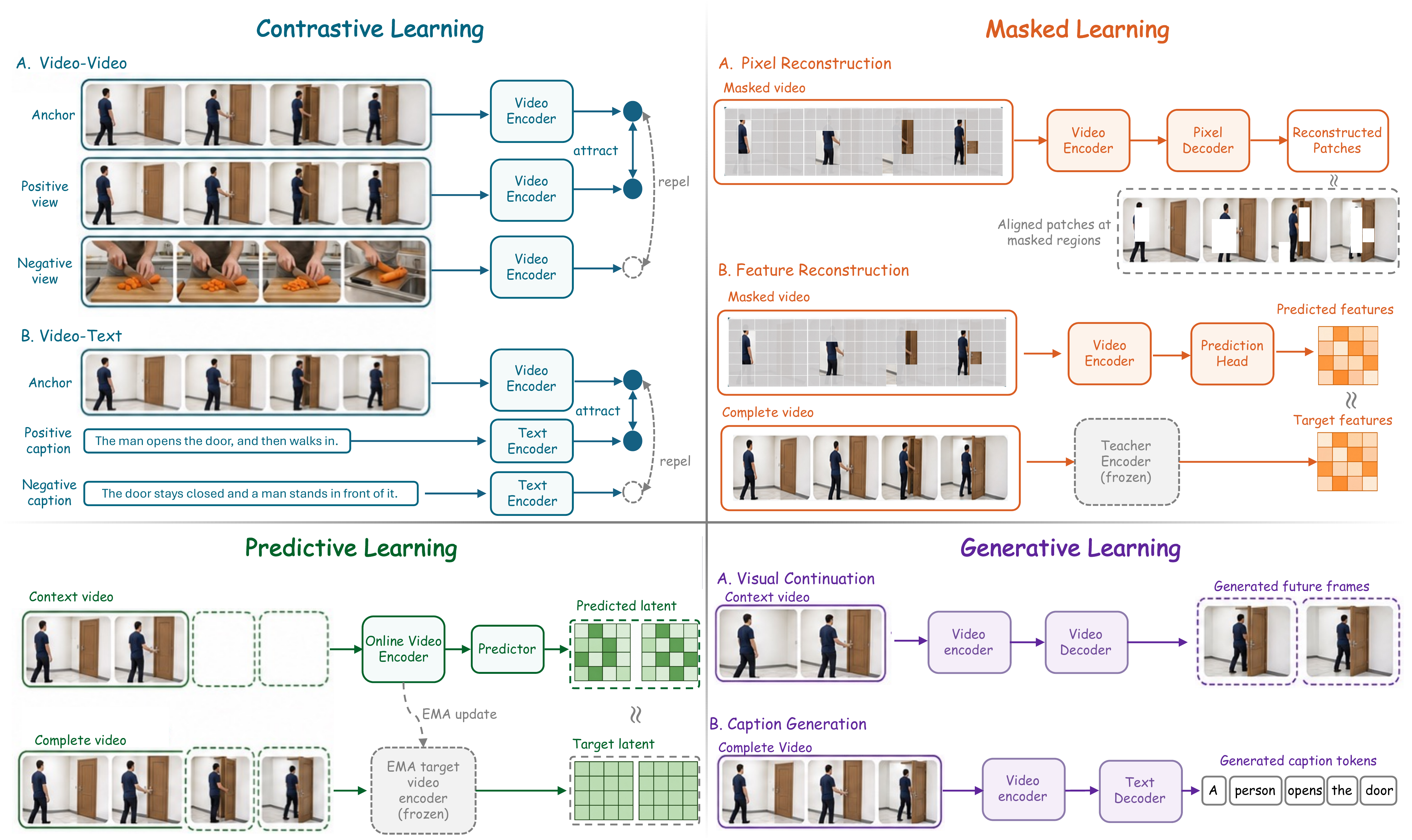}
  \caption{Four families of video representation learning.
  They differ in how supervision targets are constructed and how predictions are compared with or generated from those targets.}
  \label{fig:representation-learning}
\end{figure}

Video representation learning maps frames to features that can be reused across recognition, retrieval, grounding, reasoning, and multimodal modeling.
Convolutional backbones such as I3D~\cite{carreira2017quo} and SlowFast~\cite{feichtenhofer2019slowfast} established supervised spatiotemporal processing, and TimeSformer~\cite{bertasius21timesformer}, ViViT~\cite{arnab2021vivit}, and Video Swin~\cite{liu2022videoswin} carried the Vision Transformer~\cite{dosovitskiyimage} to video at different spatial and temporal scales.
The pretraining objective decides what these backbones retain, and we group objectives by their dominant supervision signal into the four families of Figure~\ref{fig:representation-learning}.
The families overlap: masked feature prediction, for instance, is both masked and predictive.
Variational autoencoders and tokenizers built for synthesis are covered in Section~\ref{sec:writing}.

\subsubsection{Contrastive Learning}
Contrastive learning pulls positive pairs together in embedding space and pushes mismatched pairs apart, commonly with an InfoNCE-style objective.
In video-only learning, CVRL~\cite{qian2021cvrl} and $\rho$MoCo~\cite{feichtenhofer2021large} draw positives from augmentations or clips of the same video, and the sampling distance sets a trade-off: views too close give little contrast, while views too far apart may show different events.

Video--text contrastive learning pairs clips with language.
Among dual-encoder retrieval models, Frozen in Time~\cite{Bain_2021_ICCV} trains a space--time Transformer jointly on image--text and video--text pairs, and CLIP4Clip~\cite{luo2022clip4clip} transfers the CLIP image--text encoder and adds temporal aggregation, while VideoCLIP~\cite{xu2021videoclip}, X-CLIP~\cite{ma2022xclip}, ViCLIP~\cite{wang2023internvid}, and Finsta~\cite{fei2024finsta} develop video-specific alignment.
Foundation encoders combine contrastive pretraining with masked or generative losses (InternVideo~\cite{wang2022internvideo}, InternVideo2~\cite{wang2024internvideo2}), distillation into a masked-video student (VideoPrism~\cite{zhao2024videoprism}), or later alignment stages for language and dense prediction (Perception Encoder~\cite{bolya2025pe}).
Global alignment can drop fine-grained detail: the World Embedding Benchmark~\cite{liu2026worldembedding} shows that alignment and recoverable physical state can diverge, as physics-specific contrastive tuning improves video--text matching but weakens physical-quantity regression from frozen features; the added losses aim to retain it.

\subsubsection{Masked Learning}
Masked learning removes much of the spatiotemporal input and trains the model to recover it, an idea that VideoBERT~\cite{sun2019videobert} applied early to quantized video tokens and speech transcripts jointly.
VideoMAE~\cite{tong2022videomae} reconstructs pixels behind tube masks, with a high mask ratio that exploits the redundancy of adjacent frames, and VideoMAE-V2~\cite{wang2023videomae} scales it with dual masking.
Feature targets replace raw pixels: MaskFeat~\cite{wei2022maskfeat} reconstructs handcrafted or learned features, and UMT~\cite{li2023unmasked} distills from an unmasked teacher that sees the full clip.
Learned or teacher targets shift capacity from low-level appearance toward the semantics and motion that the teacher encodes, but they also inherit its biases.

\subsubsection{Predictive Learning}
Predictive learning trains a model to predict the latent representation of hidden or future content, without pixel reconstruction or negative examples.
V-JEPA~\cite{bardes2024vjepa} predicts the features of masked regions against a slowly updated target encoder, V-JEPA 2~\cite{assran2025vjepa2} scales this to larger and more diverse data, and V-JEPA 2.1~\cite{muckley2026vjepa21} improves dense features.
InternVideo-Next~\cite{wang2025internvideonext} bridges the masked and predictive families, combining masked reconstruction with latent prediction of its first-stage targets, without video--text data.
Adding an action-conditioned transition model to such a representation yields the action-conditioned latent-state world models of Section~\ref{sec:interacting}.

\subsubsection{Generative Learning}
Generative pretraining shapes a representation by requiring it to produce missing or future content.
Visual autoregressive objectives~\cite{yan2021videogpt} predict future frames, patches, or tokens, which rewards sensitivity to appearance and motion but can spend capacity on reconstruction detail that recognition does not need.
Vision-language generation outputs text instead: GIT-style~\cite{wang2022git} training generates captions or answers from visual features, and VideoCoCa~\cite{Yan2023VideoCoCa} combines captioning with contrastive alignment.
In both cases, the representation is shaped only by what captions describe.
Large video diffusion models offer a third route: Gen4U~\cite{gen4u2026} reuses a frozen generator's intermediate activations for classification, geometry, and captioning, and GenCeption~\cite{genception2026} adapts a diffusion backbone into a text-steered model for depth, normals, pose, segmentation, and keypoints (Section~\ref{sec:sharing-learning}).

\subsection{Video-Language Foundation Modeling}
\label{sec:reading-videollm}
Video-LLMs extend language foundation models to video for open-ended description, dialogue, question answering, grounding, and reasoning.
They take visual features from encoders such as those in Section~\ref{sec:reading-representation} and must decide which frames to observe, how to encode time, and how many visual tokens the language model can afford.
We review their architectures, training, and test-time scaling in turn.

\subsubsection{Architectures}
\leadlabel{Visual front end.}
Many systems apply an image encoder such as CLIP~\cite{radford2021learning} or SigLIP~\cite{zhai2023sigmoid,tschannen2025siglip} to sampled frames, a route that is cheap and semantically strong but leaves temporal modeling to later modules.
Native-video encoders~\cite{assran2025vjepa2,zhu2026efficient} model local motion before fusion, and direct patch or visual-token inputs keep finer spatial detail at a higher context cost.
No encoder can recover an event from an unsampled frame.

\leadlabel{Time and position.}
Frame order may be encoded by temporal embeddings, timestamps, or multidimensional rotary positions.
Qwen2.5-VL~\cite{qwen25vl2025} uses M-RoPE to coordinate temporal and spatial indices, Qwen3-VL~\cite{bai2025qwen3} adds textual timestamps so that answers can cite moments, and the InternVL series~\cite{wang2025internvl3} advances visual positions in fractional steps so that long sequences stay within the trained positional range.
Temporal Transformers and state-space models such as Mamba~\cite{gu2023mamba} can also aggregate frame sequences before they reach the LLM.

\leadlabel{Vision-language connector.}
The connector maps visual features into the language backbone's embedding space and token budget.
Linear or MLP projectors turn each visual feature into a prefix token, as in LLaVA-style systems~\cite{liu2023visual}, while Q-Former and resampler designs~\cite{pmlr-v202-li23q} extract a fixed number of query tokens, as in early Video-LLaMA~\cite{zhang-etal-2023-video} and VideoChat~\cite{videochat2023}.
Gated cross-attention, as in Flamingo~\cite{alayrac2022flamingo}, instead lets language layers consult visual features at selected depths.
Structure-preserving compressors downsample while keeping spatial layout, through convolutional abstractors (Honeybee~\cite{cha2023honeybee}), spatial-temporal convolution (VideoLLaMA 2~\cite{cheng2024videollama2}), or pixel unshuffle (InternVL~\cite{wang2025internvl3}).
Projectors keep the most direct token path, resamplers and cross-attention give the most control over how much visual information enters the LLM, and compressors trade some directness for locality-aware reduction.

\leadlabel{Compression and memory.}
Video token count grows with both the number of frames and the tokens per frame, and pooling, merging, or resampling reduce it at the risk of erasing small objects, short events, or precise timestamps.
Some designs spend tokens unevenly over time: LLaVA-Video~\cite{zhang2024llavavideo} keeps a few frames dense and pools the rest, Keye-VL-1.5~\cite{yang2025keyevl15} routes frames with large visual change to a high-resolution pathway, and LLaVA-OneVision-2~\cite{an2026llavaonevision2} groups tokens adaptively like a video codec.
Others compress each unit: Chat-UniVi~\cite{jin2023chatunivi} merges tokens into dynamic multi-scale units, LLaMA-VID~\cite{li2024llamavid} reduces each frame to as few as two tokens, and Video-LaVIT~\cite{jin2024videolavit} pairs keyframes with compact motion vectors.
For long video, LongVA~\cite{longva2024} and LongVILA~\cite{longvila2024} rely on long-context transfer and token budgeting, and Video-XL~\cite{videoxl2024} adds structured compression.
One finding bounds this design space: a few tokens carry most of the discriminative information, but finding them without an oracle is hard~\cite{hao2024visualtokens}.
Memory solves a different problem from compression: MovieChat~\cite{song2024moviechat} and MA-LMM~\cite{malmm2024} maintain a memory of past segments, which decides what remains available after the original frames leave the context (agent memory is treated in Section~\ref{sec:reading-agentic}).

\leadlabel{Language backbone and output.}
An instruction-following LLM combines the retained evidence with the query and produces dialogue, descriptions, explanations, or structured predictions.
Early systems extended image-language models to video with adapters and compressed frame tokens~\cite{zhang-etal-2023-video,videochat2023,maaz-etal-2024-video,lin2024videollava,li2024llamavid}, and later open models add larger multimodal mixtures, more efficient video tokens, and grounded or reasoning outputs~\cite{li2024llava,qwen25vl2025,bai2025qwen3,wang2025internvl3,yan2026internvideo3,yao2024minicpm,videollama3_2025,videochat3_2026}.
Boxes, timestamps, and masks can be emitted as special tokens or by attached heads, which makes the evidence behind an answer easier to inspect.
Closed systems such as Gemini~\cite{gemini15_2024} show strong long-context video and audio results but disclose few design details.
Table~\ref{tab:reading-architectures} lists the encoder, connector, and LLM of representative open systems and what each training stage updates.

\leadlabel{Streaming and audio-visual input.}
Streaming models must answer from the evidence seen so far and bound their state with windows, recurrent memory, or key--value (KV) caches.
VideoLLM-online~\cite{chen2024videollmonline} introduced real-time dialogue over a continuously advancing stream, StreamingVLM~\cite{xu2025streamingvlm} keeps a compact KV cache for unbounded streams, and OneStreamer~\cite{zeng2026onestreamer} writes timestamped caption memory and learns when to record and when to respond.
For audio-video input, Qwen2.5-Omni~\cite{xu2025qwen25omni} and Qwen3-Omni~\cite{xu2025qwen3} place both modalities on one timeline with time-aligned multimodal rotary positions.
\begin{table}[!t]
\centering
\small
\caption{Video-LLM architectures and component training status. \trainedmark{} denotes trained and \frozenmark{} frozen parameters in the stage named beneath each model. Connector marks refer to learned modules, not parameter-free pooling or pruning.}
\label{tab:reading-architectures}
\surveytablecolors
\setlength{\tabcolsep}{3.5pt}
\begin{tabularx}{\textwidth}{@{}>{\raggedright\arraybackslash}p{3.35cm} >{\hsize=.90\hsize}Y >{\hsize=1.10\hsize}Y >{\raggedright\arraybackslash}p{2.70cm}@{}}
\toprule
\surveytablehead
\tableheadcell{Model / training stage} & \tableheadcell{Vision encoder} & \tableheadcell{Connector / compression} & \tableheadcell{LLM} \\
\midrule
\mbox{Video-ChatGPT}~\cite{maaz-etal-2024-video}\par {\footnotesize Instruction tuning} &
\trainingcell{\frozenmark}{CLIP ViT-L/14} &
\trainingcell{\trainedmark}{Spatiotemporal pooling; linear projector} &
\trainingcell{\frozenmark}{Vicuna-7B} \\
\addlinespace
Video-LLaVA~\cite{lin2024videollava}\par {\footnotesize Instruction tuning} &
\trainingcell{\frozenmark}{LanguageBind image/video encoders} &
\trainingcell{\trainedmark}{Shared two-layer MLP} &
\trainingcell{\trainedmark}{Vicuna-7B v1.5} \\
\addlinespace
LLaMA-VID~\cite{li2024llamavid}\par {\footnotesize Instruction tuning} &
\trainingcell{\frozenmark}{EVA-G} &
\trainingcell{\trainedmark}{Q-Former context query; context/content projection} &
\trainingcell{\trainedmark}{Vicuna-7B/13B} \\
\addlinespace
\mbox{LLaVA-OneVision}~\cite{li2024llava}\par {\footnotesize Stage 2: mixed modalities} &
\trainingcell{\trainedmark}{SigLIP} &
\trainingcell{\trainedmark}{Two-layer MLP; video pooling} &
\trainingcell{\trainedmark}{Qwen2} \\
\addlinespace
VideoLLaMA3~\cite{videollama3_2025}\par {\footnotesize Video-centric fine-tuning} &
\trainingcell{\trainedmark}{SigLIP, dynamic resolution} &
\trainingcell{\trainedmark}{Differential frame pruning; MLP} &
\trainingcell{\trainedmark}{Qwen2.5} \\
\addlinespace
Qwen2.5-VL~\cite{qwen25vl2025}\par {\footnotesize Supervised fine-tuning} &
\trainingcell{\frozenmark}{Window-attention ViT} &
\trainingcell{\trainedmark}{$2\!\times\!2$ merge; MLP} &
\trainingcell{\trainedmark}{Qwen2.5} \\
\addlinespace
InternVL3.5~\cite{wang2025internvl3}\par {\footnotesize Native pretraining} &
\trainingcell{\trainedmark}{InternViT-300M/6B} &
\trainingcell{\trainedmark}{MLP; pixel unshuffle} &
\trainingcell{\trainedmark}{Qwen3 or GPT-OSS} \\
\addlinespace
Qwen3-VL~\cite{bai2025qwen3}\par {\footnotesize S2: long-context pretraining} &
\trainingcell{\trainedmark}{SigLIP2-derived dynamic ViT} &
\trainingcell{\trainedmark}{Spatial MLP; DeepStack mergers} &
\trainingcell{\trainedmark}{Qwen3} \\
\addlinespace
Keye-VL-1.5~\cite{yang2025keyevl15}\par {\footnotesize Multi-task pretraining} &
\trainingcell{\trainedmark}{SigLIP-400M-384-14} &
\trainingcell{\trainedmark}{MLP; SlowFast and native dynamic resolution} &
\trainingcell{\trainedmark}{Qwen3-8B} \\
\addlinespace
VideoChat3~\cite{videochat3_2026}\par {\footnotesize Stage 2: instruction tuning} &
\trainingcell{\trainedmark}{MoonViT-initialized I3D-ViT} &
\trainingcell{\trainedmark}{Spatiotemporal pooling/merging; projector} &
\trainingcell{\trainedmark}{Qwen3-4B} \\
\addlinespace
OneStreamer~\cite{zeng2026onestreamer}\par {\footnotesize Supervised fine-tuning} &
\trainingcell{\frozenmark}{Qwen3-VL-4B vision encoder} &
\trainingcell{\trainedmark}{Inherited Qwen3-VL projector} &
\trainingcell{\trainedmark}{Qwen3-4B} \\
\bottomrule
\end{tabularx}
\end{table}

\subsubsection{Learning Paradigms}
Video-LLMs are usually trained in stages because the encoder, connector, and language backbone start from different representations.
In \emph{alignment pretraining}, the encoder and LLM are often frozen while the connector learns from caption pairs, video--text matching, and grounding data, following the visual instruction tuning recipe for images~\cite{liu2023visual}.
Table~\ref{tab:videollm-pretraining-data} lists pair corpora and instruction datasets for this stage and the next.
\begin{table}[!t]
\centering
\small
\caption{Datasets for Video-LLM pretraining and instruction tuning. Scale retains each source's counting unit; mixed-modality and image-only auxiliary resources are identified explicitly.}
\label{tab:videollm-pretraining-data}
\surveytablecolors
\setlength{\tabcolsep}{4.5pt}
\begin{tabularx}{\textwidth}{@{}l >{\hsize=.62\hsize}Y >{\hsize=1.38\hsize}Y@{}}
\toprule
\surveytablehead
\tableheadcell{Dataset} & \tableheadcell{Scale} &
\tableheadcell{Description} \\
\midrule
\surveygroup{3}{Web-scale video--text pairs (alignment pretraining)}
HowTo100M~\cite{Miech_2019_ICCV} &
136M clips / 1.2M videos &
instructional videos paired with automatically transcribed narration \\
\addlinespace
WebVid-10M~\cite{Bain_2021_ICCV} &
10.7M clips &
stock-footage clips paired with alt-text descriptions \\
\addlinespace
YT-Temporal-180M~\cite{zellers2021merlot} &
180M segments / 6M videos &
transcribed speech aligned with frames for temporal script knowledge \\
\addlinespace
HD-VILA-100M~\cite{xue2022hdvila} &
103M clips / 3.3M videos &
371.5K hours of 720p video spanning 15 YouTube categories \\
\addlinespace
InternVid~\cite{wang2023internvid} &
234M clips / 7.1M videos &
scalable model-generated captions with ViCLIP filtering; 760K hours \\
\addlinespace
Panda-70M~\cite{chen2024panda} &
70.8M clips &
captions selected from multiple cross-modality teacher models \\
\addlinespace
Action100M~\cite{chen2026action100m} &
147M segment annotations; 1.2M videos &
temporally localized action segments with hierarchical labels and captions from instructional video \\
\addlinespace
\surveygroup{3}{Instruction corpora and open training mixtures}
VideoInstruct-100K~\cite{maaz-etal-2024-video} &
100K QA pairs &
semi-automatic GPT-assisted video instructions from Video-ChatGPT \\
\addlinespace
VideoChat2-IT~\cite{li2024mvbench} &
$\sim$2M samples &
image/video instruction samples from 34 sources \\
\addlinespace
ShareGPT4Video~\cite{chen2024sharegpt4video} &
40K GPT-4V captions; 4.8M model captions &
dense, temporally detailed captions plus a learned video captioner \\
\addlinespace
LLaVA-Video-178K~\cite{zhang2024llavavideo} &
178K videos / 1.3M samples &
synthetic dense captions with open-ended and multiple-choice QA \\
\addlinespace
FineVision~\cite{wiedmann2025finevision} &
24.3M samples / 17.3M images &
image-based auxiliary instruction mixture; 200+ sources with deduplication and benchmark-overlap audits \\
\addlinespace
LLaVA-OneVision-2 data~\cite{an2026llavaonevision2} &
$\sim$8M video-caption pairs; 4M spatial samples &
re-captioned video pretraining pairs with a spatial-reasoning fine-tuning corpus \\
\addlinespace
\surveygroup{3}{Long-video, streaming, and reasoning corpora}
LongVid~\cite{li2025videochatflash} &
114K videos / 3.4M QA &
300K hours of long video with QA annotations across five task types \\
\addlinespace
Video-R1 corpora~\cite{feng2025videor1} &
260K RL samples; 165K CoT-SFT samples &
separate RL and supervised cold-start corpora spanning image and video reasoning \\
\addlinespace
Eagle-Video-110K~\cite{chen2025eagle25} &
110K videos &
story-level and clip-level annotations; diversity-driven video collection \\
\addlinespace
VideoMarathon~\cite{lin2025videomarathon} &
9.7K hours / 3.3M QA &
3--60 min videos with QA across 22 short- and long-horizon tasks \\
\addlinespace
Streamo-Instruct-465K~\cite{xia2025streamo} &
465K samples &
400K streaming samples plus offline video QA; narration, event grounding, and timed responses \\
\bottomrule
\end{tabularx}
\end{table}

\leadlabel{Supervised instruction tuning.}
Supervised fine-tuning (SFT) then trains on VideoQA and dialogue data for question-conditioned answers, temporal grounding data for timestamps~\cite{qian2024momentor}, and spatial grounding data for coordinates or other structured outputs.
Which components are updated varies: early Video-ChatGPT~\cite{maaz-etal-2024-video} tunes only its projector, Video-LLaVA~\cite{lin2024videollava} also updates the LLM, and VideoLLaMA3~\cite{videollama3_2025} and VideoChat3~\cite{videochat3_2026} adapt both the visual and language components (Table~\ref{tab:reading-architectures}).
The mixture of image, short-video, long-video, optical character recognition (OCR), and grounding data decides whether a model mainly describes clips or can locate and reason over the evidence behind an answer.

\leadlabel{Preference optimization and reinforcement learning.}
Post-training can then optimize preferences or task outcomes.
Reinforcement learning (RL) from human feedback~\cite{ouyang2022training} and direct preference optimization, which Qwen2.5-VL~\cite{qwen25vl2025} and VistaDPO~\cite{huang2025vistadpo} apply after SFT, improve visual faithfulness only insofar as the feedback separates grounded content from fluent but unsupported answers.
Reasoning-oriented methods instead use verifiable rewards, often with group relative policy optimization (GRPO)~\cite{guo2025deepseek}.
Video-R1~\cite{feng2025videor1} adds a temporal reward that compares rollouts on ordered and shuffled frames, LongVILA-R1~\cite{chen2025longvilar1} scales RL to long videos through its data and distributed training, and OneThinker~\cite{feng2025onethinker} broadens the task mixture and normalizes advantages with running per-task reward statistics.

Task-specific rewards target the output directly: Time-R1~\cite{wang2025timer1} rewards timestamp intersection-over-union for temporal grounding, VideoChat-R1~\cite{li2025videochatr1} jointly rewards spatiotemporal tasks without hurting general QA, and VideoP2R~\cite{jiang2025videop2r} rewards perception and reasoning separately.
Others modify GRPO itself: DeepVideo-R1~\cite{park2025deepvideor1} recasts it as advantage regression with difficulty-aware augmentation, and GRPO-CARE~\cite{chen2025grpocare} adds a bonus if a slowly updated reference model reproduces the answer from the reasoning trace.

\leadlabel{Data and scaling.}
Data and curriculum matter as much as model size.
Current recipes such as InternVideo3~\cite{yan2026internvideo3}, Qwen3-VL~\cite{bai2025qwen3}, and VideoLLaMA3~\cite{videollama3_2025} train jointly on image, video, and language data and widen task coverage with synthetic instructions; InternVideo3 and Qwen3-VL also add long-context curricula and use on-policy distillation to supply responses from states the model actually visits.
Apollo~\cite{zohar2025apollo} studies this design space systematically, measuring how frame sampling, encoder choice, and token allocation interact across model scales.
Newer corpora add finer temporal supervision: OneStreamer-1M~\cite{zeng2026onestreamer} times its captions and QA to the evidence available at each response point.

\subsubsection{Test-Time Scaling}
Test-time scaling spends extra computation after training to read the available video better.
Multimodal chain-of-thought~\cite{fei2024videoofthought,luo2024panosent} spends tokens on intermediate reasoning, which helps only if those steps stay grounded in the frames.
Query-aware frame selection instead spends compute on choosing the input: SeViLA~\cite{yu2023sevila} and VideoTree~\cite{wang2024videotree} rank, retain, or hierarchically organize frames for the query.
Iterative ``thinking with video'' methods such as FrameThinker~\cite{he2025framethinker} and VideoChat-A1~\cite{wang2025videochata1} alternate between reasoning and revisiting frames over several steps.
Search and aggregation select among several reasoning paths by self-consistency, best-of-$N$, or verifier and process-reward models, while V-Reason~\cite{sridhar2025vreason} directly modifies the model's inference-time behavior instead of sampling more outputs.
Each route trades latency and repeated visual computation for broader exploration or a second look at ambiguous evidence.

The boundary with agentic modeling is gradual: thinking-with-video methods already revisit frames, and once a model issues explicit operations to fetch new crops, frames, retrieval results, or persistent memory, it enters the agentic setting of Section~\ref{sec:reading-agentic}.

\subsection{Agentic Video Modeling}
\label{sec:reading-agentic}
Agentic video modeling turns evidence acquisition into a sequential decision problem: an LLM or Video-LLM forms subquestions, chooses a temporal region or visual detail to inspect, invokes an operation, and updates its reasoning from the result until it can answer.
Its actions change access to the recording by locating segments, sampling frames, cropping regions, querying tracks, or retrieving memory; actions that change the depicted world belong to the interactive rollouts of Section~\ref{sec:interacting}.
\begin{table}[!t]
\centering
\small
\caption{Agentic video systems. \textbf{Organization}: Single = one reasoning controller; Multi-role = coordinated specialist reasoning roles, which may share a backbone.}
\label{tab:video-agent-systems}
\surveytablecolors
\setlength{\tabcolsep}{3pt}
\begin{tabularx}{\textwidth}{@{}p{3.65cm} >{\hsize=1.05\hsize}Y >{\raggedright\arraybackslash}p{2.05cm} >{\hsize=.90\hsize}Y >{\hsize=1.05\hsize}Y@{}}
\toprule
\surveytablehead
\tableheadcell{Model} & \tableheadcell{Backbone} & \tableheadcell{Organization} & \tableheadcell{Memory / context} & \tableheadcell{Operations} \\
\midrule
\mbox{VideoAgent (long-form)}~\cite{wang2024videoagentlong} & GPT-4 controller; LaViLa or CogAgent captioner & Single & Selected-frame captions & Frame retrieval; captioning \\
\addlinespace
\mbox{VideoAgent (memory)}~\cite{fan2024videoagentmemory} & GPT-4 controller & Single & Event captions; object tracks & Caption retrieval; video QA; temporal localization; object queries \\
\addlinespace
VideoMind~\cite{liu2025videomind} & Qwen2-VL + role LoRA & Multi-role & Grounded intervals; role context & Grounding; verification; answering \\
\addlinespace
\mbox{VideoMultiAgents}~\cite{kugo2025videomultiagents} & GPT-4o roles; Gemini 2.0 Flash video tool & Multi-role & Role histories; scene graph & Caption analysis; scene-graph analysis; video QA \\
\addlinespace
Ego-R1~\cite{tian2025egor1} & Qwen2.5-3B controller & Single & Timestamped summaries & Hierarchical retrieval; video QA; image QA \\
\addlinespace
M3-Agent~\cite{long2025m3agent} & Qwen3 controller; Qwen2.5-Omni memory model & Single & Episodic + semantic memory & Memory retrieval; face/speaker identification \\
\addlinespace
VideoLucy~\cite{zuo2025videolucy} & DeepSeek-R1 reasoning; Qwen2.5-VL-7B captioner & Multi-role & Caption hierarchy & Temporal localization; recaptioning; backtracking \\
\addlinespace
LongVT~\cite{yang2026longvt} & Qwen2.5-VL-7B & Single & Current tool trace & Temporal crop; resampling \\
\addlinespace
\mbox{LongVideoAgent}~\cite{liu2025longvideoagent} & Qwen2.5-3B/7B master; Grok-4-fast-reasoning grounding; GPT-4o vision & Multi-role & Tags; accumulated observations & Grounding; visual inspection \\
\addlinespace
VideoThinker~\cite{li2026videothinker} & Qwen2.5-VL-7B & Single & Clip index; current trace & Temporal retrieval; zoom \\
\bottomrule
\end{tabularx}
\end{table}

\subsubsection{Video Agents}
Early video agents separate the controller's reasoning state from the full video stream (Table~\ref{tab:video-agent-systems}).
Two systems named VideoAgent take different routes: VideoAgent (long-form)~\cite{wang2024videoagentlong} has an LLM iteratively gather a small set of question-relevant frames, and VideoAgent (memory)~\cite{fan2024videoagentmemory} builds event and object-tracking memory and exposes segment localization and memory queries as tools.
VideoAgent2~\cite{zhi2025videoagent2} adds uncertainty-aware replanning, so that new observations can revise the search and intermediate evidence is verified.
Later controllers refine the observation policy: VCA~\cite{yang2024vca} explores unexamined segments out of curiosity, and VideoMind~\cite{liu2025videomind} splits the loop into planner, grounder, verifier, and answerer roles, all chain-of-LoRA experts on one backbone.

\emph{Agentic search} extends adaptive observation from frame selection to a structured video collection.
Deep Video Discovery~\cite{zhang2025deepvideodiscovery} segments long video into a multi-granular database and lets the controller choose search tools and parameters, and VideoExplorer~\cite{yuan2025videoexplorer} poses sub-questions inside the reasoning loop and locates the moments that answer them.

Continuous streams add another decision: when to observe and when to speak, possibly before the user asks.
StreamAgent~\cite{yang2025streamagent} anticipates task-relevant events and shifts attention before they arrive, and AViLA~\cite{zhang2025avila} answers asynchronous queries once the accumulated evidence is sufficient.

Memory determines what an agent retains between observations.
Flat caption lists are cheap to query but lose temporal scope and visual detail.
VideoLucy~\cite{zuo2025videolucy} instead backtracks through a coarse-to-fine memory hierarchy and MM-Mem~\cite{mmmem2026} builds a sensory, episodic, and schema-level memory pyramid, while InternVideo3~\cite{yan2026internvideo3} accumulates observations, reasoning, and tool actions in one evolving context.
Memory-centric agents aim for competence that grows with history instead of resetting at every query: M3-Agent~\cite{long2025m3agent} consolidates an entity-centric long-term memory from continuous audio-visual input, and WorldMM~\cite{yeo2025worldmm} maintains a dynamic multimodal memory.

\emph{Multi-agent collaboration} divides the loop across specialized roles.
VideoMultiAgents~\cite{kugo2025videomultiagents} coordinates vision, text, and graph specialists, LongVideoAgent~\cite{liu2025longvideoagent} pairs a reasoning master with grounding assistants, and Think, Then Verify~\cite{wang2026thinkverify} makes candidate answers survive adversarial checking.
MAPA~\cite{xu2026mapa} pairs lightweight perception agents with action agents to avoid redundant computation.

Other agents are trained, not only prompted: SAGE~\cite{jain2025sage} learns with RL how far ahead to look, EVA~\cite{zhang2026evaagent} makes end-to-end RL of a video agent efficient, and CASE~\cite{du2026case} learns when to stop searching.
Trained policies can lower tool cost and improve stopping, but they tie the agent to the tools seen in training.

Agent quality depends jointly on the backbone, observation policy, tool outputs, and memory: a wrong first hypothesis can steer later calls to the wrong interval, and noisy captions or detections propagate through the reasoning trace.
Agents are trained on tool-use traces for SFT and questions for RL (Ego-R1 Data~\cite{tian2025egor1}, LongVT's VideoSIAH~\cite{yang2026longvt}, VideoThinker's synthetic trajectories~\cite{li2026videothinker}).
Benchmarks (Table~\ref{tab:video-agent-evaluation}) test sparse-evidence search (VideoSIAH-Eval~\cite{yang2026longvt}), multi-hop retrieval with evidence-necessity checks and tool-call cost (LongVidSearch~\cite{yu2026longvidsearch}), trajectories against evidence milestones (AgentVidBench~\cite{an2026agentvidbench}), and combined video and web search (VideoGAIA~\cite{zhang2026videogaia}).
Others test proactive and spoken audio-visual interaction (IPIBench~\cite{ipibench2026}, OmniInteract~\cite{lu2026omniinteract}) and cross-session memory with cases where the needed history is missing (APM-Bench~\cite{huang2026apmbench}).
Beyond answer accuracy, evaluation should report whether the retrieved evidence supports the answer and what the search costs; REPAIR~\cite{alhamidi2026repair} adds repeated paired runs to separate run-to-run variance from method effects and counts the operations and latency hidden behind visible tool calls.
\begin{table}[!t]
\centering
\small
\caption{Video-agent benchmarks for evidence acquisition, interaction control, and persistent memory.}
\label{tab:video-agent-evaluation}
\surveytablecolors
\setlength{\tabcolsep}{4pt}
\begin{tabularx}{\textwidth}{@{}p{3.2cm} >{\hsize=.94\hsize}Y >{\hsize=1.04\hsize}Y >{\hsize=1.02\hsize}Y@{}}
\toprule
\surveytablehead
\tableheadcell{Benchmark} & \tableheadcell{Task / data} & \tableheadcell{Measures} & \tableheadcell{Protocol} \\
\midrule
\mbox{VideoSIAH-Eval}~\cite{yang2026longvt} &
Sparse-evidence QA; 652 questions &
Answer accuracy &
Sparse/dense sampling; tool calling \\
\addlinespace
LongVidSearch~\cite{yu2026longvidsearch} &
Multi-hop retrieval; 3,000 questions / 447 videos &
Answer accuracy; tool-call cost &
Fixed tools; 2--4 hops; gold-evidence oracle \\
\addlinespace
OmniInteract~\cite{lu2026omniinteract} &
Audiovisual streams; 1,430 response slots / 250 videos &
IA-QTF1\textsuperscript{a} &
Causal replay; interruptions; nested context \\
\addlinespace
IPIBench~\cite{ipibench2026} &
Proactive monitoring; interleaved queries &
Trigger timing; response correctness &
Add/edit/cancel tasks; past frames only \\
\addlinespace
VideoGAIA~\cite{zhang2026videogaia} &
Video--web tasks; 271 tasks &
pass@1, three-run mean &
ReAct; 40-step budget \\
\addlinespace
AgentVidBench~\cite{an2026agentvidbench} &
Multi-hop QA; 100 questions / 71 videos &
Answer accuracy; trajectory scores\textsuperscript{b} &
Evidence milestones; judged trajectories \\
\addlinespace
APM-Bench~\cite{huang2026apmbench} &
Long-term assistance; 549 sessions / 104 trajectories &
Recall; perception; proactive response; storage; latency &
Cross-session; missing-history cases \\
\bottomrule
\end{tabularx}
\par\vspace{3pt}\noindent\begin{minipage}{\textwidth}\footnotesize\raggedright
\textsuperscript{a}\,Interaction-Aware Quality--Timeliness F1; quality measures answer correctness and coverage.\par
\textsuperscript{b}\,Task understanding; evidence coverage, grounding, and completeness; reasoning faithfulness.
\end{minipage}
\end{table}

\subsubsection{Tool Use and Retrieval Augmentation}
Tools supply observations that the base model cannot obtain reliably from its current context: detectors and trackers for entity state, temporal grounders for event boundaries, croppers for spatial detail, and OCR, speech recognition, or captioners for other modalities.
In modular agents, a prompted controller composes external tools: VisProg~\cite{gupta2022visprog} and ViperGPT~\cite{suris2023vipergpt} generate executable programs over vision tools without training.
Native tool-calling models instead learn when to call a tool, with which parameters, and how to use the result.
LongVT~\cite{yang2026longvt} surveys a long video, predicts a relevant interval, and crops and densely resamples it before continuing to reason, and VideoThinker~\cite{li2026videothinker} learns temporal retrieval and zoom from tool trajectories synthesized in caption space and grounded back to video.
Ego-R1~\cite{tian2025egor1} trains a chain of tool-thought with RL over day- and week-long egocentric recordings.
Tool use can also be distilled away: Agent-of-Thoughts Distillation~\cite{shi2024aotd} distills agent traces into a single Video-LLM as chain-of-thought supervision, trading flexibility for one-pass inference.

Retrieval augmentation moves evidence into an external index.
One VideoRAG~\cite{jeong2025videoragcorpus} retrieves query-relevant videos and keeps both their visual and textual content, while another~\cite{ren2025videoragextreme} combines graph-based text grounding with multimodal encoding to search very large collections.
The retrieval unit shapes what can be recovered: fixed-length chunks may split an event, so SceneRAG~\cite{zeng2025scenerag} indexes narrative scenes and adds an entity-relation graph for multi-hop retrieval, at the cost of errors in segmentation and graph construction.
Tools and retrieval address different bottlenecks: a cropper or grounder fetches finer local evidence on demand, whereas an index narrows the search across long videos or corpora.
Both still require the system to decide what to request, detect missing or conflicting evidence, and trace the answer back to its source clips.

\subsection{Applications}
\label{sec:reading-apps}
Applications pose these problems under domain-specific viewpoints, vocabularies, and event structures.

\subsubsection{Embodied and Egocentric Perception}
Egocentric video records activity from the wearer's viewpoint, so the camera moves with the body, hands occlude manipulated objects, and key events are separated by long stretches of routine activity.
Standard resources include EPIC-KITCHENS-100~\cite{damen2022rescaling}, Ego4D~\cite{grauman2022ego4d}, EgoSchema~\cite{mangalam2023egoschema}, and the paired first- and third-person views of Ego-Exo4D~\cite{grauman2024egoexo4d} (Section~\ref{sec:video-scenarios}).
An egocentric assistant has to track the manipulated object and its state, the current task step, and the earlier event that explains the present situation, even when the object is hidden behind the hand or out of view.

Newer benchmarks target specific settings.
EgoTools~\cite{tian2026egotools} tests understanding of physical tools and their effects (not software-tool calls), covering affordance, grounding, procedure, and spatial reasoning, and EgoPolice~\cite{egopolice2026} tests police body-worn camera footage, where consequential events are rare and visibility is poor.

\subsubsection{Autonomous Driving}
Driving video must be read as a dynamic scene of road geometry, traffic controls, vehicles, pedestrians, and their relations.
BDD100K~\cite{yu2020bdd100k} supplies diverse annotated road video, and nuScenes~\cite{caesar2020nuscenes} and the Waymo Open Dataset~\cite{sun2020waymo} add synchronized camera and range-sensor sequences with 3D annotations.
Language-based benchmarks ask which participants and relations matter: DriveLM~\cite{sima2024drivelm} links key objects and their interactions to prediction and planning questions through graph-structured VideoQA, and the frame-level NuScenes-SpatialQA~\cite{tian2025nuscenesspatialqa} tests spatial relations in driving scenes.
OmniDrive~\cite{wang2025omnidrive} aligns language reasoning with 3D driving tasks and adds counterfactual questions about how alternative actions would change an outcome.
The resulting scene state links video reading to the trajectory prediction, action selection, and closed-loop control of Section~\ref{sec:interacting}.

\subsubsection{Domain-Specific Video Analysis}
\leadlabel{Medical and surgical video.}
Surgical video is structured by procedure phases, so recognizing visible instruments is not enough.
EndoNet~\cite{twinanda2016endonet} introduced Cholec80 for phase and tool recognition, CholecT50 and Rendezvous~\cite{nwoye2022rendezvous} refine the output into instrument--verb--target triplets, and SurgVLM~\cite{zeng2025surgvlm} combines perception, temporal analysis, and question answering across surgical tasks.
Evaluation must check phase, anatomy, and action against the structured labels of clinical workflows, not the fluency of a description.

\leadlabel{Sports analytics.}
A match contains long stretches of routine play broken by brief, rule-dependent events.
SoccerNet~\cite{giancola2018soccernet} defined action spotting in broadcast soccer; SoccerNet-v2~\cite{deliege2021soccernetv2} adds camera-shot segmentation and replay grounding, which require precise localization and knowledge of broadcast structure.

\leadlabel{Film and edited video.}
In edited footage, shot size, cut type, and transitions are deliberate choices.
VEU-Bench~\cite{li2025veu} tests recognition, reasoning, and judging of editing attributes within and across shots, and finds that general Video-LLMs handle them poorly, in some cases below chance.

\leadlabel{Education and instructional video.}
Instructional video follows a procedure from an initial state through intermediate steps to a goal.
HowTo100M~\cite{Miech_2019_ICCV} supplies narrated demonstrations, and CrossTask~\cite{zhukov2019crosstask} and COIN~\cite{tang2019coin} annotate steps for temporal localization.
Reading such video means tracking the progress of the procedure: separating narrated intention from visible operation, recovering omitted or repeated steps, and reasoning about order and prerequisites.

\begin{insightbox}{Video understanding is limited by state, not by context length}
Encoders compress observations, Video-LLMs map evidence into language, and agents choose what to inspect, but all of them are limited by what state they keep and ground, not by how many frames they accept.
Longer context helps only when it corrects or stabilizes that state, so evaluation should ask whether added evidence improves the state estimate, stays traceable to source clips, and survives later queries.
\end{insightbox}

\surveytheme{surveyWrite}
\section{\writeicon\opsep Video Generation: Writing the World}
\label{sec:writing}
\chapterlead{Writing means rendering a new video trace from a partial world specification supplied as text, images, audio, motion, or other conditions.}
Video generation is the writing operation $W\rightarrow V$ of Section~\ref{sec:lens}: its conditions fix some entities, relations, and dynamics of $W$ and leave the rest to the generator.
Text may specify an event, an image may fix its initial appearance, audio may set visible rhythm or articulation, and trajectories camera or object motion.

This chapter covers models whose output and training target are the video itself, judged by fidelity, condition adherence, temporal coherence, and controllable revision.
Persistent state under occlusion, revisits, long rollouts, and actions is treated in Section~\ref{sec:interacting}.
We cover tasks and evaluation, latent representations, generative paradigms, control strategies, and applications.
Xing et al.~\cite{xing2023videodiffusionsurvey} review video diffusion models in depth.

\subsection{Tasks, Conditions, and Evaluation}
\label{sec:writing-tasks}
A video-generation task is defined by the information fixed before synthesis: text, images, source videos, audio, reference identities, or prescribed motion.
Evaluation measures the visual and temporal quality of the result, its adherence to the condition, and the stability of the entities, events, and physical relations it shows.
Table~\ref{tab:gen-training-data} lists training resources, Table~\ref{tab:gen-eval} lists metrics and learned scorers, and Table~\ref{tab:gen-benchmarks} lists benchmarks and the conditions they test.

\subsubsection{Generation Tasks by Condition}
\leadlabel{Text, image, and video conditions.}
Text-to-video (T2V) maps a description of content, style, or events to a video and leaves most appearance and motion to the model, as in Sora~\cite{brooks2024video}, Wan~\cite{wan2025open}, and HunyuanVideo~\cite{hunyuanvideo2024}.
Image-to-video (I2V) fixes the initial state or reference appearance and must add plausible motion without losing that structure and identity, as in Stable Video Diffusion~\cite{blattmann2023stable} and the human-animation systems Animate Anyone~\cite{hu2024animate} and MagicAnimate~\cite{xu2023magicanimate}.
Video-to-video (V2V) starts from a source video and covers transformations with different preservation requirements.
Editing changes selected content and should stay local, as in Tune-A-Video~\cite{wu2023tune}, Video-P2P~\cite{liu2023videop2p}, FateZero~\cite{qi2023fatezero}, and TokenFlow~\cite{geyer2023tokenflow}.
Translation maps the source to another appearance or domain while keeping its content.
Interpolation connects observed endpoints, and extension continues the sequence beyond its last frame, as in NUWA-XL~\cite{yin2023nuwa}, SEINE~\cite{chen2023seine}, and StreamingT2V~\cite{henschel2025streamingt2v}.

\leadlabel{Audio and audio-visual conditions.}
In audio-driven generation, a speech, music, or sound track sets visible dynamics such as lip motion, expression, gesture, or rhythm, as in Wav2Lip~\cite{prajwal2020wav2lip}, SadTalker~\cite{zhang2023sadtalker}, EMO~\cite{tian2024emo}, and Hallo~\cite{xu2024hallo}.
Joint audio-video generation instead produces both streams and must keep their events synchronized as well as each stream plausible, as in JavisDiT~\cite{javisdit2025}, JavisDiT++~\cite{liu2026javisditpp}, and Ovi~\cite{ovi2025}.

\leadlabel{Geometric, motion, and identity conditions.}
Camera, motion, and trajectory conditions prescribe how the viewpoint or scene elements evolve.
Camera poses and paths control the observation process, as in CameraCtrl~\cite{he2024cameractrl}, while object trajectories, human poses, and action sequences constrain motion within the scene, as in MotionCtrl~\cite{wang2024motionctrl}, DragNUWA~\cite{yin2023dragnuwa}, VideoComposer~\cite{wang2023videocomposer}, and Wan-Move~\cite{chu2025wanmove}.
Identity- or subject-conditioned generation, as in ConsisID~\cite{yuan2024consisid}, fixes which person, object, or subject persists over time, and that subject must stay stable under pose, viewpoint, and occlusion changes.

\leadlabel{Any-conditioned and unified control.}
Any-conditioned systems accept a variable subset of text, image, audio, structure, motion, and reference inputs, as in FullDiT~\cite{fulldit2025}, UNIC~\cite{unic2025}, and Any2Caption~\cite{wu2025any2caption}.
They must bind each cue to the right entity, region, time span, or modality and reconcile cues of different granularity.
Section~\ref{sec:writing-control} describes how such conditions enter the model.

\leadlabel{Training supervision.}
Each task needs training pairs that expose its condition--output correspondence (Table~\ref{tab:gen-training-data}).
Caption corpora such as Panda-70M~\cite{chen2024panda} and UltraVideo~\cite{xue2025ultravideo} serve T2V, and image conditioning and extension can be learned by keeping selected frames or prefixes of a clip~\cite{jiang2025vace}.
Editing needs source--instruction--target triplets, as in OpenVE-3M~\cite{he2025openve}.
Camera control uses posed video such as RealEstate10K~\cite{zhou2018stereomagnification}, motion control uses tracked point trajectories~\cite{chu2025wanmove}, and audio-visual generation needs synchronized tracks such as VGGSound~\cite{vggsound2020}.
Identity control pairs references and targets from different contexts or shots, as in Phantom-Data~\cite{chen2025phantomdata}, MuSS~\cite{zhang2026muss}, and Omni-R2V~\cite{li2026omnivbench}, so the model cannot simply copy the reference's pose or background into the output.
\begin{table}[!t]
\centering
\small
\caption{Training resources for conditioned video generation and editing. \textbf{Condition} lists available input modalities or control signals; their use depends on the training task. \textbf{Scale} retains the source's counting unit.}
\label{tab:gen-training-data}
\surveytablecolors
\setlength{\tabcolsep}{3.5pt}
\begin{tabularx}{\textwidth}{@{}>{\hsize=.90\hsize}Y >{\hsize=1.02\hsize}Y >{\hsize=.83\hsize}Y >{\hsize=1.25\hsize}Y@{}}
\toprule
\surveytablehead
\tableheadcell{Resource} & \tableheadcell{Condition} & \tableheadcell{Scale} & \tableheadcell{Annotations / construction} \\
\midrule
Panda-70M~\cite{chen2024panda} & Text & 70.8M clips & Captions selected from multiple cross-modality teachers \\
\addlinespace
Koala-36M~\cite{wang2024koala36m} & Text; motion, aesthetic, and clarity scores & 36M clips & Structured captions; Video Training Suitability Score (VTSS) filtering \\
\addlinespace
OpenVid-1M~\cite{nan2025openvid} & Text & 1M pairs; 433K HD subset & Captions with aesthetic, temporal, motion, and clarity filtering \\
\addlinespace
UltraVideo~\cite{xue2025ultravideo} & Text & 42K short; 17K long clips & 4K/8K footage; structured captions and summaries \\
\addlinespace
RealEstate10K~\cite{zhou2018stereomagnification} & Reference image; camera parameters & $\sim$80K clips; 10M frames & Estimated camera intrinsics and poses for view synthesis \\
\addlinespace
DL3DV-10K~\cite{ling2024dl3dv} & Reference image; camera parameters & 10,510 videos; 51.2M frames & 4K multi-view scenes; camera intrinsics, poses, and scene-property labels \\
\addlinespace
OpenHumanVid~\cite{li2025openhumanvid} & Text; skeleton sequences; speech audio & 13.2M filtered clips & Appearance and motion captions; aligned pose and speech signals \\
\addlinespace
Wan-Move corpus~\cite{chu2025wanmove} & Text; reference image; point trajectories & 2M filtered videos & 720p video; dense $32\!\times\!32$ CoTracker tracks \\
\addlinespace
VGGSound~\cite{vggsound2020} & Text; audio & 200K+ 10\,s clips & Synchronized audio/video with sound-source class labels \\
\addlinespace
Phantom-Data~\cite{chen2025phantomdata} & Text; reference image & $\sim$1M pairs & Identity-consistent image--video pairs across contexts \\
\addlinespace
MuSS~\cite{zhang2026muss} & Text; reference image & 700K shots; 30K+ multi-shot clips & Shot-level captions and cross-shot subject references \\
\addlinespace
OpenVE-3M~\cite{he2025openve} & Text; source video & 3M samples & Source--instruction--target triplets; eight edit types \\
\addlinespace
Omni-R2V~\cite{li2026omnivbench} & Text; reference image or video & 340K samples & Reference--target pairs across seven task families \\
\addlinespace
Any2CapIns~\cite{wu2025any2caption} & Text; image; video; region, motion, and camera controls & 337K instances; 407K conditions & Condition-to-caption instruction pairs \\
\bottomrule
\end{tabularx}
\end{table}

\subsubsection{Evaluation and Metrics}
\label{sec:writing-eval}
Video-generation quality has several dimensions, and the right profile depends on the task: T2V emphasizes semantic alignment, editing requires source preservation and locality, audio-driven generation requires synchronization, and motion control requires trajectory adherence.
A single aggregate score hides the failure that limits a system, so we group metrics by the property they measure (Table~\ref{tab:gen-eval}) and benchmarks by the cases they test (Table~\ref{tab:gen-benchmarks}).

\leadlabel{Visual fidelity and temporal quality.}
FID~\cite{heusel2018ganstrainedtimescaleupdate} compares frame-feature distributions and ignores temporal order.
The Inception Score~\cite{salimans2016improved} rewards confident, diverse classifier predictions without reference to real data, which limits its meaning outside the classifier's domain~\cite{barratt2018inception}.
FVD~\cite{unterthiner2019accurategenerativemodelsvideo} extends the comparison to video features, but it depends on the feature extractor and sample count and can under-penalize temporal corruption in favor of static samples~\cite{ge2024fvdbias}.
JEDi~\cite{luo2024jedi} instead uses V-JEPA features and a maximum mean discrepancy, avoiding FVD's Gaussian assumption and needing fewer samples.
PSNR, SSIM~\cite{wang2004ssim}, and LPIPS~\cite{zhang2018unreasonableeffectivenessdeepfeatures} suit reconstruction and source preservation but penalize valid alternative futures.
FVMD~\cite{liu2024fvmd} and STREAM~\cite{kimstream} target motion consistency and spatiotemporal quality, and VMBench~\cite{ling2025vmbench} checks whether motion strength matches the prompt.
Diversity needs repeated samples under one condition: DiVid~\cite{hu2026divid} measures variation in semantics, style, subject, scene, motion, and camera alongside quality and faithfulness.

\leadlabel{Condition alignment and compositionality.}
Condition-aware evaluation checks whether the video follows its condition.
CLIPScore~\cite{hessel2022clipscorereferencefreeevaluationmetric} measures image--caption similarity, and its frame-averaged video variants ignore temporal order and motion.
cFreD~\cite{koo2025cfred} adds text--visual cross-covariances to a conditional Fr\'echet distance, so it penalizes videos swapped among prompts, which a marginal distance cannot detect.
FETV~\cite{liu2023fetvbenchmarkfinegrainedevaluation} and T2V-CompBench~\cite{sun2025t2vcompbenchcomprehensivebenchmarkcompositional} separate objects, attributes, relations, actions, and temporal composition.
Structured conditions have their own tests: MultiRef-Compass~\cite{multirefcompass2026} checks how several references bind to audio-visual content.
OmniVBench~\cite{li2026omnivbench} applies case-specific checklists for reference fidelity and binding, and MoveBench~\cite{chu2025wanmove} measures trajectory endpoint error.
For source-conditioned tasks, V2V-Bench~\cite{liu2026v2vbench} scores what the instruction should change separately from what the source should preserve.

\leadlabel{Physical and world consistency.}
Realistic local motion can still show impossible interactions, unstable objects, or broken persistence.
VideoPhy~\cite{bansal2024videophy} and VideoPhy-2~\cite{bansal2025videophy2} test physical commonsense, the latter in harder action-centric interactions.
PhysicsMind~\cite{physicsmind2025} and VideoScience-Bench~\cite{hu2025videoscience} test mechanics and basic scientific phenomena, and PhyAVBench~\cite{phyavbench2025} tests whether generated sound changes with paired physical conditions.
PhyProbe~\cite{ku2026phyprobe} instead trains a physical-violation scorer on frozen visual features and shows that prior learned evaluators depend on their annotation distributions, though it retains some dataset dependence itself.
Persistence under revisits and occlusion is tested by the world-model benchmarks of Section~\ref{sec:wm-benchmarks}.

\leadlabel{Human preference, learned evaluators, and suites.}
Human comparisons capture coherence, visual appeal, and instruction adherence that fixed feature distances miss.
Learned evaluators model such judgments: VideoScore~\cite{he2024videoscore} predicts several human-rated properties, VideoScore2~\cite{he2025videoscore2} adds physical and commonsense scores with rationales, and MJ-VIDEO~\cite{tong2025mjvideo} trains a reward model on fine-grained preferences.
All three inherit the coverage and biases of their annotations, which for VideoScore2 are partly model-generated.
Suites such as VBench~\cite{huang2023vbenchcomprehensivebenchmarksuite} and EvalCrafter~\cite{liu2024evalcrafterbenchmarkingevaluatinglarge} combine perceptual, semantic, and temporal dimensions, and VBench-2.0~\cite{zheng2025vbench20advancingvideogeneration} extends them to human fidelity, controllability, creativity, physics, and commonsense.

\leadlabel{Multi-shot and audio-visual generation.}
Multi-shot and audio-visual outputs also need evaluation across shot boundaries and modalities.
MSVBench~\cite{shi2026msvbench} scores story alignment, cross-shot consistency, visual quality, and motion, while the Cinematic Narrative Benchmark of MuSS~\cite{zhang2026muss} separates fidelity to external references from identity persistence across shots and detects reference copying.
JavisBench~\cite{javisdit2025} tests joint generation of scenes with multiple or overlapping sound events, T2AV-Compass~\cite{t2avcompass2025} standardizes text-to-audio-video evaluation, and LongAV-Compass~\cite{longavcompass2026} extends it to minute-scale settings where identity, narrative continuity, and synchronization can fail at different times.
At the metric level, Fr\'echet Audio Distance~\cite{kilgour2018fad} compares audio-feature distributions but ignores the video, LSE-D and LSE-C~\cite{prajwal2020wav2lip} measure speech--mouth alignment, and JavisScore~\cite{javisdit2025} measures audio-video correspondence in broader sounding scenes.
\begin{table}[!t]
\centering
\small
\caption{Measurement functions and learned scorers for video generation. Benchmarks appear in Table~\ref{tab:gen-benchmarks}.}
\label{tab:gen-eval}
\surveytablecolors
\setlength{\tabcolsep}{5pt}
\begin{tabularx}{\textwidth}{@{}>{\hsize=.74\hsize}Y >{\hsize=.95\hsize}Y >{\hsize=1.31\hsize}Y@{}}
\toprule
\surveytablehead
\tableheadcell{Evaluation aspect} & \tableheadcell{Metrics and scorers} &
\tableheadcell{What is measured} \\
\midrule
\surveygroup{3}{Perceptual and temporal quality}
Distributional quality &
IS~\cite{salimans2016improved}, FID~\cite{heusel2018ganstrainedtimescaleupdate}, FVD~\cite{unterthiner2019accurategenerativemodelsvideo}, tFVD~\cite{guo2026vrae}, JEDi~\cite{luo2024jedi} &
classifier confidence/class coverage (IS); feature-distribution distances (FID/FVD/JEDi), not direct physical validity \\
\addlinespace
Reference fidelity &
PSNR, SSIM~\cite{wang2004ssim}, LPIPS~\cite{zhang2018unreasonableeffectivenessdeepfeatures} &
pixel, structural, or perceptual agreement with an aligned target; unsuitable as sole criteria for open-ended futures \\
\addlinespace
Conditional diversity &
DiVid descriptors~\cite{hu2026divid} &
variation across repeated samples in semantics, style, subject, scene, motion, and camera, alongside faithfulness \\
\addlinespace
Motion and temporal quality &
FVMD~\cite{liu2024fvmd}, STREAM~\cite{kimstream} &
motion-feature distributions and separated spatial/temporal quality; dependent on tracking or feature extraction \\
\addlinespace
\surveygroup{3}{Condition fidelity}
Prompt alignment &
CLIPScore~\cite{hessel2022clipscorereferencefreeevaluationmetric}, cFreD~\cite{koo2025cfred} &
text--visual similarity or prompt-conditioned distributional agreement; not a direct event-order test \\
\addlinespace
Trajectory adherence &
Tracked endpoint error (EPE)~\cite{chu2025wanmove} &
deviation from prescribed point trajectories; tracker failures and occlusion affect the result \\
\addlinespace
Audio quality and synchronization &
FAD~\cite{kilgour2018fad}, LSE-D/LSE-C~\cite{prajwal2020wav2lip}, JavisScore~\cite{javisdit2025} &
audio-feature distributions, speech--mouth alignment, and event-level audio-video correspondence \\
\addlinespace
\surveygroup{3}{World consistency and holistic quality}
Physical violation scoring &
PhyProbe~\cite{ku2026phyprobe} &
learned physical-violation scores from frozen visual features; human agreement and cross-dataset transfer qualify the score \\
\addlinespace
Multidimensional learned assessment &
VideoScore~\cite{he2024videoscore}, VideoScore2~\cite{he2025videoscore2} &
human-annotation-derived quality and alignment scores; VideoScore2 adds physical/common-sense scores and rationales \\
\addlinespace
Human preference and learned rewards &
MJ-VIDEO~\cite{tong2025mjvideo}, VideoReward~\cite{liu2025videoreward},
LiFT~\cite{wang2024lift} &
pairwise preference and multidimensional reward; annotation coverage and preference aggregation affect interpretation \\
\bottomrule
\end{tabularx}
\end{table}

\begin{table}[!t]
\centering
\small
\caption{Benchmarks for conditioned video and audio-video generation and editing. \textbf{Condition} is the supplied modality or control signal; \textbf{Evaluation target} is the tested property. Alternative input settings are labeled where needed.}
\label{tab:gen-benchmarks}
\surveytablecolors
\setlength{\tabcolsep}{3.5pt}
\begin{tabularx}{\textwidth}{@{}>{\hsize=.88\hsize}Y >{\hsize=.94\hsize}Y >{\hsize=1.18\hsize}Y@{}}
\toprule
\surveytablehead
\tableheadcell{Benchmark} & \tableheadcell{Condition} & \tableheadcell{Evaluation target} \\
\midrule
VBench / EvalCrafter~\cite{huang2023vbenchcomprehensivebenchmarksuite,liu2024evalcrafterbenchmarkingevaluatinglarge} & Text & Frame appearance, temporal consistency, and text--video alignment \\
\addlinespace
VBench++~\cite{huang2024vbenchpp} & Text or reference image & Video appearance and motion; image--video subject/background consistency \\
\addlinespace
FETV / T2V-CompBench~\cite{liu2023fetvbenchmarkfinegrainedevaluation,sun2025t2vcompbenchcomprehensivebenchmarkcompositional} & Text & Content coverage; attribute, action, motion, and relation composition \\
\addlinespace
VMBench~\cite{ling2025vmbench} & Text & Object integrity, motion amplitude/smoothness, temporal coherence, and commonsense \\
\addlinespace
MoveBench~\cite{chu2025wanmove} & Text; reference image; point tracks or masks & Trajectory adherence and motion consistency \\
\addlinespace
OmniVBench~\cite{li2026omnivbench} & Text; reference image or video & Reference fidelity, attribute binding, and separation of content, motion, and style \\
\addlinespace
OpenVE-Bench~\cite{he2025openve} & Text; source video & Edit-instruction adherence; unedited-content preservation and temporal consistency \\
\addlinespace
V2V-Bench~\cite{liu2026v2vbench} & Text; source video & Edit faithfulness, source preservation, and temporal consistency \\
\addlinespace
JavisBench / T2AV-Compass~\cite{javisdit2025,t2avcompass2025} & Text & Audio fidelity, video appearance/motion, and audio--video synchronization \\
\addlinespace
LongAV-Compass~\cite{longavcompass2026} & Text; optional reference image or video & Minute-scale visual, acoustic, and semantic continuity \\
\addlinespace
MultiRef-Compass~\cite{multirefcompass2026} & Text; reference image; optional video and audio & Reference identity and role binding across visual and audio content \\
\addlinespace
MSVBench / MuSS narrative~\cite{shi2026msvbench,zhang2026muss} & Text; reference image & Story alignment and subject consistency across shots \\
\addlinespace
VideoPhy / VideoPhy-2~\cite{bansal2024videophy,bansal2025videophy2} & Text & Physical plausibility of interactions and adherence to described actions \\
\addlinespace
VBench-2.0~\cite{zheng2025vbench20advancingvideogeneration} & Text & Human fidelity, controllability, creativity, physics, and commonsense \\
\addlinespace
PhysicsMind (generation)~\cite{physicsmind2025} & Text; reference image & Center-of-mass, lever-equilibrium, and inertial-motion consistency \\
\addlinespace
VideoScience-Bench~\cite{hu2025videoscience} & Text; optional reference image & Scientific setup, dynamics, continuity, and expected physical/chemical phenomena \\
\addlinespace
PhyAVBench~\cite{phyavbench2025} & T2AV: text; I2AV: text + reference image; V2A: source video & Sound changes under paired physical conditions; contrastive response scores \\
\bottomrule
\end{tabularx}
\end{table}

\subsection{Video Latent Representation Learning}
\begin{figure}[t]
  \centering
  \includegraphics[width=0.96\textwidth]{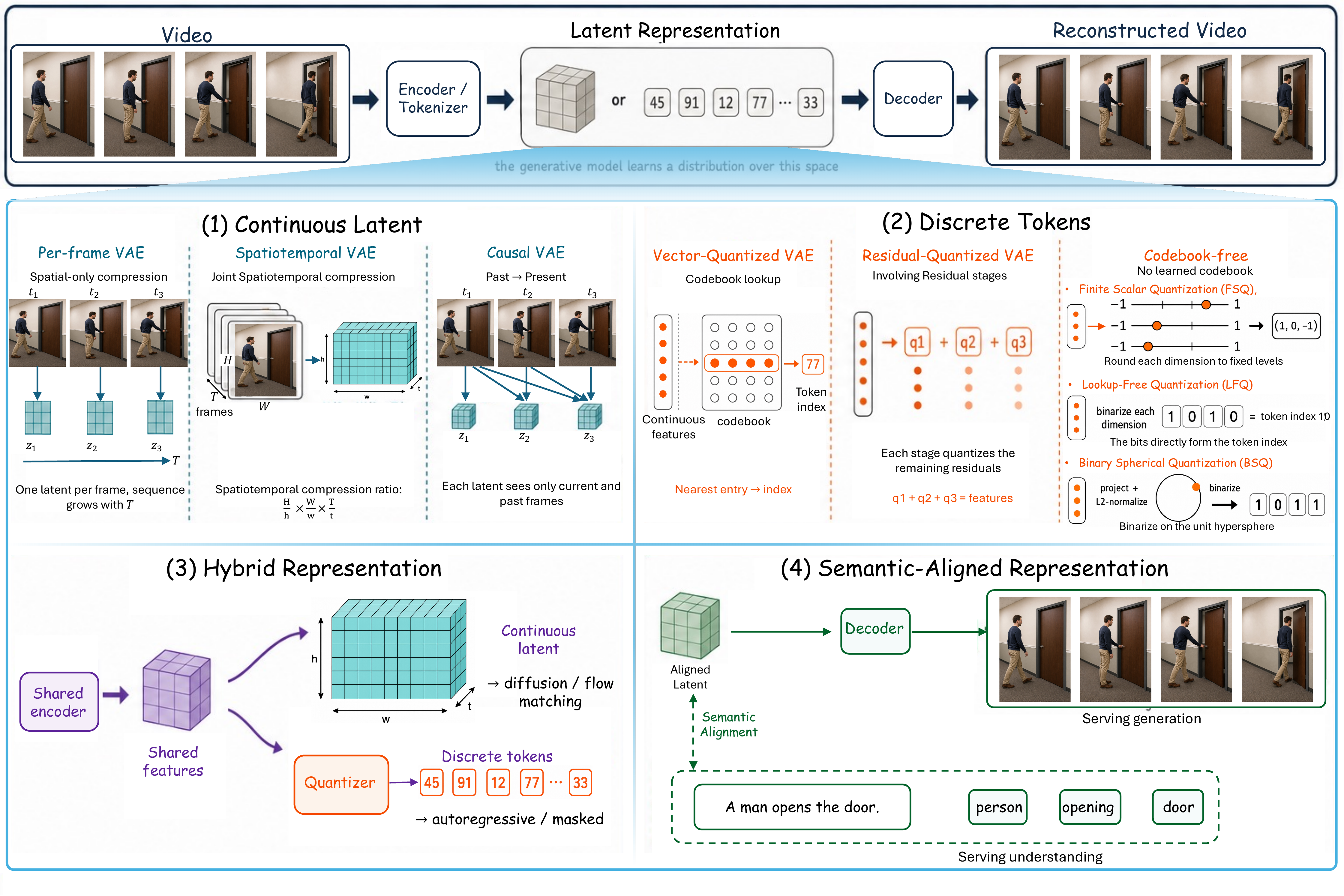}
  \caption{Latent representations for video generation.
  Continuous, discrete, hybrid, and semantically aligned designs differ in how tokenizers and VAEs shape reconstruction fidelity, compression, sequence modeling, and unification with understanding.
  Semantic alignment can be applied to either continuous or discrete representations.}
  \label{fig:generation-latents}
\end{figure}

Many current video generators work in a compressed latent space: an encoder maps video to a shorter representation, the generator models its distribution, and a decoder reconstructs the frames.
Compression lowers training and sampling cost, especially for long videos, while the bottleneck and decoder decide whether fine texture, fast motion, and small objects survive.
Figure~\ref{fig:generation-latents} contrasts the four families: continuous latents for diffusion and flow objectives, discrete tokens for autoregressive or masked prediction, hybrids that mix continuous and discrete forms or pair a tokenizer with a diffusion decoder, and semantically aligned representations that also serve understanding.

\subsubsection{Continuous Latents}
\leadlabel{Per-frame image VAEs.}
Latent diffusion~\cite{rombach2022high} established the value of training a generative model in the continuous space of a pretrained image autoencoder, and the simplest video extension encodes each frame independently.
VideoLDM~\cite{blattmann2023videoldm} and AnimateDiff~\cite{guo2024animatediff} reuse image autoencoders and add temporal layers to image diffusion backbones.
This keeps the image prior and eases reuse of text-to-image backbones.
However, the latent sequence still grows linearly with the number of frames, and temporal coherence must be learned by the generator or by temporal layers in the decoder.

\leadlabel{Spatiotemporal and Causal VAEs.}
Video VAEs use 3D or factorized spatiotemporal operations to compress both resolution and time.
Temporal compression shortens the sequence, lowers attention cost, and exploits redundancy between nearby frames, but an aggressive bottleneck can smooth brief events or fast motion.
CV-VAE~\cite{zhao2024cvvae} keeps its video latent space compatible with a pretrained image VAE so that image-generation priors remain usable, whereas CogVideoX~\cite{yang2024cogvideox} and Cosmos~\cite{nvidia2025cosmos} use dedicated 3D tokenizers for high spatial and temporal compression.
A causal 3D VAE computes the latent at time $t$ without future frames and encodes the first frame without preceding context, so one tokenizer handles both images and videos and streaming or autoregressive generators see no future leakage.

\subsubsection{Discrete Tokens}
\leadlabel{From vector quantization to codebook-free schemes.}
VQ-VAE~\cite{van2017neural} replaces each encoder output with the index of a learned codebook entry, producing categorical symbols that next-token or masked-token models can predict.
VQGAN~\cite{esser2021taming} improves high-resolution reconstruction with perceptual and adversarial losses, and RQ-VAE~\cite{lee2022rqvae} stacks residual codes to increase precision without enlarging the token grid.
Learned codebooks, however, may be used unevenly or collapse to a small active subset.
Later quantizers drop the learned codebook: FSQ~\cite{mentzer2024fsq} rounds a few scalar dimensions to fixed levels, LFQ in MAGVIT-v2~\cite{yu2024magvit} uses binary dimensions to obtain a very large implicit vocabulary, and BSQ~\cite{zhao2024bsq} projects embeddings onto a hypersphere before binary quantization.
They differ in optimization stability, vocabulary size, and distortion at a given token budget.

\leadlabel{Video tokenization.}
The video encoder also fixes the spatiotemporal extent of each token and hence the sequence length.
C-ViViT in Phenaki~\cite{villegas2023phenaki} factorizes spatial and temporal encoding into causal discrete tokens, MAGVIT-v2 pairs a causal tokenizer with LFQ for both autoregressive and masked generation, and HiTVideo~\cite{zhou2025hitvideo} shortens the sequence with hierarchical codes that place semantics and fine detail in different layers.
Fixed grids, however, spend as many tokens on a static shot as on a fast-changing scene.
Adaptive tokenizers allocate the budget by content: EVATok~\cite{xiong2026evatok} predicts how many tokens each segment needs, and VideoFlexTok~\cite{atanov2026videoflextok} emits a variable-length coarse-to-fine sequence whose early tokens carry semantics and motion and whose later tokens add detail.

\subsubsection{Hybrid and Semantically Aligned Representations}
Hybrid tokenizers combine continuous and discrete forms or change the decoder.
OneVAE~\cite{zhou2025onevae} initializes a discrete video VAE from a continuous one and trains both in one network, so continuous latents can feed diffusion or flow models and discrete codes can feed sequence models.
Divot~\cite{ge2024divot} and DiCoDe~\cite{li2024dicode} instead decode with a diffusion model.
Divot learns continuous features for LLM-based understanding and reconstruction, and DiCoDe compresses video into deep tokens that an LLM predicts autoregressively.

Semantic alignment adds a further requirement.
A reconstruction-optimal latent may keep texture and local motion without organizing them into the objects, events, and relations needed for understanding, while a highly semantic latent may discard the detail that generation needs.
VILA-U~\cite{wu2025vilauunifiedfoundationmodel} trains one discrete tokenizer for both reconstruction and text alignment, and the SEED tokenizer~\cite{ge2024seedllama} gives a language model a semantically organized visual code.
PyraTok~\cite{pyratok2026} quantizes features at several spatiotemporal scales against one text-guided codebook, so coarse semantics and fine detail occupy different levels.
Section~\ref{sec:sharing} pursues this question in models that share LLMs, parameters, or objectives across understanding and generation.

Representation autoencoders start from the semantic side: they freeze a pretrained encoder as the tokenizer and train only a projector or pooler and a decoder.
V-RAE~\cite{guo2026vrae} compares image and video encoders such as DINOv3 and V-JEPA 2.1 and finds that reconstruction ranking does not consistently predict generation quality, so a semantic latent must be tested inside the target generator.

\subsection{Generative Video Foundation Modeling}
\label{sec:writing-models}
\begin{figure}[t]
  \centering
  \includegraphics[width=0.90\textwidth]{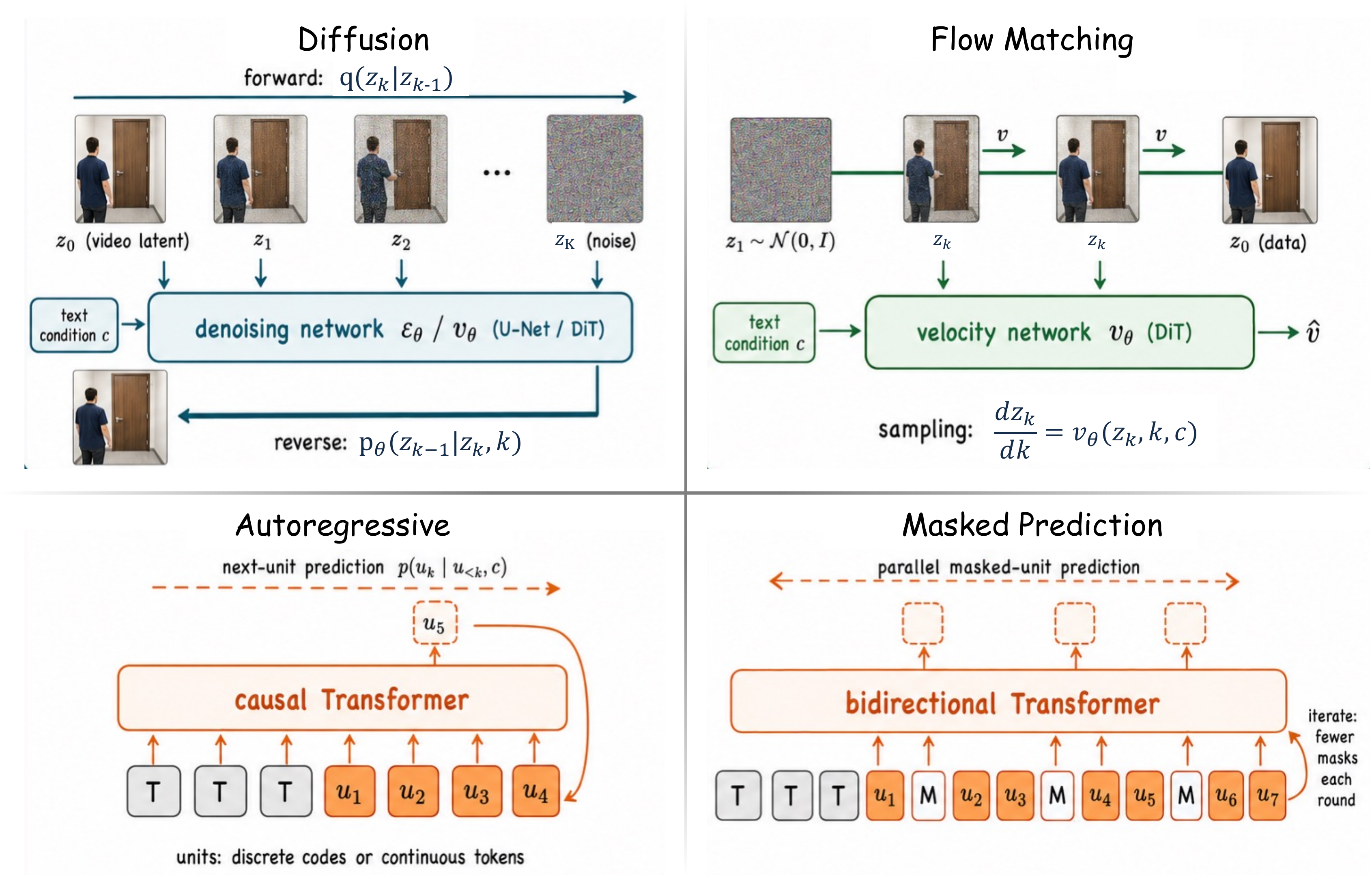}
  \caption{Generative modeling paradigms.
  Diffusion and flow matching learn denoising or velocity targets over video latents.
  Autoregressive modeling factorizes an ordered sequence, while masked prediction fills selected positions from visible context through parallel conditional predictions.}
  \label{fig:generation-paradigms}
\end{figure}

Video generators differ in their generative objective, the order in which space-time units are predicted, and the system that organizes synthesis.
Diffusion, autoregressive or masked prediction, and flow matching define model-level generation processes (Figure~\ref{fig:generation-paradigms}), whereas LLM/MLLM-assisted and agentic systems add planning, tool use, and revision around a visual generator that may use any of the three.
The levels combine in practice: a system may denoise each chunk while rolling out autoregressively in time, and an agent may repeatedly call a flow-based generator.
Preference optimization is applied on top of these generators.
VideoReward~\cite{liu2025videoreward} aligns a flow-based generator with Flow-DPO on a multidimensional preference signal, and LiFT~\cite{wang2024lift} applies reward-weighted fine-tuning with a reward model learned from human annotations.
World-R1~\cite{worldr1_2026} uses feedback from pretrained 3D and vision-language models to reinforce geometric consistency through Flow-GRPO.
Table~\ref{tab:generation-systems} compares representative systems whose technical routes are publicly documented.

\begingroup
\small
\setlength{\tabcolsep}{4pt}
\renewcommand{\arraystretch}{1.06}
\setlength{\surveytableunit}{\dimexpr(\textwidth-3.25cm-10\tabcolsep)/4\relax}
\setlength{\LTcapwidth}{\textwidth}
\setlength{\LTpre}{8pt}
\setlength{\LTpost}{8pt}
\Needspace{.35\textheight}
\surveytablecolors
\begin{longtable}{@{}>{\raggedright\arraybackslash}p{1.20\surveytableunit} >{\raggedright\arraybackslash}p{.85\surveytableunit} >{\raggedright\arraybackslash}p{\surveytableunit} >{\centering\arraybackslash}p{1.25cm} >{\raggedright\arraybackslash}p{.95\surveytableunit} >{\centering\arraybackslash}p{2cm}@{}}
\hiderowcolors
\caption{Video-generation systems. \textbf{Visual space} describes generated variables; \textbf{MLLM online} identifies a multimodally pretrained backbone used at inference, excluding offline annotators. Modes refer to the cited configurations; MAGVIT-v2 is a tokenizer without mode assignments. Closed systems appear only where an official technical source documents the listed capability.}
\label{tab:generation-systems}\\
\showrowcolors
\toprule
\surveytablehead
\tableheadcell{System} & \tableheadcell{Architecture} & \tableheadcell{Learning} & \tableheadcell{Visual space} & \tableheadcell{MLLM online} & \tableheadcell{Modes}\par\generationmodehead \\
\midrule
\endfirsthead
\hiderowcolors
\caption[]{Video-generation systems (continued).}\\
\showrowcolors
\toprule
\surveytablehead
\tableheadcell{System} & \tableheadcell{Architecture} & \tableheadcell{Learning} & \tableheadcell{Visual space} & \tableheadcell{MLLM online} & \tableheadcell{Modes}\par\generationmodehead \\
\midrule
\endhead
\bottomrule
\rowcolor{white}\multicolumn{6}{@{}p{\textwidth}@{}}{\generationtablekey}\\
\noalign{\vskip0pt}
\endfoot
\surveylonggroup{6}{Diffusion over pixels or continuous latents}
Video Diffusion Models (text)~\cite{ho2022video} &
U-Net & Diffusion & \spacecode{Px} & \tableno & \generationmodes{1}{0}{0}{0}{0} \\
Video Diffusion Models (prediction)~\cite{ho2022video} &
U-Net & Diffusion & \spacecode{Px} & \tableno & \generationmodes{0}{1}{1}{0}{0} \\
Imagen Video~\cite{ho2022imagen} &
Cascaded U-Nets & Diffusion & \spacecode{Px} & \tableno & \generationmodes{1}{0}{0}{0}{0} \\
Make-A-Video~\cite{singer2023makeavideo} &
U-Net cascade & Diffusion & \spacecode{Px} & \tableno & \generationmodes{1}{1}{1}{0}{0} \\
\addlinespace
Video LDM (text)~\cite{blattmann2023videoldm} &
U-Net & Diffusion & \spacecode{VAE} & \tableno & \generationmodes{1}{0}{0}{0}{0} \\
Video LDM (RDS prediction)~\cite{blattmann2023videoldm} &
Frame-conditioned U-Net & Diffusion & \spacecode{VAE} & \tableno & \generationmodes{0}{1}{1}{0}{1} \\
Stable Video Diffusion (I2V)~\cite{blattmann2023stable} &
U-Net & Diffusion & \spacecode{VAE} & \tableno & \generationmodes{0}{1}{0}{0}{0} \\
Sora~\cite{brooks2024video} &
DiT & Diffusion & \spacecode{VAE} & \spacecode{NR} & \generationmodes{1}{1}{1}{0}{1} \\
CogVideoX~\cite{yang2024cogvideox} &
DiT & Diffusion & \spacecode{VAE} & \tableyes{} I2V prompt expansion & \generationmodes{1}{1}{0}{0}{0} \\
\surveylonggroup{6}{Flow matching over continuous video latents}
Mochi~\cite{genmo2024mochi} &
Asymmetric DiT & Flow matching & \spacecode{VAE} & \tableno & \generationmodes{1}{0}{0}{0}{0} \\
LTX-Video~\cite{hacohen2024ltxvideo} &
DiT & Flow matching & \spacecode{VAE} & \tableno & \generationmodes{1}{1}{0}{0}{0} \\
HunyuanVideo (T2V)~\cite{hunyuanvideo2024} &
Dual-stream DiT & Flow matching & \spacecode{VAE} & \tableyes{} text encoder & \generationmodes{1}{0}{0}{0}{0} \\
Step-Video-T2V~\cite{ma2025stepvideo} &
DiT & Flow matching & \spacecode{VAE} & \tableno{} LLM only & \generationmodes{1}{0}{0}{0}{0} \\
HunyuanVideo 1.5~\cite{wu2025hunyuan15} &
DiT & Flow matching & \spacecode{VAE} & \tableyes{} text encoder & \generationmodes{1}{1}{0}{0}{0} \\
Wan / Wan-VACE~\cite{wan2025open,jiang2025vace} &
DiT & Flow matching & \spacecode{VAE} & \tableoptional{} Wan I2V prompt expansion & \generationmodes{1}{1}{1}{0}{0} \\
Pyramid Flow~\cite{jin2024pyramidflow} &
DiT & Flow matching & \spacecode{VAE} & \tableno & \generationmodes{1}{1}{0}{0}{0} \\
\addlinespace
\surveylonggroup{6}{Joint audio-video generation}
Ovi~\cite{ovi2025} &
Twin DiTs & Flow matching & \spacecode{VAE} & \tableno & \generationmodes{1}{0}{0}{1}{0} \\
LTX-2~\cite{hacohen2026ltx2} &
Dual-stream DiT & Flow matching & \spacecode{VAE} & \tableyes{} text encoder & \generationmodes{1}{1}{0}{1}{0} \\
\surveylonggroup{6}{Autoregressive and masked generation}
VideoPoet~\cite{kondratyuk2024videopoet} &
Causal Transformer & Next-token AR & \spacecode{Tok} & \tableyes{} generative core & \generationmodes{1}{1}{1}{1}{0} \\
Loong~\cite{wang2024loong} &
Causal Transformer & Next-token AR & \spacecode{Tok} & \tableyes{} generative core & \generationmodes{1}{0}{0}{0}{1} \\
Phenaki~\cite{villegas2023phenaki} &
Masked Transformer & Masked prediction & \spacecode{Tok} & \tableno & \generationmodes{1}{1}{0}{0}{1} \\
MAGVIT-v2~\cite{yu2024magvit} &
Tokenizer + Transformer & AR / masked prediction & \spacecode{Tok} & \tableno & \generationmodes{0}{0}{0}{0}{0} \\
Lumos-1~\cite{yuan2025lumos} &
Temporal causal Transformer & Temporal AR; masked discrete diffusion & \spacecode{Tok} & \tableyes{} generative core & \generationmodes{1}{1}{0}{0}{0} \\
NOVA~\cite{deng2024nova} &
Temporal AR Transformer & Continuous AR with diffusion loss & \spacecode{VAE} & \tableno & \generationmodes{1}{1}{0}{0}{0} \\
VideoMAR~\cite{yu2025videomar} &
Temporal AR Transformer & Continuous AR with diffusion loss & \spacecode{VAE} & \tableno & \generationmodes{0}{1}{0}{0}{0} \\
\addlinespace
\surveylonggroup{6}{Streaming and long-horizon hybrids}
StreamingT2V~\cite{henschel2025streamingt2v} &
U-Net with memory & Diffusion & \spacecode{VAE} & \tableno & \generationmodes{1}{0}{0}{0}{1} \\
MAGI-1~\cite{teng2025magi} &
Chunk-AR DiT & Causal flow matching & \spacecode{VAE} & \tableyes{} I2V prompt refinement & \generationmodes{1}{1}{0}{0}{1} \\
LongTake~\cite{park2026longtake} &
Chunk-AR DiT & Flow matching; distribution-matching distillation & \spacecode{VAE} & \tableno & \generationmodes{1}{0}{0}{0}{1} \\
\surveylonggroup{6}{LLM/MLLM-assisted generation}
VideoDirectorGPT~\cite{lin2024videodirectorgpt} &
Planner + diffusion model & Diffusion; prompted planning & \spacecode{VAE} & \tableno{} LLM only & \generationmodes{1}{0}{0}{0}{0} \\
\addlinespace
MetaCanvas~\cite{lin2025metacanvas} &
MLLM + latent-canvas generator & Planner alignment; flow backbone & \spacecode{VAE} & \tableyes{} planner & \generationmodes{1}{1}{1}{0}{0} \\
\addlinespace
TV2TV~\cite{han2025tv2tv} &
Mixture of Transformers & Text AR + video flow matching & \spacecode{VAE} & \tableyes{} generative core & \generationmodes{1}{0}{0}{0}{0} \\
\addlinespace
\surveylonggroup{6}{Agentic generation workflows}
MovieAgent~\cite{wu2025movieagent} &
Agents + visual tools & Prompted CoT planning; backend-dependent & \spacecode{BD} & \tableoptional{} GPT-4o planner & \generationmodes{1}{0}{0}{0}{1} \\
\addlinespace
Anim-Director~\cite{li2024animdirector} &
MLLM + visual tools & Prompted planning; backend-dependent & \spacecode{BD} & \tableyes{} planner / observer & \generationmodes{1}{0}{0}{0}{0} \\
\addlinespace
Any2Caption~\cite{wu2025any2caption} &
Condition interpreter + generator & Condition alignment; caption instruction tuning & \spacecode{BD} & \tableyes{} interpreter & \generationmodes{1}{1}{1}{0}{0} \\
\addlinespace
Mora~\cite{yuan2024mora} &
Agents + visual tools & Self-modulated joint fine-tuning of visual experts & \spacecode{VAE} & \tableno{} LLM only; MLLM offline & \generationmodes{1}{1}{1}{0}{0} \\
\end{longtable}
\endgroup

\subsubsection{Diffusion-Based Generation}
Diffusion models learn to reverse a gradual corruption process: a clean video latent is perturbed according to a noise schedule, and the model learns to predict the noise, a velocity, or an equivalent denoising target at any noise level~\cite{ho2020denoising,rombach2022high}.
The $\epsilon$- and $v$-prediction objectives and EDM-style schedules are alternative parameterizations of this problem (Section~\ref{sec:prelim-generative}); they affect optimization and sampling, not the video architecture.
Flow-matching generators (Section~\ref{sec:writing-flow}) share these backbones, so the backbone and long-video designs below apply to both.

\leadlabel{Backbone and attention.}
Early video diffusion systems, including Video Diffusion Models~\cite{ho2022video}, Imagen Video~\cite{ho2022imagen}, and Make-A-Video~\cite{singer2023makeavideo}, extended convolutional U-Nets with temporal convolutions or factorized spatial and temporal attention.
Lumiere~\cite{bar2024lumiere} instead generates the whole clip in one pass with a space-time U-Net that downsamples in both space and time, avoiding keyframe generation followed by temporal super-resolution.
Diffusion Transformers (DiTs)~\cite{Peebles_2023_ICCV} then made it easier to scale attention over compressed space-time tokens.
CogVideoX~\cite{yang2024cogvideox}, Mochi~\cite{genmo2024mochi}, and LTX-Video~\cite{hacohen2024ltxvideo} pair a causal or highly compressed video VAE with a DiT.
CogVideoX uses an expert Transformer with separate text and video normalization, Mochi gives the visual stream more capacity in an asymmetric DiT, and LTX-Video compresses aggressively so that full spatiotemporal attention stays affordable.
Joint audio-video backbones couple two modality streams through bidirectional cross-modal attention, so timing and semantics are exchanged throughout synthesis.
LTX-2~\cite{hacohen2026ltx2} extends LTX-Video with separate modality VAEs and asymmetric audio and video streams, and Ovi~\cite{ovi2025} couples matched audio and video DiTs trained with flow matching.

The main backbone trade-off concerns attention.
Factorized attention separates spatial and temporal interactions to save cost, whereas full spatiotemporal attention lets all space-time tokens interact at a higher memory cost.
Sparse and linear attention cut this cost for long or high-resolution clips.
HunyuanVideo 1.5~\cite{wu2025hunyuan15} uses selective and sliding tile attention within a unified T2V and I2V training framework, and SANA-Video 2.0~\cite{chen2026sanavideo2} interleaves gated linear attention with periodic softmax layers and routes block summaries across depth.
Conditions enter the backbone through cross-attention, adaptive normalization such as AdaLN, joint text-video attention, or channel and token concatenation (Section~\ref{sec:writing-control}).

\leadlabel{Streaming and long videos.}
Standard diffusion denoises a fixed-length clip as a whole, whereas streaming variants generate causally, chunk by chunk.
Diffusion Forcing~\cite{chen2024diffusionforcing} gives each frame its own noise level, MAGI-1~\cite{teng2025magi} denoises chunks autoregressively, CausVid~\cite{yin2025causvid} distills a bidirectional model into a few-step causal generator, Self-Forcing~\cite{selfforcing2025} trains on the model's own rollouts, and Causal Forcing~\cite{zhu2026causalforcing} distills from an autoregressive teacher (Section~\ref{sec:wm-memory}, Table~\ref{tab:forcing-family}).
S2PD~\cite{hu2026s2pd} instead changes the sampling order within a clip, generating blocks autoregressively at high noise, then refining the sequence in parallel at low noise.

Long videos raise a different problem: repeated local generation keeps short-term motion but gradually loses appearance and global structure.
StreamingT2V~\cite{henschel2025streamingt2v} carries anchor frames and short- and long-term memory across windows, FreeLong~\cite{lu2024freelong} blends low-frequency global features with high-frequency local ones during denoising without training, and FramePack~\cite{zhang2025framepack} compresses past frames by importance to bound the context.
Other methods change training or the history representation.
LongTake~\cite{park2026longtake} supervises later frames from long real-video prefixes before self-rollout distribution matching, and its evaluation separates sustained motion from appearance quality, since a nearly static continuation can score well on consistency.
Test-time training~\cite{dalal2025tttvideo} stores context in MLP weights that a self-supervised loss updates while the sequence is processed, and has been demonstrated on one-minute, storyboard-conditioned cartoons.

\subsubsection{Autoregressive and Masked Token Generation}
Autoregressive video models factorize the distribution over an ordered sequence of video units.
With discrete tokens, this resembles language modeling: VideoGPT~\cite{yan2021videogpt} paired a VQ-VAE with a Transformer prior over video codes, CogVideo~\cite{hongcogvideo} scaled the recipe to text-conditioned generation, and VideoPoet~\cite{kondratyuk2024videopoet} applied it in a multimodal sequence model.
Loong~\cite{wang2024loong} places text and video tokens in one sequence and uses short-to-long training and inference-time re-encoding to limit error accumulation in minute-level videos.
InfinityStar~\cite{infinitystar2025} extends discrete autoregression to 720p image, video, and long-video generation, reporting about 30 times lower latency than Wan 2.1-14B in its own single-GPU setting.
The route offers a standard causal likelihood and flexible length, at the cost of long decoding chains, dependence on tokenizer quality, and errors that propagate through the prefix.

Masked generation relaxes the one-token-at-a-time order: the model predicts sets of masked positions from visible tokens, often refining coarse-to-fine over several rounds.
Phenaki~\cite{villegas2023phenaki} applies a bidirectional masked Transformer to causal video tokens, MAGVIT-v2~\cite{yu2024magvit} improves the tokenizer and the masked generation framework, and Lumos-1~\cite{yuan2025lumos} keeps frames causal while decoding each frame in parallel with mask-based discrete diffusion.
Continuous tokens support the same factorization.
MAR~\cite{li2024mar} introduced a diffusion loss for the conditional distribution of a continuous image token, and NOVA~\cite{deng2024nova} and VideoMAR~\cite{yu2025videomar} carry it to video, predicting frames causally in time and continuous tokens within each frame with a diffusion loss.

\subsubsection{Flow-Based Generation}
\label{sec:writing-flow}
Flow matching~\cite{lipman2023flow} trains a time-dependent vector field that transports samples from a simple base distribution to the data along a prescribed probability path, and generation integrates the corresponding ordinary differential equation.
It differs from the usual diffusion account in its training target and sampling path: rectified-flow formulations favor nearly straight paths, which can reduce the number of solver steps, and the same DiT or multimodal DiT (MMDiT) can parameterize either target.
Stable Diffusion 3~\cite{esser2024sd3} made the rectified-flow and MMDiT combination prominent for images, and video systems adapted it to space-time latents.
HunyuanVideo~\cite{hunyuanvideo2024}, Step-Video-T2V~\cite{ma2025stepvideo}, Goku~\cite{goku2025}, and Wan~\cite{wan2025open} scale flow matching with DiT backbones and video VAEs, HunyuanVideo with a dual-stream-to-single-stream Transformer.
Pyramid Flow~\cite{jin2024pyramidflow} changes the path itself, linking spatial pyramid stages and compressing full-resolution history with a temporal pyramid so that one DiT is trained end to end across resolutions.

Normalizing flows are a distinct, likelihood-based route that should not be confused with flow matching.
STARFlow-V~\cite{gu2025starflowv} applies invertible autoregressive transformations to pretrained video-VAE latents, with a global--local decomposition and a light causal denoiser, so the likelihood is tractable in latent space.

\begin{figure}[t]
  \centering
  \includegraphics[width=0.99\textwidth]{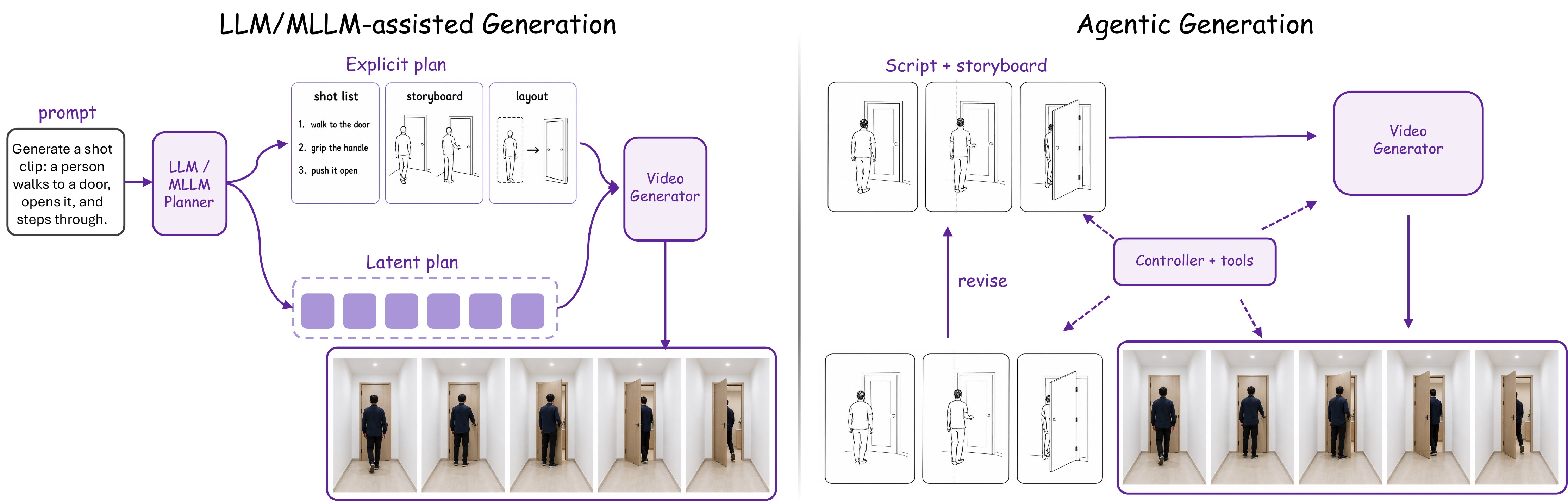}
  \caption{System-level organization for video generation.
  An LLM or MLLM plans for a visual generator through an explicit plan, such as shot lists, storyboards, or layouts, or through learned plan embeddings, and agentic systems close the loop with scripting, generation, evaluation, and revision under a controller.}
  \label{fig:generation-paradigms-llm}
\end{figure}

\subsubsection{LLM/MLLM-Assisted Generation}
LLM/MLLM-assisted generation covers systems in which a language model plans, expands, captions, or structures the conditions for a separate visual sampler.
A language model that predicts video tokens directly belongs to autoregressive generation instead.
In the common arrangement (Figure~\ref{fig:generation-paradigms-llm}), the language model turns an underspecified request into scenes, shots, layouts, or refined prompts that a diffusion, flow, or autoregressive generator consumes.
VideoDirectorGPT~\cite{lin2024videodirectorgpt} produces a grounded video plan before layout-conditioned generation, and LLM-grounded Video Diffusion~\cite{lian2024llmgrounded} and Dysen-VDM~\cite{fei2024dysenvdm} draft dynamic scene layouts or scene graphs that guide the denoiser, so the language model handles composition and timing while the video model handles appearance and motion.
Any2Caption~\cite{wu2025any2caption} interprets text, image, video, region, motion, and camera cues and translates them into dense structured captions for a downstream generator.

The language model and the generator can also be coupled more tightly than through a written prompt.
MetaCanvas~\cite{lin2025metacanvas} lets an MLLM plan in spatial or spatiotemporal latent spaces connected to diffusion backbones, including image-to-video and editing settings.
TV2TV~\cite{han2025tv2tv} couples the two more tightly, interleaving language next-token prediction with video flow matching in one framework.

\subsubsection{Agentic Generation}
\label{sec:writing-agentic}
Agentic generation turns planning into an executable loop: a controller decomposes a request, calls specialized models or tools, inspects intermediate results, and revises the plan or conditions when the output falls short (Figure~\ref{fig:generation-paradigms-llm}).
The recurring stages are scripting, storyboarding, visual generation, evaluation, and revision.

Current agents mainly decompose long-form creation or coordinate tools.
MovieAgent~\cite{wu2025movieagent} assigns screenplay, directing, storyboard, and scene construction to cooperating agents, and Anim-Director~\cite{li2024animdirector} has an MLLM write a storyline and script, produce consistent scene references, and select outputs from generative tools.
Mora~\cite{yuan2024mora} routes among generation and editing skills in a multi-agent workflow.

\subsection{Control Strategies}
\label{sec:writing-control}
\begin{figure}[t]
  \centering
  \includegraphics[width=0.99\textwidth]{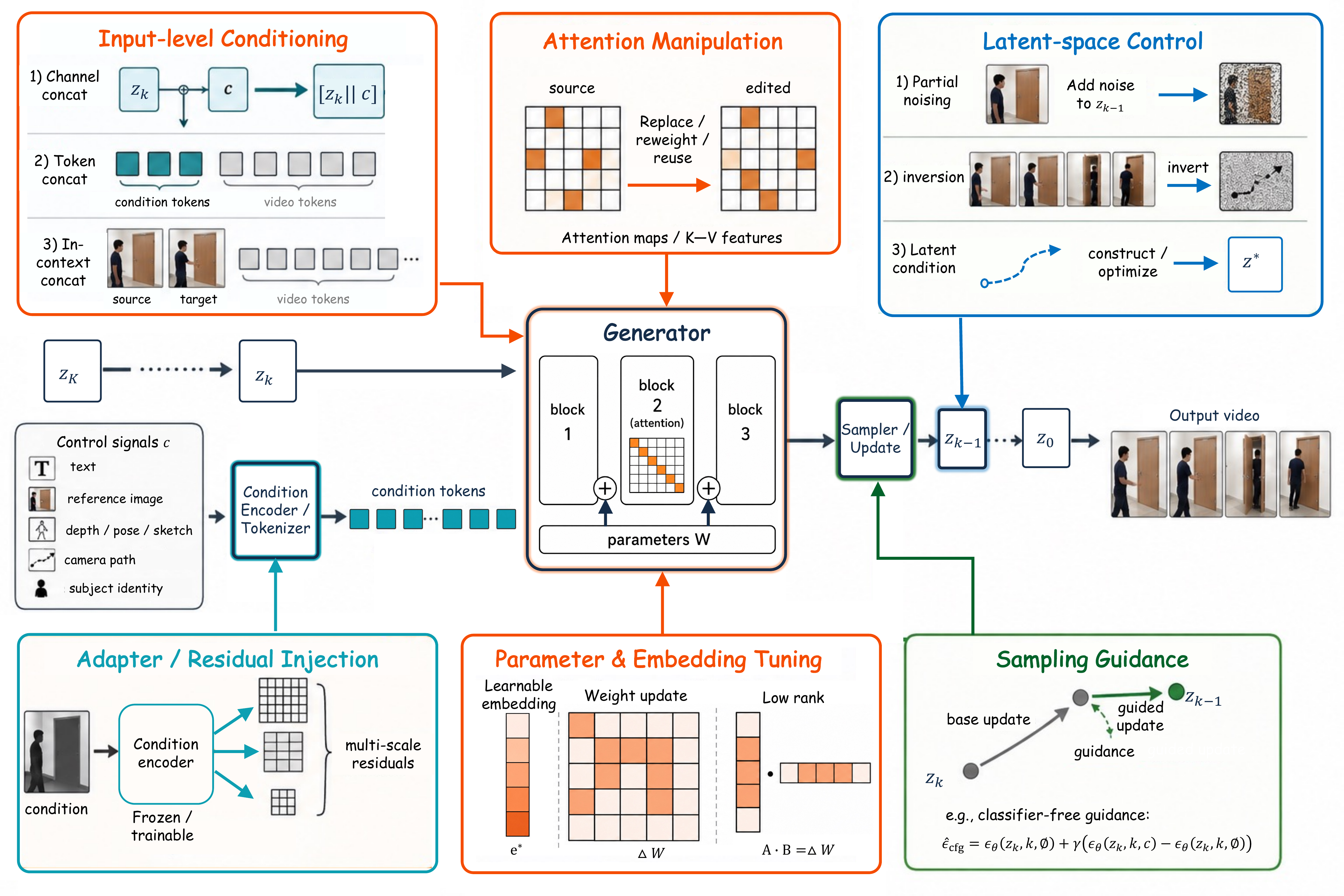}
  \caption{Control mechanisms by intervention site.
  Each route is identified by the component through which its principal control signal enters generation: the input, the attention maps, the latent state, block features, parameters and embeddings, or the sampling update.}
  \label{fig:control-injection}
\end{figure}

Section~\ref{sec:writing-tasks} classifies generation by the information supplied; here we classify control by the component through which that information alters generation.
A camera path, pose sequence, or reference image may enter as input tokens, adapter features, a latent trajectory, sampling guidance, or tuned parameters.
This entry point determines what must be trained, how strong and local the control is, and how easily several signals combine.
Figure~\ref{fig:control-injection} locates the entry points relative to the backbone, and Table~\ref{tab:control-strategies} compares the six families.
A single system often combines several, as in source inversion followed by attention reuse and classifier-free guidance.
\begin{table}[!t]
\centering
\small
\caption{Control strategies for video generation, grouped by where the control signal enters the generator (Figure~\ref{fig:control-injection}). Practical systems often combine several families.}
\label{tab:control-strategies}
\setlength{\tabcolsep}{3pt}
\renewcommand{\arraystretch}{1.10}
\surveytablecolors
\begin{tabularx}{\textwidth}{@{}>{\hsize=.85\hsize}Y >{\hsize=1.00\hsize}Y >{\hsize=.80\hsize}Y >{\hsize=1.05\hsize}Y >{\hsize=1.15\hsize}Y >{\hsize=1.15\hsize}Y@{}}
\toprule
\surveytablehead
\tableheadcell{Family} & \tableheadcell{Where control enters} & \tableheadcell{What is trained} & \tableheadcell{Strengths} & \tableheadcell{Limitations} & \tableheadcell{Examples} \\
\midrule
\textbf{Attention} &
Cross- and self-attention maps, replaced, reweighted, or reused from a source run &
No separate control network &
Local semantic edits that keep the source layout &
Depends on learned correspondences; large motion and occlusion break them; strong map reuse blocks geometric change &
Prompt-to-Prompt~\cite{hertz2022prompttoprompt}, Video-P2P~\cite{liu2023videop2p}, FateZero~\cite{qi2023fatezero}, TokenFlow~\cite{geyer2023tokenflow} \\
\addlinespace
\textbf{Adapter} &
Block features of a frozen backbone, through a side branch &
Adapter or side network &
Keeps the pretrained generator; encodes dense spatial and temporal structure &
Needs paired control data; one branch per signal; separately trained adapters may conflict &
ControlNet~\cite{zhang2023adding}, T2I-Adapter~\cite{mou2023t2iadapter}, MotionCtrl~\cite{wang2024motionctrl}, VideoControlNet~\cite{hu2023videocontrolnet}, VACE~\cite{jiang2025vace} \\
\addlinespace
\textbf{Latent space} &
Noisy latent or denoising trajectory: partial noising, inversion, or constructed and optimized latents &
Nothing, or the base model for constructed conditions &
Noise level sets fidelity against edit strength; accurate inversion preserves background, identity, and motion &
Limited by what the latent retains; inversion errors accumulate into temporal drift &
SDEdit~\cite{meng2021sdedit}, DDIM inversion~\cite{song2021ddim}, Wan-Move~\cite{chu2025wanmove}, VLIPP~\cite{yang2025vlipp} \\
\addlinespace
\textbf{Concatenation / in-context} &
Model input, as extra channels or tokens, including in-context examples and retrieved evidence &
The backbone and its condition encoders &
Simple; many conditions combine in one sequence and one model &
Training must teach the model to bind each token type; longer sequences raise attention cost; retrieved evidence can be wrong &
SVD~\cite{blattmann2023stable}, VideoComposer~\cite{wang2023videocomposer}, FullDiT~\cite{fulldit2025}, UNIC~\cite{unic2025}, PhysRAG~\cite{physrag2026} \\
\addlinespace
\textbf{Guidance} &
Sampling update at each denoising step &
Nothing at test time; CFG needs condition dropout during training &
Works at test time; can impose constraints absent from the training data &
Extra model or gradient evaluations; high scales reduce diversity; only as good as the surrogate objective &
CFG~\cite{ho2022classifierfree}, universal guidance~\cite{bansal2023universalguidance}, FreeDoM~\cite{yu2023freedom}, self-guidance~\cite{epstein2023selfguidance} \\
\addlinespace
\textbf{Tuning} &
Model parameters or text embeddings &
Embeddings, full or low-rank weights, or a hypernetwork that predicts them &
Preserves a specific subject or style more faithfully than prompts &
Needs adaptation data and compute; adds parameters; can bind the subject to reference poses, backgrounds, or motion &
Textual Inversion~\cite{gal2022textualinversion}, DreamBooth~\cite{ruiz2023dreambooth}, LoRA~\cite{hu2021lora}, Tune-A-Video~\cite{wu2023tune}, Video2LoRA~\cite{wu2026video2lora} \\
\bottomrule
\end{tabularx}
\end{table}

\subsubsection{Attention-Based Control}
Attention-based methods reuse the correspondences that a text-conditioned generator already forms: cross-attention links prompt tokens to spatial features, and self-attention relates visual features to one another.
Prompt-to-Prompt~\cite{hertz2022prompttoprompt} injects cross-attention maps from a source prompt into the denoising run for an edited prompt, keeping the layout while changing a word, an attribute, or a concept's strength.

Video editing also needs the change to stay stable across frames.
Video-P2P~\cite{liu2023videop2p} and FateZero~\cite{qi2023fatezero} combine attention control with inversion of the source video.
Video-P2P adds separate guidance for source and target prompts, and FateZero fuses source and target cross- and self-attention maps recorded during inversion.
TokenFlow~\cite{geyer2023tokenflow} instead propagates diffusion features between frames through feature-space correspondences, so an image editor produces temporally consistent edits without training.

\subsubsection{Adapter-Based Control}
Adapters add a trainable path for external structure while keeping most or all of the pretrained generator frozen.
ControlNet~\cite{zhang2023adding} copies the encoder blocks of a frozen image diffusion model into a conditional branch joined by zero-initialized convolutions, whereas T2I-Adapter~\cite{mou2023t2iadapter} adds features from a small condition encoder at several resolutions.
They represent the two main variants: a heavy side network that mirrors the backbone and a light adapter that produces residual features.

Video adapters must also encode how a condition changes over time.
MotionCtrl~\cite{wang2024motionctrl} represents camera motion with poses and object motion with trajectories so that the two can be composed.
VideoControlNet~\cite{hu2023videocontrolnet} brings ControlNet to video-to-video translation with motion guidance that keeps the injected structure aligned across frames.
Unified adapters train several controls together.
VACE~\cite{jiang2025vace} encodes reference frames, source video, and masks in a Video Condition Unit and injects them into a frozen DiT through a context adapter, supporting reference generation, editing, and masked editing with shared adapter parameters.

\subsubsection{Latent-Space Control}
Latent-space methods act on the noisy sample or the denoising trajectory itself.
Partial noising, as in SDEdit~\cite{meng2021sdedit}, perturbs an input to an intermediate noise level and denoises it under a new condition, so the noise level trades fidelity against edit strength.
Inversion maps an observed video to a noise trajectory from which the generator approximately reconstructs it, often through deterministic DDIM sampling~\cite{song2021ddim}; editing then changes the condition or selected latents along the reverse path.

A method may also construct latent conditions directly.
Wan-Move~\cite{chu2025wanmove} projects dense point trajectories into latent space and propagates first-frame features along them, giving an image-to-video model a motion condition without a separate motion encoder.
VLIPP~\cite{yang2025vlipp} has a vision-language model plan coarse object motion, renders the plan as a synthetic motion video, and converts its optical flow into structured noise that initializes a motion-controllable image-to-video model (Section~\ref{sec:wm-physics}).

\subsubsection{Concatenation and In-Context Conditioning}
The most direct route places the condition in the model input.
Channel concatenation appends aligned condition latents to the noisy video latent, as in the image-to-video variants of Stable Video Diffusion~\cite{blattmann2023stable} and CogVideoX~\cite{yang2024cogvideox} and in VideoComposer~\cite{wang2023videocomposer}, which fuses spatial and temporal conditions before concatenation.
Token concatenation instead appends condition embeddings to the sequence processed by self- or joint attention.

FullDiT~\cite{fulldit2025} maps heterogeneous conditions into one sequence and models them with full attention, avoiding separately trained side branches.
In-context generation places reference examples or source--target pairs in the sequence so that the model infers the requested transformation, as Emu2~\cite{sun2024emu2} showed for multimodal image generation and UNIC~\cite{unic2025} applies to unified video editing.
Retrieval can supply further condition tokens: PhysRAG~\cite{physrag2026} retrieves videos of relevant physical phenomena, extracts physical information, and injects it into a video diffusion model through learnable queries.

\subsubsection{Guidance-Based Control}
Guidance changes the sampling direction instead of adding a persistent condition branch.
Classifier-free guidance (CFG)~\cite{ho2022classifierfree} combines conditional and unconditional predictions at each denoising step,
\begin{equation}
  \widehat{\epsilon}_{\mathrm{cfg}}
  = \epsilon_{\theta}(z_k,k,\varnothing)
  + \gamma\bigl(\epsilon_{\theta}(z_k,k,c)
  - \epsilon_{\theta}(z_k,k,\varnothing)\bigr),
\end{equation}
where $z_k$ is the noisy latent at denoising step $k$, $\varnothing$ denotes the null condition, and $\gamma$ is the guidance scale; $\gamma=1$ recovers the conditional prediction.
CFG needs a model trained with condition dropout, and raising $\gamma$ usually strengthens alignment but reduces diversity.

More general guidance uses the gradient of an energy, attention, correspondence, or physical objective to modify each update and can impose constraints absent from the training data.
Universal guidance~\cite{bansal2023universalguidance} and training-free energy guidance~\cite{yu2023freedom} use external objectives, whereas diffusion self-guidance~\cite{epstein2023selfguidance} defines them on the generator's internal attention and feature maps.

\subsubsection{Tuning-Based Control}
Tuning stores a new subject, style, or source-video prior in learned embeddings or weights.
For images, Textual Inversion~\cite{gal2022textualinversion} learns new text embeddings for a user concept, DreamBooth~\cite{ruiz2023dreambooth} fine-tunes the model on a few subject images with a prior-preservation loss, and LoRA~\cite{hu2021lora} learns low-rank weight updates while keeping the base weights fixed.
Video adds the requirement that the learned identity stay stable under motion and viewpoint change.

Tune-A-Video~\cite{wu2023tune} and LoRA-Edit~\cite{gao2025loraedit} tune on each source video.
Tune-A-Video adapts an image diffusion model from one text--video pair with spatiotemporal attention and DDIM inversion, and LoRA-Edit applies mask-aware LoRA tuning to an image-to-video model, taking structure and motion from the source and edited appearance from reference images.
Video2LoRA~\cite{wu2026video2lora} avoids per-reference optimization: a hypernetwork predicts reference-specific LoRA weights in one forward pass.

\subsection{Applications}
\label{sec:writing-apps}
Generated video serves as content, as an avatar, as a visual plan, or as training data, and each use places different demands on conditioning, latency, consistency, and evaluation.
Systems that also model action-conditioned transitions are treated in Section~\ref{sec:interacting}.

\subsubsection{Creative Content Production}
Video generation enters production at several stages, from concept shots and storyboards to narrative footage, advertising and educational variants, and visual-effects edits.
The agentic systems of Section~\ref{sec:writing-agentic} handle planning and tool orchestration, while foundation generators increasingly bundle creation and revision operations.
Following Sora~\cite{brooks2024video}, Movie Gen~\cite{polyak2024moviegen} combines text-to-video generation with personalization, instruction-based editing, and synchronized audio.
Closed systems such as Kling~\cite{kling2024} and Sora~2~\cite{sora2_2025} offer text- and image-to-video generation as products, and Sora~2 adds synchronized audio.
Open models such as VideoCrafter2~\cite{chen2024videocrafter2}, Open-Sora~\cite{zheng2024opensora}, Wan~\cite{wan2025open}, HunyuanVideo~\cite{hunyuanvideo2024}, and Step-Video-T2V~\cite{ma2025stepvideo} make text-to-video generation available through released weights.
Seedance 1.0~\cite{seedance2025} adds native multi-shot generation with consistent subjects, and Seedance 2.0~\cite{seedance2_2026} accepts text, image, audio, and video inputs for joint audio-video generation, reference-based generation, and editing.

Production also needs what per-clip realism does not measure: keeping a character, location, or style across shots, revising one element without disturbing the rest, and controlling camera and motion.
These needs tie personalization and control (Section~\ref{sec:writing-control}) to cross-shot and preference evaluation (Section~\ref{sec:writing-eval}).

\subsubsection{Avatars and Character Animation}
Avatar generation maps speech, motion, or other control signals to the visible behavior of a person or character.
Its lineage runs through the audio-driven systems of Section~\ref{sec:writing-tasks}: Wav2Lip~\cite{prajwal2020wav2lip} edits the mouth region with a lip-sync expert, and SadTalker~\cite{zhang2023sadtalker} predicts expression and head-pose coefficients of a 3D face model.
Diffusion models such as EMO~\cite{tian2024emo} and Hallo~\cite{xu2024hallo} generate expressive portrait motion directly from audio, and VLOGGER~\cite{corona2025vlogger} extends the output to the upper body with hand and gesture motion from a single reference image.
Pose-driven animation, as in Animate Anyone~\cite{hu2024animate} and MagicAnimate~\cite{xu2023magicanimate}, replaces audio with a motion sequence and is judged against that motion instead of speech.
In both settings, identity, clothing, background, and non-speech motion must stay stable over long or varied inputs.

Real-time avatars must also produce frames at a sustained rate with bounded delay while staying synchronized with incoming audio or motion.
MuseTalk~\cite{zhang2024musetalk} dubs video in latent space and reports 30 frames per second at $256\times256$ resolution on a V100 GPU.
Vidu S1~\cite{vidus1_2026} lets voice instructions redirect an ongoing character stream, and Vidu S2~\cite{vidus2_2026} adds references that can be updated during streaming and frame-aligned editing of an incoming video stream.
Mid-stream control adds action-to-output latency to throughput as a criterion and links avatars to the real-time interactive world models of Section~\ref{sec:wm-realtime}.

\subsubsection{Video Plans for Embodied Agents}
Generated video can serve as a plan between a task specification and physical execution, and methods differ mainly in how they recover actions from it.
UniPi~\cite{du2023unipi} generates a future video from a text task and extracts controls with an inverse dynamics model, using images as a planning space shared by embodiments with different action spaces.
AVDC~\cite{ko2024avdc} infers actions from dense correspondences between generated frames, so policies can be learned from videos without action labels.
RoboDreamer~\cite{zhou2024robodreamer} composes video predictions of reusable instruction primitives to generalize to new object--action combinations, and Dreamitate~\cite{liang2024dreamitate} fine-tunes a video model on human demonstrations and tracks the generated tool motion to control a robot.
A generated plan stays fixed, whereas an interactive simulator such as UniSim~\cite{yang2024unisim} updates its predictions as actions arrive (Section~\ref{sec:interacting}).

\subsubsection{Synthetic Data for Embodied AI and Driving}
Embodied and driving systems need observations of rare tasks, environments, and failures that are costly or unsafe to collect.
Video generation can supply them offline by varying appearance, layout, traffic, camera pose, or actions while keeping the annotations carried by the condition.
In driving, MagicDrive~\cite{gao2023magicdrive} conditions multi-camera street views on camera poses, road maps, 3D bounding boxes, and text with cross-view attention, and Panacea~\cite{wen2023panacea} generates annotated multi-view videos from bird's-eye-view layouts.
DriveDreamer~\cite{wang2024drivedreamer} and GAIA-1~\cite{hu2023gaia1} go further and predict future driving observations under scene or action conditions (Section~\ref{sec:interacting}).
In robotics, the data must also carry actions: DreamGen~\cite{jang2025dreamgen} adapts an image-to-video model to a target embodiment and labels its generated trajectories with a latent-action or inverse-dynamics model.
GigaWorld-0~\cite{gigaworld2025} pairs a controllable video generator with 3D generation and system identification to train vision-language-action policies (Section~\ref{sec:interacting}).
Generated data are judged by downstream perception and policy results on real observations, not by visual fidelity.

\begin{insightbox}{Rendering fidelity and stateful control are distinct scaling axes}
Video priors increasingly make each clip convincing, yet visual fidelity and stateful control are separate goals.
A good control changes only its target, keeps that change over time, and combines with other controls without disturbing identity, geometry, or motion.
Progress should therefore be measured by comparing matched rollouts with and without the control, not by the quality of each rollout alone.
\end{insightbox}

\surveytheme{surveyShare}
\section{\shareicon\opsep Unified Video Models: Sharing the World}
\label{sec:sharing}
\suppressfloats[t]
\chapterlead{Sharing means understanding and generation use common representations, parameters, interfaces, or objectives.}
Video understanding and generation have traditionally relied on separate encoders and generators.
Unified video models implement the sharing relation $V\leftrightarrow W$ of Section~\ref{sec:lens}: inferred content can condition generation, and generative structure can support understanding and reasoning.

Definition~\ref{def:operations} evaluates sharing at three levels: capability coverage (one system performs both directions), architectural sharing (the directions use common components), and functional synergy (one direction improves the other under a controlled comparison).
We cover systems with video input or output that link understanding and generation, and cite image-only work where it supplies a design or evaluation protocol missing for video.

\subsection{Unified Architectures}
\label{sec:sharing-arch}
\begin{figure}[!t]
  \centering
  \includegraphics[width=0.99\textwidth]{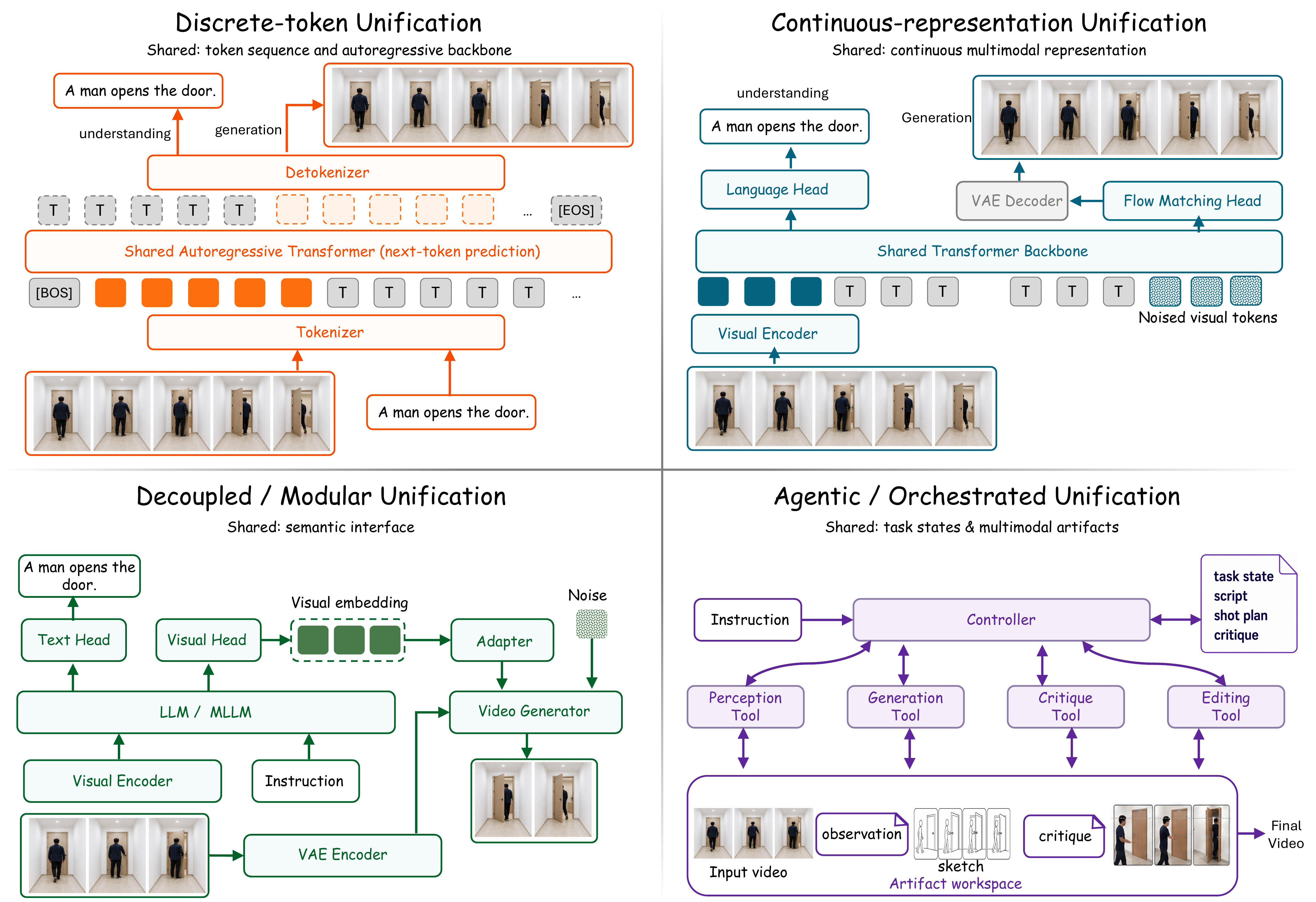}
  \caption{Four architecture families for unified video understanding and generation, ordered by the abstraction of what is shared: discrete-token unification shares a token sequence and autoregressive backbone; continuous latent-space unification shares a continuous representation with direction-specific heads; decoupled and modular unification links separate experts through a semantic bridge; and agentic and tool-augmented unification shares a task state maintained by a controller.
  The families can be combined: an agent may call tools built on any of the other three.}
  \label{fig:unified-architectures}
\end{figure}

Unified architectures differ in where understanding and generation meet (Figure~\ref{fig:unified-architectures}): in a shared token sequence, a shared continuous latent space, a semantic bridge between decoupled experts, or the task state of an agent that calls specialized tools.
Table~\ref{tab:unified-models} compares representative video systems along the component dimensions of Zhao et al.~\cite{zhao2026unifiedmultimodalunderstandinggeneration}.
The first three families have image-domain precedents: Chameleon~\cite{team2024chameleon} for early-fusion token sequences, Transfusion~\cite{zhou2024transfusion} for autoregressive text and diffusion images in one Transformer, Janus~\cite{wu2025janus} for decoupled encoders, and MetaQueries~\cite{pan2025transfer} for learned-query bridges.
On the video side, UniVL~\cite{luo2020univl} earlier unified video--language understanding and caption generation, though its generated output was text, not video.

\begingroup
\footnotesize
\setlength{\tabcolsep}{2.7pt}
\renewcommand{\arraystretch}{1.08}
\setlength{\surveytableunit}{\dimexpr(\textwidth-1.20cm-10\tabcolsep)/5\relax}
\setlength{\LTcapwidth}{\textwidth}
\setlength{\LTpre}{8pt}
\setlength{\LTpost}{8pt}
\Needspace{.35\textheight}
\surveytablecolors
\begin{longtable}{@{}>{\raggedright\arraybackslash}p{1.05\surveytableunit} >{\raggedright\arraybackslash}p{.90\surveytableunit} >{\raggedright\arraybackslash}p{.98\surveytableunit} >{\raggedright\arraybackslash}p{1.03\surveytableunit} >{\raggedright\arraybackslash}p{1.04\surveytableunit} >{\centering\arraybackslash}p{1.20cm}@{}}
\hiderowcolors
\caption{Representative unified video systems, grouped by their integration mechanism. For agentic and tool-augmented systems, the core is a controller and the interfaces are perception, synthesis, or execution tools. Entries follow the cited configurations.}
\label{tab:unified-models}\\
\showrowcolors
\toprule
\surveytablehead
\tableheadcell{System} & \tableheadcell{Core} & \tableheadcell{Read interface} & \tableheadcell{Write / act interface} & \tableheadcell{Training / notes} & \videocaphead \\
\midrule
\endfirsthead
\hiderowcolors
\caption[]{Unified video systems (continued).}\\
\showrowcolors
\toprule
\surveytablehead
\tableheadcell{System} & \tableheadcell{Core} & \tableheadcell{Read interface} & \tableheadcell{Write / act interface} & \tableheadcell{Training / notes} & \videocaphead \\
\midrule
\endhead
\bottomrule
\rowcolor{white}\multicolumn{6}{@{}p{\textwidth}@{}}{\unifiedtablekey}\\
\noalign{\vskip0pt}
\endfoot
\surveylonggroup{6}{Discrete-token unification}
Emu3~\cite{wang2024emu3} & 8B Transformer & Spatiotemporal VQ & VQ $\rightarrow$ decoder & Joint AR $\cdot$ C & \videocaps{\tableyes}{\tableyes}{\tableno} \\
\addlinespace
Emu3.5~\cite{cui2025emu3} & 34B Transformer & IBQ over frames & IBQ $\rightarrow$ frame decoder; keyframe diffusion & AR + SFT + RL $\cdot$ C & \videocaps{\tableyes}{\tableyes}{\tableno} \\
\addlinespace
Video-LaVIT~\cite{jin2024videolavit} & Llama-2-7B & Keyframe + motion VQ & Codes $\rightarrow$ diffusion decoders & Video--language AR $\cdot$ C & \videocaps{\tableyes}{\tableyes}{\tableno} \\
\addlinespace
Vega~\cite{vega2026} & Qwen2.5-3B & Tar semantic tokens & Tar $\rightarrow$ Wan2.1 renderer & Semantic AR + FM $\cdot$ C & \videocaps{\tableyes}{\tableyes}{\tableno} \\
\addlinespace
VILA-U~\cite{wu2025vilauunifiedfoundationmodel} & Llama-2-7B + RQ depth & SigLIP/RQ tower & RQ $\rightarrow$ RQ-VAE & Alignment + joint AR $\cdot$ C & \videocaps{\tableyes}{\tableyes}{\tableno} \\
\addlinespace
NExT-OMNI~\cite{luo2025nextomni} & Qwen2.5-7B & CLIP discrete tokens & Codes $\rightarrow$ VQ-VAE & Joint discrete FM $\cdot$ B & \videocaps{\tableyes}{\tableyes}{\tableno} \\
\surveylonggroup{6}{Continuous latent-space unification}
TUNA~\cite{liu2025tuna} & Qwen2.5-1.5B + flow & Wan2.2 VAE + SigLIP2 & Shared $\rightarrow$ flow/VAE & Joint AR + FM $\cdot$ H & \videocaps{\tableyes}{\tableyes}{\tableno} \\
\addlinespace
Show-o2~\cite{xie2025show} & Qwen2.5-1.5B + flow & Wan2.1 VAE & Shared $\rightarrow$ flow/VAE & Staged AR/FM $\cdot$ H & \videocaps{\tableyes}{\tableyes}{\tableno} \\
\addlinespace
Uni-ViGU~\cite{qin2026univigu} & Wan2.1 DiT & Wan VAE + diffusion features & VAE $\rightarrow$ video flow/VAE & Continuous/discrete FM; shared attention & \videocaps{\tableyes}{\tableyes}{\tableno} \\
\addlinespace
HYDRA-X~\cite{hydrax2026} & Qwen2.5-7B + flow & Hydra-XTok & XTok $\rightarrow$ visual flow & Joint AR/FM; mask NR & \videocaps{\tableyes}{\tableyes}{\tableno} \\
\addlinespace
Gen4U~\cite{gen4u2026} & Veo 3~\cite{veo3_2025} / Wan2.2 & Video codec + diffusion features & Original codec $\rightarrow$ decoder & Frozen-feature readouts & \videocaps{\tableyes}{\tablepartial}{\tableno} \\
\addlinespace
PixelUMM~\cite{wei2026pixelumm} & Qwen3 Transformer experts & Pixel patch/tubelet & Pixels $\rightarrow$ pixel flow & Joint AR/FM; edit fine-tuning $\cdot$ H & \videocaps{\tableyes}{\tableyes}{\tableyes} \\
\surveylonggroup{6}{Decoupled and modular unification}
Lance~\cite{fu2026lance} & Qwen2.5-VL-3B dual experts & Qwen2.5-VL ViT & Wan2.2 VAE $\rightarrow$ flow & Joint AR/FM $\cdot$ H & \videocaps{\tableyes}{\tableyes}{\tableyes} \\
\addlinespace
NExT-GPT~\cite{wu2023nextgpt} & Vicuna-7B + projectors & ImageBind & Tokens $\rightarrow$ Zeroscope & Alignment + LoRA $\cdot$ C & \videocaps{\tableyes}{\tableyes}{\tableno} \\
\addlinespace
GPT4Video~\cite{wang2023gpt4video} & LLaMA + LoRA & CLIP + video abstractor & Text $\rightarrow$ external T2V & Alignment; generator frozen $\cdot$ C & \videocaps{\tableyes}{\tableyes}{\tableno} \\
\addlinespace
Omni-Video~\cite{tan2025omnivideo} & VILA + clue adapter & VILA + SigLIP2 clues & Wan VAE $\rightarrow$ Wan2.1 & Staged clue/decoder training $\cdot$ C & \videocaps{\tableyes}{\tableyes}{\tableyes} \\
\addlinespace
UniVid~\cite{luo2025univid} & BAGEL-7B + adapter & ViT/SigLIP2 selector & Wan VAE $\rightarrow$ Wan2.2 & Connector/DiT LoRA; reflection & \videocaps{\tableyes}{\tableyes}{\tableno} \\
\addlinespace
UniVideo~\cite{univideo2025} & Qwen2.5-VL-7B + bridge & Qwen2.5-VL & Hunyuan VAE $\rightarrow$ MMDiT & Staged FM; MLLM frozen & \videocaps{\tablepartial}{\tableyes}{\tableyes} \\
\addlinespace
Omni-Video 2~\cite{yang2026omnivideo2} & Frozen MLLM + adapter & MLLM encoder & VAE $\rightarrow$ video DiT & Adapter/decoder FM & \videocaps{\tablepartial}{\tableyes}{\tableyes} \\
\addlinespace
MoRoute~\cite{gao2026moroute} & Qwen3.5 + routers & Qwen3.5 encoder & Wan VAE $\rightarrow$ Wan2.1 & Interface/decoder FM; routed attention & \videocaps{\tablepartial}{\tableyes}{\tableyes} \\
\addlinespace
ReViSE~\cite{liu2025revise} & Omni-Video + VLM critic & Inherited VLM encoder & Wan VAE $\rightarrow$ Wan2.1 & FM + differentiable VLM reward & \videocaps{\tablepartial}{\tableyes}{\tableyes} \\
\surveylonggroup{6}{Agentic and tool-augmented unification}
UniVA~\cite{liang2025univa} & Claude Sonnet 4 agents & InternVL3; SAM2/Sa2VA & Seedance; VACE/Aleph; audio/image & Orchestration; artifact memory & \videocaps{\tableyes}{\tableyes}{\tableyes} \\
\addlinespace
VideoAgent (editing)~\cite{zhou2026videoagentediting} & Agent graph; tested LLMs & Shot planning; retrieval & Video/audio editing tools & Textual-gradient graph refinement & \videocaps{\tableyes}{\tableno}{\tableyes} \\
\addlinespace
VISTA~\cite{long2025vista} & Gemini 2.5 Flash agents & Visual/audio critique & Veo 3~\cite{veo3_2025}, Veo 2 & Prompt refinement; candidate state & \videocaps{\tablepartial}{\tableyes}{\tableno} \\
\addlinespace
VideoAgent (planning)~\cite{soni2024videoagent} & GPT-4 Turbo feedback & VLM plan judgment & AVDC diffusion; action execution & Self-conditioning; online tuning & \videocaps{\tablepartial}{\tableyes}{\tableno} \\
\end{longtable}
\endgroup

\subsubsection{Discrete-Token Unification}
Discrete-token models convert text, images, and video into symbols that are interleaved in one sequence.
A single Transformer then serves both directions with next-token prediction: answering predicts text tokens after video context, and generation predicts visual tokens after a text or visual condition.
Emu3~\cite{wang2024emu3} follows this design without diffusion heads, covering image and video understanding and generation, and Emu3.5~\cite{cui2025emu3} scales the same objective to video-heavy data and reports world-modeling behavior such as spatiotemporally consistent exploration.
Video-LaVIT~\cite{jin2024videolavit} cuts the token cost of video by tokenizing keyframes and motion separately, and VILA-U~\cite{wu2025vilauunifiedfoundationmodel} trains its vision tower to align discrete tokens with language while keeping the detail needed for generation.
SEED-LLaMA~\cite{ge2024seedllama} pairs discrete visual tokens with a diffusion image decoder; it answers video questions but generates only images.
Two variants change the role of the tokens or the objective: Vega~\cite{vega2026} predicts semantic keyframe tokens autoregressively and leaves appearance to a diffusion renderer, and NExT-OMNI~\cite{luo2025nextomni} replaces autoregression with discrete flow matching over text, image, video, and audio tokens.

This route gives many input--output combinations one vocabulary, one sequence format, and one likelihood objective.
However, the tokenizer becomes the bottleneck: codes must preserve appearance and motion without making sequences unmanageably long, and quantization discards information before the Transformer sees it.
Long autoregressive outputs also accumulate errors through their generated prefix.

\subsubsection{Continuous Latent-Space Unification}
Continuous latent-space models avoid quantization: an encoder maps images or video into a continuous feature space that an understanding head reads and a decoder, diffusion model, or flow head turns back into pixels.
The latent space is shared, but the losses and output heads can differ, because language prediction and pixel generation have different numerical requirements.
TUNA~\cite{liu2025tuna} cascades a VAE encoder with a representation encoder, and Show-o2~\cite{xie2025show} replaces the discrete vocabulary of Show-o~\cite{xie2024show} with a 3D causal VAE; both predict text autoregressively and generate images and video with flow matching.
Uni-ViGU~\cite{qin2026univigu} starts from the other end, extending a video diffusion generator with discrete flow matching for text and lightweight understanding layers.
AToken~\cite{lu2025atoken} and HYDRA-X~\cite{hydrax2026} make the tokenizer itself the shared component, training it with reconstruction and semantic objectives over images, video, and, for AToken, 3D.
PixelUMM~\cite{wei2026pixelumm} drops the pretrained encoder and VAE: pixel patches and video tubelets enter a mixture-of-transformers in which the two directions keep separate projections and feed-forward layers but share attention, supporting video question answering, generation, and editing.

Continuous spaces fit diffusion and flow generators naturally, but they expose a tension that recurs across unified models.
Understanding benefits from invariance to texture and small appearance changes, whereas generation must retain enough detail to reconstruct them.

\subsubsection{Decoupled and Modular Unification}
Decoupled systems sidestep this tension by keeping separate visual pathways for the two directions, linked through an LLM, a projection layer, or a learned adapter.
The understanding encoder can drop nuisance detail and emphasize objects, relations, actions, or text, while a separate tokenizer or decoder keeps the fine spatial and temporal detail needed to generate video.
NExT-GPT~\cite{wu2023nextgpt} and GPT4Video~\cite{wang2023gpt4video} set the modular baseline around an instruction-following LLM and frozen pretrained encoders and generators, so only a small part of the system is trained: NExT-GPT links the LLM to diffusion decoders through lightweight projections, whereas GPT4Video has the LLM write textual prompts for an off-the-shelf text-to-video model.
Later systems train the bridge.
Omni-Video~\cite{tan2025omnivideo}, UniVid~\cite{luo2025univid}, and UniVideo~\cite{univideo2025} let an MLLM interpret the instruction and pass continuous features through an adapter or bridge to a diffusion decoder; UniVid adds a reflection mechanism for temporal reasoning.
Omni-Video 2~\cite{yang2026omnivideo2} keeps the MLLM frozen and conditions the decoder on a predicted target caption and source-video latents for editing, and ReViSE~\cite{liu2025revise} trains an Omni-Video model with feedback from its own VLM (Section~\ref{sec:sharing-synergy}).

Other systems redesign the bridge itself.
Lumos-Nexus~\cite{xing2026lumosnexus} aligns a small generator with the understanding model and mixes its velocity predictions with those of a larger pretrained generator at inference.
MoRoute~\cite{gao2026moroute} learns which layer of a frozen VLM conditions each diffusion block, and Mamoda2.5~\cite{shi2026mamoda25} feeds semantic features to a sparse-expert diffusion backbone through learned queries.
Lance~\cite{fu2026lance} replaces the bridge with shared attention: separate semantic and VAE encoders feed dual experts that attend over one interleaved sequence.
JavisGPT~\cite{javisgpt2025} extends the encoder--LLM--decoder pattern to sounding video, passing synchrony-aware queries to a joint audio--video generator.

Decoupling keeps strong pretrained components and makes each one replaceable.
However, the bridge can become an information bottleneck: a semantic clue sufficient for following an instruction may omit the appearance, geometry, or motion that the decoder needs.

\subsubsection{Agentic and Tool-Augmented Unification}
Agentic systems leave perception, generation, editing, and evaluation to specialized models.
A controller reads the request and media, plans operations, calls tools, inspects their outputs, and revises the plan; what is shared is a task state such as a script, shot plan, prompt, edit graph, or critique.
VideoAgent (editing)~\cite{zhou2026videoagentediting} plans shots and retrieves source clips before calling a library of editing agents, and VISTA~\cite{long2025vista} wraps a video generator in a loop of temporal planning, candidate selection, multimodal critique, and prompt rewriting.
UniVA~\cite{liang2025univa} plans and calls understanding, generation, editing, and segmentation tools that share global, task, and user memories.

Agents can combine strong tools without forcing their internal representations to agree, and each tool can be swapped as better models appear.
However, they depend on tool selection, intermediate formats, and evaluator reliability: one wrong caption, retrieval result, or critique can redirect the whole workflow.

\begin{figure}[t]
  \centering
  \includegraphics[width=0.99\textwidth]{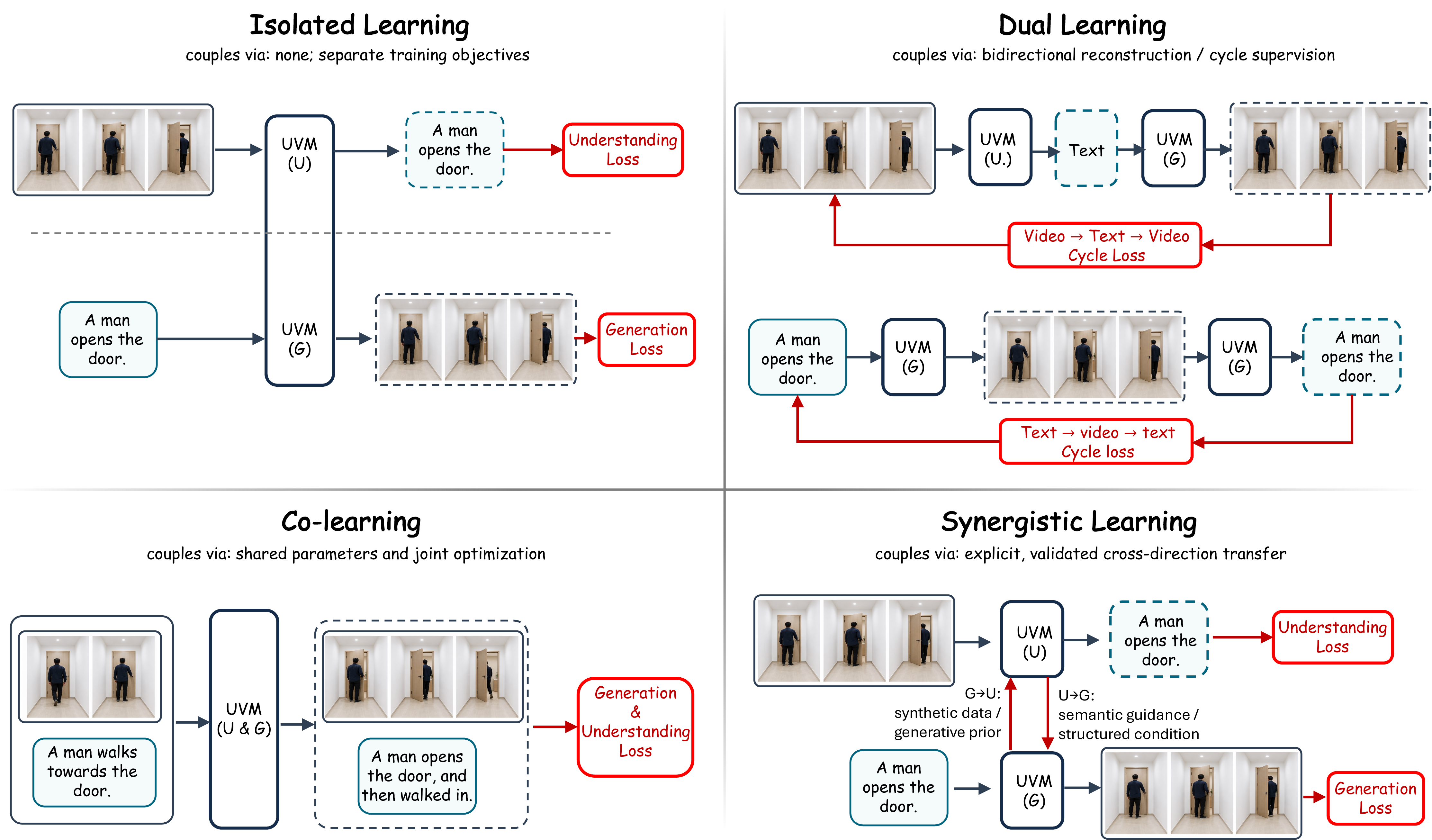}
  \caption{Learning paradigms in unified video models.
  Isolated learning keeps independently pretrained experts; dual learning relates inverse tasks or reconstruction cycles; co-learning optimizes both directions jointly; and synergistic learning targets cross-direction benefit.
  The paradigms can coexist.}
  \label{fig:unified-learning}
\end{figure}

\subsection{Learning Paradigms}
\label{sec:sharing-learning}
Architecture fixes where understanding and generation meet; training determines what signal flows between them (Figure~\ref{fig:unified-learning}).
Isolated learning connects independently trained experts, dual learning relates the two directions through inverse or cycle objectives, co-learning optimizes both over shared parameters or representations, and synergistic learning uses one direction to improve the other.
The paradigms can coexist; for example, a cycle objective can be part of joint training.

\subsubsection{Isolated Learning}
Isolated learning trains understanding and generation as independent experts and connects them after pretraining through a small trained module, while most expert parameters stay frozen.
NExT-GPT~\cite{wu2023nextgpt} and GPT4Video~\cite{wang2023gpt4video} follow this recipe (Section~\ref{sec:sharing-arch}), as GILL~\cite{koh2023gill} did for a frozen language model and a frozen image generator.
Isolated learning is cheap and modular, and the source of each capability stays easy to identify.
However, generation losses never update the video encoder and understanding losses never update the decoder, so no cross-direction transfer can occur, which makes isolated systems the natural baseline for measuring what tighter coupling adds.

\subsubsection{Dual Learning}
\label{sec:sharing-dual}
Dual learning treats understanding and generation as approximately inverse problems: a model reconstructs a condition from generated content, regenerates an observation from an inferred representation, or enforces agreement over a round trip.
Such objectives let one direction supervise the other when paired labels are scarce.
As many videos fit one description and one condition allows many valid futures, a video cycle must preserve task-relevant events, identities, viewpoints, and motion while tolerating appearance variation.

Uni-ViGU~\cite{qin2026univigu} is the clearest video-native example.
Starting from a video generator, it recovers understanding as the inverse task: it first reconstructs prompts to recall learned text--video associations and then trains on detailed captions, keeping the original generation path.
Dedicated cycle objectives remain image-based: CyCLeGen~\cite{shan2026cyclegen} couples image--layout cycles, UniCorn~\cite{han2026unicorn} distills text--image--text self-play among proposer, solver, and judge roles, and Reconstruction Alignment~\cite{xie2026reconstruction} trains a model to reconstruct an image from its own understanding embeddings.

\subsubsection{Co-learning}
Co-learning optimizes understanding and generation objectives in one shared model or a jointly trained bridge, and each of the first three families of Section~\ref{sec:sharing-arch} supports it.
Discrete-token models such as Emu3~\cite{wang2024emu3}, Video-LaVIT~\cite{jin2024videolavit}, and VILA-U~\cite{wu2025vilauunifiedfoundationmodel} apply one next-token objective to both directions, while Vega~\cite{vega2026} shapes a shared semantic token space and keeps a separate diffusion renderer.
Continuous and hybrid models share a visual space or Transformer but keep distinct losses, as in TUNA~\cite{liu2025tuna}, Show-o2~\cite{xie2025show}, HYDRA-X~\cite{hydrax2026}, PixelUMM~\cite{wei2026pixelumm}, and the single-Transformer HaploOmni~\cite{xiao2025haploomni}.
BAGEL~\cite{deng2505emerging} co-trains a mixture of transformer experts that share attention but keep direction-specific projections and feed-forward layers, a design Lance applies to video (Section~\ref{sec:sharing-arch}).
Bridge models such as Omni-Video~\cite{tan2025omnivideo}, UniVid~\cite{luo2025univid}, and UniVideo~\cite{univideo2025} co-train the adapter between an MLLM and a video decoder.

Joint training creates an optimization imbalance between the two objectives.
Understanding data rewards semantic invariance and concise answers, whereas generation data rewards retaining appearance and motion detail.
Differences in dataset scale, loss magnitude, and convergence speed can let one direction dominate the shared parameters, so the data mixture and schedule decide how strongly each capability updates the model.
Lance~\cite{fu2026lance} counters this interference with staged multitask training and adaptive data scheduling.
Tong et al.~\cite{tong2026multimodalpretraining} study related choices when pretraining a Transfusion-style backbone from scratch, comparing representation autoencoders, data mixtures, and dense or sparse-expert architectures with compute-matched image--text analyses.

\subsubsection{Synergistic Learning}
\label{sec:sharing-synergy}
Synergistic learning uses one direction to improve the other; its evidence is a measured gain in the receiving direction.
From understanding to generation, OmniWeaving~\cite{pan2026omniweaving} conditions a video generator on an MLLM that can first write a reasoning trace about interleaved text, image, and video inputs; after end-to-end training, enabling this thinking mode raises its scores on intent-driven and interpolative tasks of IntelligentVBench.
ReViSE~\cite{liu2025revise} turns understanding into a training signal for editing: the frozen VLM inside an Omni-Video model scores a one-step estimate of the edited video, and its answer-token probabilities serve as a differentiable reward alongside flow matching.
Ablations against supervised fine-tuning on RVE-Bench support this understanding-to-editing transfer.
Lance's data-mixture ablations~\cite{fu2026lance} give video evidence in both directions: adding understanding data improves generation on VBench, and adding generation data improves understanding on MVBench.

From generation to understanding, Uni-ViGU~\cite{qin2026univigu} transfers a generator's text--video associations through inverse-task training (Section~\ref{sec:sharing-dual}), and Thinking with Video~\cite{tong2026thinkingvideovideogeneration} generates intermediate frames at inference (Section~\ref{sec:sharing-reasoning}).
CounterVid~\cite{poppi2026countervid} transfers through data: it synthesizes counterfactual action clips and trains a video-language model with a preference objective, reducing action and temporal hallucinations more than generic visual negatives do.
UnityVideo~\cite{huang2025unityvideounifiedmultimodalmultitask} jointly learns depth, flow, segmentation, pose, and other visual conditions, improving both world-aware generation and the prediction of those signals.
Gen4U~\cite{gen4u2026} and GenCeption~\cite{genception2026} reuse frozen or adapted video diffusion features for perception (Section~\ref{sec:reading-representation}).

Integrated tasks provide a third kind of evidence.
UniUGP~\cite{lu2025uniugp} tests whether scene reasoning and predicted video improve driving trajectories within one system (Section~\ref{sec:sharing-planning}).
EWAM~\cite{wang2026ewam} shows that masking or counterfactually replacing predicted-future features changes a manipulation policy's actions and success.
Tong et al.~\cite{tong2026multimodalpretraining} also find that general multimodal pretraining data, especially raw video, helps offline navigation planning through action-conditioned prediction.
Apart from Lance's data-mixture ablations, controlled bidirectional tests remain image-based (Section~\ref{sec:sharing-eval}).

\subsection{Applications}
\label{sec:sharing-apps}
Applications test whether information passed between understanding and generation improves a task outcome.

\subsubsection{Instruction-Driven Video Editing and Creation}
Instruction-driven editing follows a read--modify--write pattern: the model identifies the entities, events, motion, and time span an instruction refers to, then changes them while preserving everything else.
A plausible output is still wrong if it edits the wrong interval, loses subject identity, or damages surrounding content.
Unified models place both steps in one system: Omni-Video~\cite{tan2025omnivideo} and UniVideo~\cite{univideo2025} pass MLLM interpretations to a diffusion decoder (Section~\ref{sec:sharing-arch}), and UNIC~\cite{unic2025} casts editing as in-context generation over one shared sequence.
ReViSE~\cite{liu2025revise} adds reasoning to editing data: it pairs reasoning-enriched instructions with target edits and separates modifying a clip from generating a plausible continuation.

Agentic systems extend editing into production workflows: VideoAgent (editing)~\cite{zhou2026videoagentediting} plans and assembles multi-shot edits, and VISTA~\cite{long2025vista} uses its own critiques to rewrite prompts across rounds (Section~\ref{sec:sharing-arch}).
Evaluation should therefore score localization, preservation of untouched content, and improvement across rounds, not only the quality of the final video.

\subsubsection{Generative Reasoning}
\label{sec:sharing-reasoning}
Generative reasoning uses generation as an intermediate computation.
A video can show a changing configuration that is hard to express in one image or a textual chain of thought, as objects move, occlusions change, and the consequences of an operation unfold.
The generated sequence then serves as an inspectable hypothesis about the process.
Thinking with Video~\cite{tong2026thinkingvideovideogeneration} uses a video generator to produce intermediate frames during multimodal reasoning and evaluates them on VideoThinkBench.
OpenCoF~\cite{opencof2026} trains chain-of-frame generation on temporally organized reasoning videos, while MME-CoF-Pro~\cite{mmecofpro2026,luo2026evaluation} shows that answer accuracy can rise while intermediate frames stay inconsistent or hallucinated.
ProAR~\cite{shen2026proar} makes autoregressive video diffusion prospective by predicting its own goal frame alongside each chunk.

However, generated reasoning traces add a failure mode: a coherent but unsupported intermediate video can make a later answer look well grounded.
Reliable use thus needs observed evidence separated from generated hypotheses, and tests of whether the intermediate video changes the answer for the right reason.

\subsubsection{Unified Perception-Generation-Planning Systems}
\label{sec:sharing-planning}
In planning, generated video predicts what an action will do: the system reads the current observation, predicts how the scene may evolve under a plan, and uses that prediction to choose or refine the action.
Its value lies in the resulting decision, not in visual quality.
UniUGP~\cite{lu2025uniugp} applies this loop to autonomous driving: from multi-frame observations and instructions, its hybrid experts produce scene reasoning, future video, and a trajectory, connected through a four-stage curriculum.
The question is whether future generation improves planning in difficult or long-tail situations.
VideoAgent (planning)~\cite{soni2024videoagent} applies the same idea to embodied tasks, refining generated video plans through self-conditioning and execution feedback.

Recent models extend sharing to actions.
Cosmos 3~\cite{cosmos3_2026} places language, image, video, audio, and action sequences in one omnimodal mixture-of-transformers.
UniWAM~\cite{chen2026uniwam} couples a physical reasoner, a video generator, and an action predictor through joint attention, and perturbs predicted future latents during post-training so that actions do not over-rely on imagined futures.
EWAM~\cite{wang2026ewam} tests whether a predicted-future pathway matters for action (Section~\ref{sec:sharing-synergy}), and Section~\ref{sec:interacting} details action-conditioned prediction.

\subsection{Evaluation}
\label{sec:sharing-eval}

Table~\ref{tab:unified-eval} expands the sharing row of Table~\ref{tab:evidence-ladder} into evaluation levels: separate competence, integrated behavior, and measured synergy.
Video coverage is mature at the first level, emerging at the second, and mostly directional at the third.
Levels~1 and~2 refine the capability-coverage level of Definition~\ref{def:operations} and Level~3 its functional-synergy level; architectural sharing is a property of the design (Section~\ref{sec:sharing-arch}), not a benchmark score, and decision utility is covered by the integrated tasks of Section~\ref{sec:sharing-synergy}.

\begin{table}[!t]
\centering
\small
\caption{Evidence ladder for unified video models, from separate competence to measured cross-direction transfer.}
\label{tab:unified-eval}
\setlength{\tabcolsep}{3pt}
\surveytablecolors
\begin{tabularx}{\textwidth}{@{}>{\raggedright\arraybackslash}p{2.70cm} >{\hsize=.83\hsize}Y >{\hsize=1.47\hsize}Y >{\hsize=.70\hsize}Y@{}}
\toprule
\surveytablehead
\tableheadcell{Evidence level} & \tableheadcell{Required test} &
\tableheadcell{Benchmarks and probes} & \tableheadcell{Video-native status} \\
\midrule
\surveybadge{Level 1}\par Separate competence &
evaluate understanding and generation separately &
Video-MME~\cite{fu2025video}, MVBench~\cite{li2024mvbench}, LongVideoBench~\cite{wu2024longvideobench}, LVBench~\cite{wang2025lvbench}, MLVU~\cite{zhou2025mlvu}; VBench~\cite{huang2023vbenchcomprehensivebenchmarksuite}, VBench-2.0~\cite{zheng2025vbench20advancingvideogeneration} &
well established \\
\addlinespace
\surveybadge{Level 2}\par Integrated behavior &
test integrated transformations and source/intent fidelity &
UniVBench~\cite{wei2026univbenchunifiedevaluationvideo}, UniM~\cite{li2026unimunifiedanytoanyinterleaved}; agentic UniVA-Bench~\cite{liang2025univa}; reasoning-conditioned RVE-Bench~\cite{liu2025revise} and VR-Bench~\cite{xing2026lumosnexus} &
emerging video coverage \\
\addlinespace
\surveybadge{Level 3}\par Measured synergy &
ablate cross-direction information or training under matched settings &
\emph{Video controls:} Lance task mixtures~\cite{fu2026lance}, ReViSE reward~\cite{liu2025revise}, CounterVid preferences~\cite{poppi2026countervid}, UnifiedReward critic transfer~\cite{wang2026unifiedreward}, EWAM future-feature interventions~\cite{wang2026ewam}. \emph{Image--text:} RealUnify~\cite{shi2026realunifyunifiedmodelstruly}, UniSandbox~\cite{niu2025understandinginformgenerationunified} &
directional video evidence; general bidirectional tests remain limited \\
\addlinespace
\gaphl \surveybadge{Open gap}\par Video round trip &
specify invariants, block answer leakage, and compare cycle / non-cycle controls &
\emph{image-based precedents:} UniCorn~\cite{han2026unicorn}, CyCLeGen~\cite{shan2026cyclegen} &
no established video-native standard \\
\bottomrule
\end{tabularx}
\end{table}

\subsubsection{Per-Direction Evaluation}
Each direction is evaluated with its own community's protocols: MVBench~\cite{li2024mvbench}, Video-MME~\cite{fu2025video}, LongVideoBench~\cite{wu2024longvideobench}, LVBench~\cite{wang2025lvbench}, and MLVU~\cite{zhou2025mlvu} for understanding (Section~\ref{sec:reading-tasks}), and VBench~\cite{huang2023vbenchcomprehensivebenchmarksuite} and VBench-2.0~\cite{zheng2025vbench20advancingvideogeneration} for generation (Section~\ref{sec:writing-eval}).
Results should be reported separately, since perceptual and faithfulness criteria are not interchangeable with question-answer accuracy.
Matched single-task baselines can also reveal interference, where joint training improves one direction but degrades the other.

\subsubsection{Unified, Interleaved, and Synergy Evaluation}
Integrated benchmarks test whether one system handles a wider family of input--output transformations.
UniVBench~\cite{wei2026univbenchunifiedevaluationvideo} is the most direct video-centered example: it evaluates understanding, generation, editing, and reconstruction on multi-shot videos with a shared agentic evaluator, and its reconstruction and editing tasks check whether source content survives.
RVE-Bench~\cite{liu2025revise} and VR-Bench~\cite{xing2026lumosnexus} test whether reasoning-conditioned edits and generations do what the inferred intent implies, and UniVA-Bench~\cite{liang2025univa} scores agentic workflows on both final media and planning behavior.
UniM~\cite{li2026unimunifiedanytoanyinterleaved} extends input and output to text, image, video, and audio, while MME-Unify~\cite{xie2025mmeunify} covers integrated tasks in image--text settings.

Synergy evaluation instead compares matched routes, such as understanding-guided generation against plain generation, or understanding with and without generated intermediates.
Image-domain studies lead here: DreamLLM~\cite{dong2023dreamllm} was an early study of synergy between understanding and generation in interleaved documents, and Uni-MMMU~\cite{zou2026unimmmumassivemultidisciplinemultimodal}, RealUnify~\cite{shi2026realunifyunifiedmodelstruly}, and UniSandbox~\cite{niu2025understandinginformgenerationunified} score both directions with coupled tasks, matched routes, or decoupled tracks.
UniG2U-Bench~\cite{wen2026unig2ubenchunifiedmodelsadvance} finds that an extra generation step helps understanding on tasks that need visual transformation but can be neutral or harmful elsewhere.
Pseudo-Unification~\cite{yang2026pseudounificationentropyprobingreveals} probes internal information patterns and finds divergent processing behind nominally shared parameters.
Video evidence so far comes from training ablations and interventions and is mostly directional (Section~\ref{sec:sharing-synergy}); Section~\ref{sec:challenge-synergy} discusses what a video synergy test would need.

\subsubsection{Protocols and Consistency}
Round-trip evaluation tests whether content survives an understanding--generation cycle: a model infers an event or layout, renders it, and reads it back, or generates a video, describes it, and regenerates it.
Existing round-trip training objectives are image-based (Section~\ref{sec:sharing-dual}), and the reconstruction task of UniVBench is the closest video protocol.
A video protocol must name entities, event order, geometry, or dynamics to keep, score intended changes separately, and add non-unified and no-cycle controls.

Reliable measurement also needs information controls.
Train--test overlap can confound transfer with memorization, intermediate representations can leak answers across directions, and text accuracy, video fidelity, and cross-modal consistency may rely on different, possibly biased judges.
UniSandbox~\cite{niu2025understandinginformgenerationunified} reduces leakage with synthetic tasks and separated modules, TorchUMM~\cite{luo2026torchummunifiedmultimodalmodel} standardizes model interfaces and task implementations, and UniSAFE~\cite{lee2026unisafecomprehensivebenchmarksafety} evaluates safety across input--output modality combinations.
Judges can themselves be unified: UnifiedReward~\cite{wang2026unifiedreward} trains one reward model to rank and score image and video outputs for both understanding and generation.

\begin{insightbox}{Synergy requires controlled cross-direction transfer}
Shared tokens or parameters show coupling, separate scores show competence, and mixed input--output tasks show interoperability.
Synergy needs a controlled comparison: ablate the cross-direction path under matched data, capacity, and compute, and measure the change in generation, understanding, or an integrated task.
\end{insightbox}

\surveytheme{surveyInteract}
\section{\interacticon\opsep Video World Models: Interacting with the World}
\label{sec:interacting}
\chapterlead{Interacting means modeling state transition itself: how an established world evolves over time and, when actions, controls, or new observations are supplied, how they change that evolution.}

Interacting is the operation $W\circlearrowleft$ of Section~\ref{sec:lens}: the model advances its state, $s_{t+1}\sim p_\theta(s_{t+1}\mid s_{\leq t},a_t,c_t)$, and decodes observations when visible output is needed.
Rollout exposes failures that a single plausible clip hides: generated errors feed back as context, off-screen content is forgotten, smooth motion breaks physical or causal structure, and actions miss their intended effects.
We first place current models in the taxonomy of Section~\ref{sec:wm-taxonomy}, then examine control, memory, responsiveness, grounding, applications, and evaluation.
Zhu et al.~\cite{zhu2024soraworldsimulator} survey general world models starting from Sora; this chapter focuses on video-centered state.

\begin{figure}[t]
  \centering
  \includegraphics[width=0.99\textwidth]{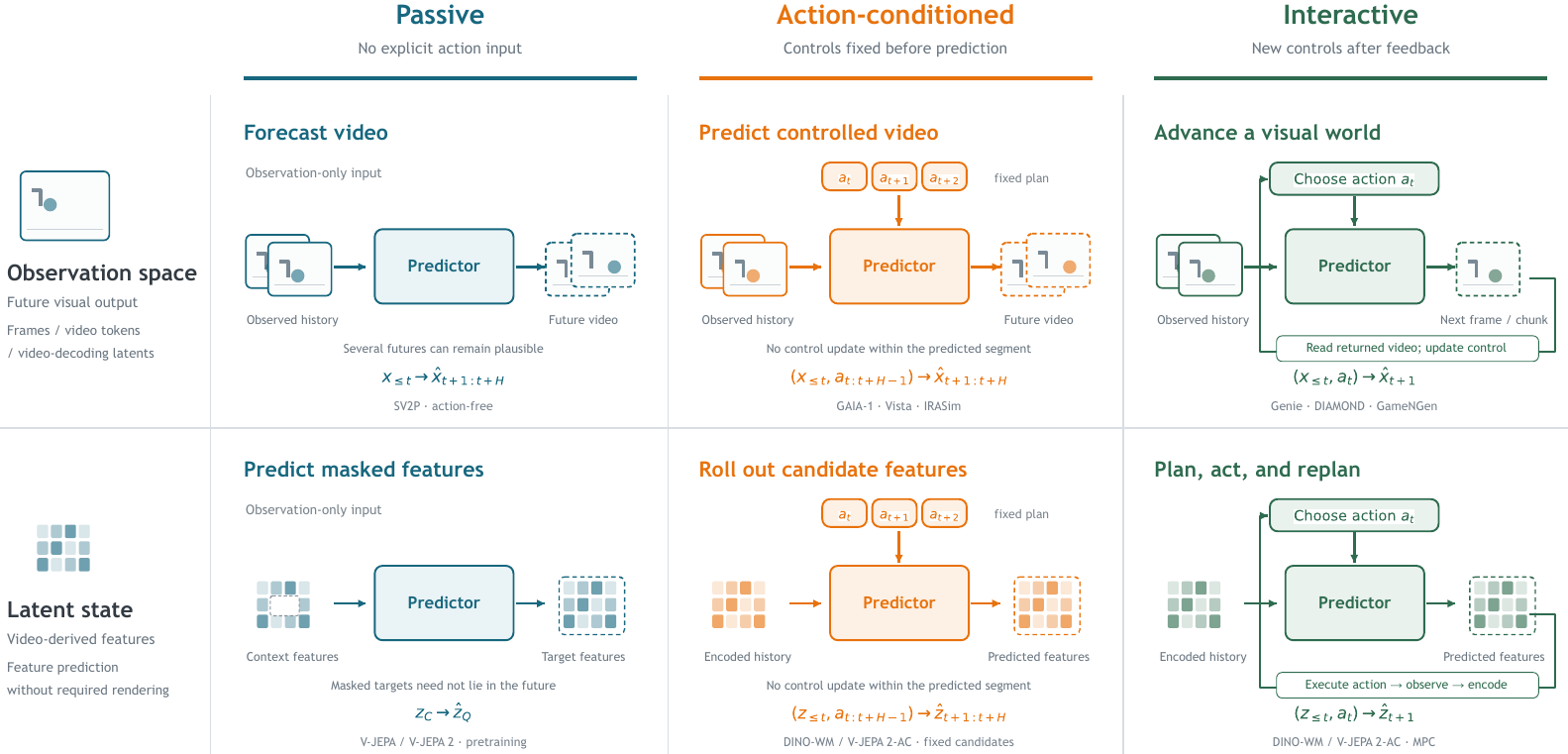}
  \caption{The $2\times3$ taxonomy of video world models.
  Rows give the prediction space: observations (frames, video tokens, or video-decoding latents) or latent states (video-derived features predicted without required rendering).
  Columns give the transition: passive prediction without action input, action-conditioned prediction with controls fixed before prediction, and interactive prediction that receives new controls after feedback.
  Each cell shows its prediction form and representative models; $x$ denotes observations, $z$ features, $a$ actions, $H$ the horizon, and $C$ and $Q$ the context and target positions of masked feature prediction.}
  \label{fig:worldmodel-taxonomy}
\end{figure}

\subsection{Architectures and Training Paradigms}

Crossing the two axes of Section~\ref{sec:wm-taxonomy} gives the $2\times3$ grid of Figure~\ref{fig:worldmodel-taxonomy}.
A cell describes how a model is trained and used, so one model can occupy several cells: an action-conditioned latent predictor can score fixed candidate actions or run inside a planner that acts, observes, and replans.
Likewise, the attention pattern does not decide the column; a model is interactive when it accepts new controls between emitted frames or chunks.
Table~\ref{tab:worldmodel-architectures} lists representative implementations, one row per evaluated protocol.

\leadlabel{Training data.}
Each cell needs data that expose the transition it learns.
Passive predictors such as V-JEPA 2 and V-JEPA 2.1 pretrain on large image--video mixtures~\cite{assran2025vjepa2,muckley2026vjepa21}, in which video supplies temporal change but no actions.
Action-conditioned learning needs synchronized observations and controls, such as the DROID robot demonstrations~\cite{khazatsky2024droid} that train the action-conditioned V-JEPA predictors.
Open X-Embodiment~\cite{openx2023embodiment} pools robot data across embodiments in one format, although action units and coordinate frames still need alignment.
Annotated video adds further supervision: camera geometry and depth (OmniWorld~\cite{omniworlddata2025}, SolarWM~\cite{huang2026solarwm}), multi-view tactile contact (Deform360~\cite{li2026deform360}), game actions with ground-truth world state (WildWorld~\cite{wildworld2026}), and recorded intervention branches (CG-World~\cite{cgworld2026}).

\begin{table}[!t]
\centering
\small
\caption{Video world-model implementations and evaluated protocols. \textbf{Observe} predicts video observations; \textbf{Latent} predicts feature states. \textbf{Passive} has no action input; \textbf{Action} uses a fixed control sequence; \textbf{Interactive} closes the loop. Marks refer to the listed protocol.}
\label{tab:worldmodel-architectures}
\setlength{\tabcolsep}{3.5pt}
\renewcommand{\arraystretch}{1.06}
\surveytablecolors
\begin{tabularx}{\textwidth}{@{}p{.41\textwidth} *{5}{>{\centering\arraybackslash}X}@{}}
\toprule
\surveytablehead
\tableheadcell{Model / protocol} & \multicolumn{2}{c}{\tableheadcell{Prediction space}} & \multicolumn{3}{c}{\tableheadcell{Transition protocol}} \\
\cmidrule(lr){2-3}\cmidrule(l){4-6}
\surveytablehead
 & \tableheadcell{Observe} &
\tableheadcell{Latent} & \tableheadcell{Passive} &
\tableheadcell{Action} & \tableheadcell{Interactive} \\
\midrule
SV2P, action-free~\cite{babaeizadeh2018sv2p} & \cmark & -- & \cmark & -- & -- \\
SV2P, fixed actions~\cite{babaeizadeh2018sv2p} & \cmark & -- & -- & \cmark & -- \\
GAIA-1, driving controls~\cite{hu2023gaia1} & \cmark & -- & -- & \cmark & -- \\
Vista~\cite{gao2024vista} & \cmark & -- & -- & \cmark & -- \\
IRASim~\cite{zhu2025irasim} & \cmark & -- & -- & \cmark & -- \\
WorldDirector, planned controls~\cite{wang2026worlddirector} & \cmark & -- & -- & \cmark & -- \\
\addlinespace
Genie, playable rollout~\cite{bruce2024genie} & \cmark & -- & -- & -- & \cmark \\
DIAMOND, policy loop~\cite{alonso2024diamond} & \cmark & -- & -- & -- & \cmark \\
GameNGen~\cite{valevski2025gamengen} & \cmark & -- & -- & -- & \cmark \\
Matrix-Game~\cite{zhang2025matrixgame} & \cmark & -- & -- & -- & \cmark \\
SolarWM, causal interaction~\cite{huang2026solarwm} & \cmark & -- & -- & -- & \cmark \\
Waypoint-1.5~\cite{rajpal2026waypoint15} & \cmark & -- & -- & -- & \cmark \\
\addlinespace
V-JEPA, pretraining~\cite{bardes2024vjepa} & -- & \cmark & \cmark & -- & -- \\
V-JEPA 2, pretraining~\cite{assran2025vjepa2} & -- & \cmark & \cmark & -- & -- \\
V-JEPA 2.1, pretraining~\cite{muckley2026vjepa21} & -- & \cmark & \cmark & -- & -- \\
DINO-WM, fixed candidates~\cite{zhou2024dinowm} & -- & \cmark & -- & \cmark & -- \\
V-JEPA 2-AC, fixed candidates~\cite{assran2025vjepa2} & -- & \cmark & -- & \cmark & -- \\
V-JEPA 2.1, navigation~\cite{muckley2026vjepa21} & -- & \cmark & -- & \cmark & -- \\
LeWorldModel, fixed candidates~\cite{maes2026leworldmodel} & -- & \cmark & -- & \cmark & -- \\
DINO-WM, MPC~\cite{zhou2024dinowm} & -- & \cmark & -- & -- & \cmark \\
V-JEPA 2-AC, robot MPC~\cite{assran2025vjepa2} & -- & \cmark & -- & -- & \cmark \\
V-JEPA 2.1, robot MPC~\cite{muckley2026vjepa21} & -- & \cmark & -- & -- & \cmark \\
H-JEPA, hierarchical MPC~\cite{zhang2026hjepa} & -- & \cmark & -- & -- & \cmark \\
LeWorldModel, MPC~\cite{maes2026leworldmodel} & -- & \cmark & -- & -- & \cmark \\
\bottomrule
\end{tabularx}
\end{table}

\subsubsection{Observation-Space Passive Prediction}

Passive visual models predict future observations from history alone.
They can represent uncertainty over several plausible futures, as the action-free setting of SV2P~\cite{babaeizadeh2018sv2p} does by sampling latent variables to predict multiple future frames.
Large video generators used as forecasters also belong here: Sora~\cite{brooks2024video} was presented as a step toward world simulation, Physics-IQ~\cite{motamed2025physicsiq} scores how such generators continue a real initial segment, and the base video-to-world models of Cosmos~\cite{nvidia2025cosmos} predict futures from a video prefix before any action-conditioned post-training.
Training targets the next frame or a future segment with reconstruction, likelihood, adversarial, diffusion, or flow losses.
Autoregressive models feed their predictions back as history, whereas block predictors generate a pending segment jointly.
Passive prediction supplies dynamics priors and uncertainty estimates for the controlled cells.

\subsubsection{Observation-Space Action-Conditioned Prediction}
\label{sec:wm-obs-action}

This cell receives the controls for a whole predicted segment before generation, such as ego motion, a robot trajectory, camera poses, road layout, or planned object states.
GAIA-1~\cite{hu2023gaia1} predicts discrete image tokens conditioned on text and action tokens before decoding video, and DrivingWorld~\cite{hu2024drivingworld} predicts driving observations together with ego state.
Conditions range from ego dynamics, agent configurations, and road semantics (GAIA-2~\cite{russell2025gaia2}) and commands, goal points, and trajectories (Vista~\cite{gao2024vista}) to frame-level robot actions (IRASim~\cite{zhu2025irasim}).
DriveDreamer~\cite{wang2024drivedreamer} and post-trained Cosmos variants~\cite{nvidia2025cosmos} condition on bird's-eye-view maps, 3D boxes, and camera configurations.

Most models in this cell adapt a pretrained video generator through condition encoders, cross-attention, or control branches, and add structure, multi-view, or action-fidelity losses to the visual objective.
Making them interactive requires causal generation that accepts new controls during rollout.
Vid2World~\cite{huang2025vid2world} makes a video diffusion model autoregressive and action-guided, minWM~\cite{zhao2026minwm} combines control tuning, autoregressive training, and few-step distillation, and BiWM~\cite{rui2026biwm} keeps bidirectional computation within causal generation.

\subsubsection{Observation-Space Interactive World Models}

Interactive visual models generate causally: emitted frames become history, and each new action changes the future that has not yet been generated.
Genie~\cite{bruce2024genie} predicts video tokens from learned latent actions, GameNGen~\cite{valevski2025gamengen} diffuses the next game frame from past frames and actions, and DIAMOND~\cite{alonso2024diamond} both trains policies inside its generated observations and runs as an interactive simulator.
GameFactory~\cite{gamefactory2025} extends playable generation to newly constructed scenes, and Matrix-Game~\cite{zhang2025matrixgame} couples visual context to keyboard and mouse input, with a later version~\cite{he2025matrixgame2} adding chunked, few-step streaming diffusion.
Oasis~\cite{oasis2024} renders a playable Minecraft-like world from keyboard and mouse input in real time, Hunyuan-GameCraft~\cite{gamecraft2025} maps keyboard and mouse actions into a shared camera representation and conditions on hybrid history to keep long play consistent, and Genie~3~\cite{genie3_2025} generates worlds from text prompts that users navigate in real time for several minutes.

Controls can also be speech, and several actors can share one world.
Vidu S1 and S2~\cite{vidus1_2026,vidus2_2026} take speech and updated references as controls on a streaming character (Section~\ref{sec:writing-apps}), and Wan-Streamer~\cite{wanstreamer2026} keeps audio-visual context to respond full-duplex to incoming events (Section~\ref{sec:wm-realtime}).
Solaris~\cite{solaris2026} pairs synchronized actions and observations from two Minecraft players, and MultiWorld~\cite{multiworld2026} binds actions to agent identities and updates a shared geometry-aware representation between chunks, so that every actor affects the same world.

\subsubsection{Latent-State Passive Prediction}

Latent passive models predict video-derived features without actions, often as pretraining for later transition models.
V-JEPA~\cite{bardes2024vjepa} learns by predicting the features of masked video regions, and the pretraining stage of V-JEPA 2~\cite{assran2025vjepa2} scales this objective over images and video.
Because the masked region can be predicted from frames on both sides, this objective is not causal forecasting; future prediction requires a separate, causal protocol.
We keep these models in the taxonomy because their predictors supply the feature space and initialization of the action-conditioned latent cells; on their own, they are pretraining for transition models rather than complete world models.
Later variants change the prediction unit or the loss.
Causal-JEPA~\cite{causaljepa2026} infers masked object-level latents from surrounding objects to encourage relational prediction.
V-JEPA 2.1~\cite{muckley2026vjepa21} adds a dense context-token loss and multi-level supervision for stronger local structure.

\subsubsection{Latent-State Action-Conditioned Prediction}

Adding actions keeps a feature-space target but conditions its evolution on controls.
V-JEPA 2-AC~\cite{assran2025vjepa2} post-trains a predictor on robot trajectories, DINO-WM~\cite{zhou2024dinowm} learns how frozen DINOv2 features evolve under actions, and LeWorldModel~\cite{maes2026leworldmodel} instead learns the encoder and predictor jointly from pixels with latent regularization.
V-JEPA 2.1 also trains a diffusion predictor over its features for navigation~\cite{muckley2026vjepa21}.

Such models can roll a candidate action sequence forward and compare the result with a goal representation without rendering a single frame.
LaDi-WM~\cite{huang2025ladiwm} predicts geometric and semantic features for a manipulation policy, and MoWM~\cite{shang2025mowm} combines motion-aware latent prediction with pixel-space signals.
The JEPA-WM design study~\cite{terver2025jepawmstudy} compares encoders, proprioception, context, multi-step losses, and planning optimizers in one recipe and finds that the best combination depends on the task.

\subsubsection{Latent-State Interactive World Models}

An interactive latent system closes the loop with model predictive control (MPC): the planner chooses an action prefix, executes it, encodes the new observation, and plans again.
V-JEPA 2-AC~\cite{assran2025vjepa2} and DINO-WM~\cite{zhou2024dinowm} use this loop, and V-JEPA 2.1~\cite{muckley2026vjepa21} applies it to reaching, grasping, and pick-and-place on a real robot arm.
H-JEPA~\cite{zhang2026hjepa} stacks latent spaces at coarser timescales so that upper levels supply subgoals to lower-level planning.
Latent rollout spends no capacity on texture and scores candidates without a decoder, but latent distance can omit task-relevant details.
Because corrective replanning can hide poor scores, scorer-level audits complement closed-loop success (Section~\ref{sec:wm-benchmarks}).

\begin{insightbox}{Rendering, causal rollout, and decision state can be specialized}
A visual generator can supply appearance and motion priors, an online generator can provide action-conditioned rollout, and a latent predictor can supply compact state for planning.
These components need not share a representation or training stage, but for the same history and action their rendered consequences, retained state, and decision scores must agree.
\end{insightbox}

\subsection{Core Technical Challenges}

Four challenges recur across the grid: control, memory, latency, and physical grounding.
Memory and physical consistency matter in every cell; controlled prediction must also bind each control to its consequence, and interaction must do so within a response budget.
The challenges interact: longer memory raises inference cost, fewer denoising steps change rollout stability, and stronger action constraints can disturb the visual prior inherited from pretraining.

\subsubsection{Action-Conditioned Controllability}
\label{sec:wm-control}

Action control poses two problems.
Without action records, a latent-action model must learn a control variable from observed transitions; with supplied controls, conditioning must bind each signal to the right time, view, object, and consequence.

\leadlabel{Latent action learning.}
Latent-action models couple inverse and forward dynamics.
An inverse model infers a variable that explains the change between two nearby frames, and a forward model predicts the later frame from the earlier one and that variable.
A discrete codebook or continuous bottleneck limits how much information the variable carries, which favors a compact control code.
LAPO~\cite{schmidt2024lapo}, LAPA~\cite{ye2025lapa}, and Moto~\cite{chen2024moto} use discrete latent actions to pretrain policies from unlabeled video, and Genie~\cite{bruce2024genie} uses them to make platformer videos playable.
LAPA then maps its codes to robot commands with a small labeled set, while Moto treats them as a language of motion tokens.
Continuous latent actions offer more capacity.
AdaWorld~\cite{gao2025adaworld} conditions an autoregressive world model on them and adapts to new action spaces with little interaction, while IGOR~\cite{chen2024igor} encodes initial-to-goal image changes in one space for humans and robots.

However, a latent action can lock onto the wrong source of change.
LAOM~\cite{nikulin2025laom} shows that visual distractors lead latent actions to encode background or camera motion, whereas a little action supervision anchors them to the agent.
Latent codes should therefore be tested for decoding into executable actions and for transfer under distractors, not only for reconstruction.
Action Forcing~\cite{sundar2026actionforcing} avoids an unconstrained code: it recovers signed egomotion axes from principal components of tracked pixel motion and uses them as controls that can be tested for reversal, scaling, and composition.

\leadlabel{Action conditioning mechanisms.}
Known actions remove the discovery problem but must still be aligned with the frames they affect.
A single sequence embedding is too coarse when each frame must respond to a different command, so IRASim~\cite{zhu2025irasim} injects actions frame by frame throughout its diffusion Transformer.
Other systems express controls where their effects appear.
EA-WM~\cite{yang2026eawm} projects joint or end-effector kinematics into the camera view as visual action fields, and Aero-World~\cite{alradi2026aeroworld} feeds acceleration and angular velocity as tokens and checks inertial consistency with a frozen latent physics probe.
X-World~\cite{zheng2026xworld} conditions multi-camera driving on ego actions plus optional controls over traffic agents, road elements, and appearance.
Models that predict both states and actions must also stay responsive to actions; FLEX-WAM~\cite{khorrambakht2026flexwam} rebalances the two losses and tracks an action-sensitivity proxy for this purpose.

\subsubsection{Long-Horizon Consistency and Memory}
\label{sec:wm-memory}

Long-horizon failure occurs at two scales.
Locally, small prediction errors return as context and accumulate across frames or chunks; globally, a finite context drops entities, layouts, and events that are no longer visible.
The first calls for aligning training histories with deployment, the second for storing, compressing, or recurrently updating information beyond the active window.
TECO~\cite{yan2023teco} is an early example, compressing past frames into temporally consistent latent codes to keep long rollouts coherent.

\leadlabel{Training-inference alignment: the forcing family.}
Under teacher forcing, the model predicts from clean ground-truth history, whereas at deployment it conditions on its own outputs, so a small error in geometry or identity becomes the next input.
Diffusion Forcing~\cite{chen2024diffusionforcing} exposes the model to uncertain histories by giving sequence elements independent noise levels; GameNGen similarly perturbs its conditioning frames with noise, and LIVE~\cite{liveworldmodel2026} adds forward--reverse cycle consistency.
MAGI-1~\cite{teng2025magi} denoises chunks block-causally.
Autoregressive diffusion adds a second mismatch, between a many-step bidirectional teacher and a few-step causal student.
CausVid~\cite{yin2025causvid} makes a pretrained video diffusion model causal and distills its sampler through distribution matching, Self-Forcing~\cite{selfforcing2025} trains the student on its own rollouts, and Causal Forcing~\cite{zhu2026causalforcing} distills from an autoregressive teacher whose sampling path already follows causal inference.
Later variants change the denoising window (Rolling Forcing~\cite{liu2025rollingforcing}), the prediction horizon (Next Forcing~\cite{xu2026nextforcing}), or the distillation objective (One-Forcing~\cite{feng2026oneforcing}, Delta Forcing~\cite{wu2026deltaforcing}); Table~\ref{tab:forcing-family} lists their mechanisms and intended effects.

\leadlabel{Retrieval-augmented and structured memory.}
Retrieval extends the context with stored observations that are recalled when the camera or agent returns to a relevant region.
WorldMem~\cite{xiao2025worldmem} stores frames with pose and time metadata and attends to them to reconstruct earlier scenes; Context-as-Memory~\cite{yu2025contextmemory} retrieves only scene-relevant past views, and WorldWeaver~\cite{liuworldweaver} builds its memory on predicted depth, which drifts less than RGB.
Spatial Memory Intelligence~\cite{yang2026smi} uses a multimodal LLM to cluster, sparsify, filter, and retrieve stored views by action, while Keepsake~\cite{alradi2026keepsake} caps the archive by evicting views that others can replace.
Such archives trade growing storage and retrieval cost for detailed evidence that can be revisited.

Geometry makes recall more precise and reduces false matches between similar-looking places.
VMem~\cite{li2025vmem} indexes views by 3D surfels, RELIC~\cite{hong2025relic} stores compressed history with relative actions and absolute camera poses in its key--value (KV) cache, and AnchorWeave~\cite{anchorweave2026} retrieves local memories around persistent anchors.
Long-term spatial memory~\cite{wu2025longtermspatialmemory} keeps a reconstructed static point map for layout, recent frames for motion, and retrieved past frames for appearance.
Matrix-Game 3.5~\cite{qian2026matrixgame35} keeps geometrically projected latent patches for the static scene and reference tokens for moving subjects, so that past positions of an object are not mistaken for additional objects.

However, remembering what was seen differs from tracking what changes off-screen.
WorldDirector~\cite{wang2026worlddirector} conditions causal chunks on externally planned object and camera trajectories plus retrieved object crops, so planned objects keep moving while unseen.
LiveWorld~\cite{duan2026liveworld} keeps a static background and localized dynamic entities and uses virtual monitors to generate their progress outside the view before projecting it back into the scene; its LiveBench checks this off-screen progression.

\leadlabel{Context compression.}
Compression keeps a bounded working context.
FramePack~\cite{zhang2025framepack} gives older frames smaller representation budgets than recent ones and adds drift prevention to sampling.

\leadlabel{Fixed-size state representations.}
Fixed-size memories keep a constant-size state and pay a compression cost at every update.
Long-context state-space world models~\cite{po2025long} pair a block-wise state-space scan with dense local attention, and Echo-Infinity~\cite{bian2026echoinfinity} recurrently updates memory queries as frames leave its local window.
GIM-World~\cite{wei2026gimworld} encodes selected past latents into geometry-supervised tokens, and Owl-1~\cite{huang2024owl1} keeps learned state queries inside a multimodal LLM alongside predicted textual dynamics.
Honeycomb~\cite{shi2026honeycomb} recurrently writes six fixed-size spatial and spatiotemporal feature planes that cover larger ranges at coarser resolution.
WorldTrace~\cite{wu2026worldtrace} keeps a fixed-size KV memory of averaged scene summaries and selected landmark keys that later queries can reach beyond the recent window, and evaluates it with its LoopBench revisit test.
Recurrent states can drift, test-time weight updates~\cite{dalal2025tttvideo} add optimization latency, and compact memories may drop details needed later for rendering.
Retrieval, compression, and fixed state thus trade capacity against latency differently and can be combined (Table~\ref{tab:memory-mechanisms}).
\begin{table}[!t]
\centering
\small
\caption{Training mechanisms for causal and streaming generation. DMD: distribution-matching distillation.}
\label{tab:forcing-family}
\surveytablecolors
\setlength{\tabcolsep}{4pt}
\begin{tabularx}{\textwidth}{@{}>{\hsize=.72\hsize}Y >{\hsize=1.28\hsize}Y Y@{}}
\toprule
\surveytablehead
\tableheadcell{Method} & \tableheadcell{Training mechanism} & \tableheadcell{Intended effect} \\
\midrule
Teacher forcing & Clean ground-truth history; next-step prediction & Fit observed transitions; exposure mismatch remains \\
\addlinespace
Diffusion Forcing~\cite{chen2024diffusionforcing} & Independent noise levels across sequence elements & Denoise from histories with varying uncertainty \\
\addlinespace
CausVid~\cite{yin2025causvid} & DMD from a bidirectional teacher to a causal student & Few-step causal generation \\
\addlinespace
Self-Forcing~\cite{selfforcing2025} & Self-generated histories; video-level distribution matching & Align training context with autoregressive inference \\
\addlinespace
Causal Forcing~\cite{zhu2026causalforcing} & Causal-teacher ODE initialization followed by DMD on rollouts & Align initialization with causal prediction \\
\addlinespace
Causal Forcing++~\cite{zhao2026causalforcingpp} & Causal consistency initialization followed by DMD & Reduce initialization cost for frame-wise few-step generation \\
\addlinespace
Rolling Forcing~\cite{liu2025rollingforcing} & Self-generated history; joint window denoising with progressive noise & Reduce drift in streaming generation \\
\addlinespace
Next Forcing~\cite{xu2026nextforcing} & Ground-truth history; auxiliary future-chunk denoising targets & Extend predictive supervision; parallelize next-chunk inference \\
\addlinespace
One-Forcing~\cite{feng2026oneforcing} & Frame-wise rollouts; DMD plus adversarial learning & Stabilize one-step autoregressive generation \\
\addlinespace
Delta Forcing~\cite{wu2026deltaforcing} & Transition-calibrated weighting of DMD and continuity regularization & Balance condition changes with trajectory continuity \\
\bottomrule
\end{tabularx}
\end{table}

\begin{table}[!t]
\centering
\small
\caption{Memory mechanisms for long-horizon video generation and world modeling. Horizon scaling distinguishes active context from the full archive.}
\label{tab:memory-mechanisms}
\setlength{\tabcolsep}{3pt}
\renewcommand{\arraystretch}{1.05}
\surveytablecolors
\begin{tabularx}{\textwidth}{@{}>{\hsize=.72\hsize}Y >{\hsize=1.00\hsize}Y >{\hsize=.73\hsize}Y >{\hsize=1.35\hsize}Y >{\hsize=1.20\hsize}Y@{}}
\toprule
\surveytablehead
\tableheadcell{Memory type} & \tableheadcell{Stored state} &
\tableheadcell{Horizon scaling} & \tableheadcell{Examples} &
\tableheadcell{Main limitation} \\
\midrule
Visual retrieval &
frames, tokens, or anchors selected by similarity or pose &
growing or capped archive &
WorldMem~\cite{xiao2025worldmem}, RELIC~\cite{hong2025relic},
Context-as-Memory~\cite{yu2025contextmemory},
AnchorWeave~\cite{anchorweave2026},
SMI~\cite{yang2026smi}, Keepsake~\cite{alradi2026keepsake} (capped) &
archives grow; eviction loses detail; false matches corrupt recall \\
\addlinespace
Spatial caches &
static maps, surfels, projected features, and reference views &
grows with retained scene coverage &
VMem~\cite{li2025vmem}, Long-term Spatial Memory~\cite{wu2025longtermspatialmemory},
Matrix-Game 3.5~\cite{qian2026matrixgame35} &
geometry errors persist; static caches alone do not model off-screen change \\
\addlinespace
Planned dynamics with retrieval &
object/camera trajectories, crops, and selected history &
bounded selected context; growing video archive &
WorldDirector~\cite{wang2026worlddirector} &
off-screen evolution is externally planned; chunks do not establish online interaction \\
\addlinespace
Addressable KV summaries &
recent window plus Field summaries or Landmark slots &
fixed GPU cache; host scaling depends on writer &
WorldTrace~\cite{wu2026worldtrace}; evaluated with LoopBench &
host history grows by default; bounded aggregates lose information \\
\addlinespace
Compressed context &
selected chunks or summaries with unequal temporal budgets &
bounded working context &
StreamingT2V~\cite{henschel2025streamingt2v}, FramePack~\cite{zhang2025framepack} &
discarded detail requires another retained source or a new observation \\
\addlinespace
Fixed-size memory &
memory queries; spatial and spatiotemporal feature planes &
fixed carrier; history budget accounted separately &
Echo-Infinity~\cite{bian2026echoinfinity}, Owl-1~\cite{huang2024owl1},
GIM-World~\cite{wei2026gimworld}, Honeycomb~\cite{shi2026honeycomb} &
compression loses detail; state updates or history re-encoding add cost \\
\bottomrule
\end{tabularx}
\end{table}

\subsubsection{Real-Time Responsiveness}
\label{sec:wm-realtime}

Responsiveness is the delay between a new action and the first output that reflects it; a slow transition breaks the control loop even when offline throughput is high.
End-to-end cost combines sampling depth, work per step, active history, cache construction, latent decoding, and the rate at which controls enter the rollout.
Recent systems report real-time rates: Vidu S1~\cite{vidus1_2026} reaches up to 42 frames per second (FPS) at 540p on consumer GPUs, and Wan-Streamer~\cite{wanstreamer2026} reaches 25 FPS.
Display rate can still overstate responsiveness: with temporal compression, several frames are decoded per predicted latent, so actions may enter less often than frames appear.
Waypoint-1.5~\cite{rajpal2026waypoint15}, a causal diffusion Transformer with bounded KV caches for keyboard-mouse play on consumer hardware, accordingly reports rendered and latent FPS and control rate separately.

\leadlabel{Few-step distillation and adversarial post-training.}
Training-based acceleration turns a many-step generator into a causal student with fewer evaluations per frame or chunk, as CausVid, Self-Forcing, and Causal Forcing do while aligning rollouts.
Adversarial post-training goes further: Seaweed-APT~\cite{lin2025seaweedapt} distills a video diffusion model into a one-step generator, AAPT~\cite{lin2025aapt} applies autoregressive adversarial post-training to real-time interactive video, and Causal Forcing++~\cite{zhao2026causalforcingpp} scales few-step autoregressive distillation.
Once sampling depth is reduced, the backbone, active context, and decoder dominate the cost, and the student must also stay stable over long rollouts.

\leadlabel{Training-free acceleration.}
Training-free methods exploit redundancy at inference.
Sparse attention suits streaming because the current query is short and the history long: Light Forcing~\cite{lv2026lightforcing} keeps the anchors and recent frames that control the next output.
Caching reuses features that change little across denoising steps, as in TeaCache~\cite{liu2024teacache} and FasterCache~\cite{lv2024fastercache}.
WorldCache~\cite{feng2026worldcache} adapts caching to world models, where an error in a few control-sensitive action, image, or memory tokens can propagate through the rollout.
TurboDiffusion~\cite{turbodiffusion2025} combines both routes, stacking step distillation with low-bit and sparse attention and weight--activation quantization, reporting two orders of magnitude of end-to-end acceleration on a single consumer GPU.

\subsubsection{Physical and Causal Grounding}
\label{sec:wm-physics}

Physical grounding asks whether motion and interaction respect contact, rigidity, gravity, and persistence.
Causal grounding asks how the predicted future changes when an action or initial condition is intervened on, and whether unaffected factors stay fixed.
Current methods offer much machinery for plausible dynamics, whereas intervention-level supervision and evaluation remain less developed.

\leadlabel{Explicit physics simulation integration.}
Explicit methods let a simulator produce the dynamics and a generator render them.
PhysGen~\cite{liu2024physgen} and PhysMotion~\cite{tan2024physmotion} estimate object shape and material from a single image, simulate the motion, and use it to guide video generation.
PhysGen3D~\cite{chen2025physgen3d} extends this to interactive 3D scenes with user-set initial conditions, and PhysCtrl~\cite{wang2025physctrl} instead replaces the simulator with a generative physics network trained on simulated animations, conditioned on material parameters and forces.
Mass, material, force, or initial velocity thus become editable, but deformable bodies, fluids, fracture, articulated agents, and partly observed contact remain hard to recover from pixels and to simulate.

\leadlabel{Implicit physics alignment.}
Implicit approaches keep a learned generator and change what it learns from.
VideoREPA~\cite{zhang2025videorepa} distills token-level relations from a self-supervised video encoder, which carries stronger physical structure, into a text-to-video generator.
Phantom~\cite{shen2026phantom} couples visual trajectories with auxiliary physical latents, and RoboScape~\cite{shang2025roboscape} trains with depth and temporal structure as additional targets.
PhyGDPO~\cite{cai2025phygdpo} post-trains on preferences between candidate videos ranked by physical plausibility, and WROP~\cite{zhang2026wrop} supplies procedurally generated Blender data for object permanence and solidity.
Inference-time planning instead adds no physics training: VLIPP~\cite{yang2025vlipp} uses a vision-language model to plan coarse physical motion and converts the plan into structured noise for a motion-controllable image-to-video diffusion model.

\leadlabel{Intervention-level causal evidence.}
Plausibility objectives do not record which force, action, or initial condition produced a motion.
Causal evidence instead holds the initial state fixed, varies one action or physical parameter, and compares both the affected future and the unchanged state.
PhysEditWorld~\cite{physeditworld2026} supplies such data by replaying identical game episodes under different gravity settings, with synchronized video, action, engine-state, and parameter signals (see also Section~\ref{sec:challenge-physics}).
OneWorld~\cite{he2026oneworld} requires several action-conditioned futures of one scene to share a latent physical explanation, scored by a mechanism interpreter during flow training and sampling.
Different actions should then change outcomes, not the world's mechanism.

\subsection{Applications}

Applications differ in who consumes the prediction.
A world model may produce a dataset, train a policy, evaluate a policy, supply forecasts to a planner, or respond directly to a person.
The first three support decisions before deployment, whereas planning and human simulation consume predictions online.
The same action-conditioned generator can move between these roles as its control fidelity and temporal stability improve.

\subsubsection{Data Engines}

As a data engine, a world model turns controllable variation into training examples where real observations are expensive, dangerous, or unevenly distributed.
Each example must include the labels the learner needs, often actions.
In robotics, DreamGen~\cite{jang2025dreamgen} and GigaWorld-0~\cite{gigaworld2025} synthesize action-labeled trajectories for policy training (Section~\ref{sec:writing-apps}).
In driving, the action-conditioned generators of Section~\ref{sec:wm-obs-action}, such as GAIA-2~\cite{russell2025gaia2}, Vista~\cite{gao2024vista}, and Cosmos~\cite{nvidia2025cosmos}, supply unusual layouts and safety-critical cases that fleet logs rarely contain.

\subsubsection{Policy Training in Imagination}
\label{sec:wm-imagination}

Imagination training keeps the world model inside the optimization loop: a policy acts, the model predicts the next observation or state, and a reward updates the policy.
Unlike a data engine, the generated distribution then follows the current policy, and errors compound along the trajectories it chooses.
Action-conditioned video models such as UniSim~\cite{yang2024unisim}, iVideoGPT~\cite{ivideogpt2024}, and Pandora~\cite{pandora2024} can serve as this simulator if their action effects, rewards, and resets are accurate enough.

World-VLA-Loop~\cite{liu2026worldvlaloop} refines vision-language-action policies by reinforcement learning in a world model that predicts future frames and binary rewards.
It feeds the updated policy's rollouts back into the model, since a fixed simulator fails most often on states that a changed policy visits.
Dreamer 4~\cite{hafner2025dreamer4} trains a Minecraft agent with reinforcement learning inside a video world model learned largely from offline video and action data.
Because each step of a diffusion world model may need many denoising evaluations, Horizon Imagination~\cite{cohen2026horizon} denoises several future observations in parallel and decouples the denoising budget from the rollout horizon, making longer on-policy rollouts affordable.

\subsubsection{Policy Evaluation Environments}
\label{sec:wm-policy-eval}

For policy evaluation, a world model replaces repeated deployment with reproducible generated trials.
WorldGym~\cite{quevedo2025worldgym} rolls out robot policies in an action-conditioned video model and scores them with a vision-language model, while WorldEval~\cite{li2025worldeval} encodes policy behavior with Policy2Vec to drive its simulation.
Both are validated by whether generated success rates preserve real-world policy rankings.
Such agreement depends on the policies tested, since a simulator tuned on familiar checkpoints may misrank a controller that visits new states or exploits a modeling artifact.

Later systems add multi-view, memory-supported rollouts (Ctrl-World~\cite{guo2025ctrlworld}) or task-progress and success judges (GE-Sim 2.0~\cite{qiu2026gesim2}), and generalist robot (DreamDojo~\cite{gao2026dreamdojo}) and driving (GAIA-3~\cite{wayve2025gaia3}) world models list policy evaluation among their uses.
OmniDreams~\cite{basant2026omnidreams} takes a hybrid route in driving, generating sensor observations online inside the AlpaSim simulator, whose own services handle traffic and physics.

\subsubsection{Predict-then-Act Planning}

Predict-then-act methods use a forecast as evidence for the next decision and differ in how much of the future they decode.
At the pixel end, UniPi~\cite{du2023unipi} and AVDC~\cite{ko2024avdc} generate a video plan and recover actions with an inverse-dynamics model or dense correspondences (Section~\ref{sec:writing-apps}).
Such plans are easy to inspect but add generation cost and another source of error.
Navigation World Models~\cite{nwm2024} instead predict egocentric observations under candidate navigation actions with a conditional diffusion Transformer and plan by checking which simulated trajectory reaches the goal.
VPP~\cite{hu2025vpp} instead conditions an inverse-dynamics policy on future-aware features from a fine-tuned video diffusion model.
LaDi-WM~\cite{huang2025ladiwm} and MoWM~\cite{shang2025mowm} combine latent prediction with feature or pixel signals, and LaWAM~\cite{chen2026lawam} predicts latent visual subgoals without reconstructing video.
At the latent end, V-JEPA 2-AC and DINO-WM score candidate actions without decoding pixels, which cuts prediction cost but makes their forecasts harder to inspect.

Prediction can also serve only as training supervision.
WLA-0~\cite{yang2026wla0} trains shared representations with a future-frame World Expert and an Action Expert, so the World Expert can be dropped at test time or re-enabled to select among sampled action chunks.

\subsubsection{Interactive Simulation for Humans}

In human-interactive simulation, world prediction is the application: a person supplies actions, and the model produces the evolving observation stream.
Among playable systems, Genie made unlabeled video playable and Genie~3~\cite{genie3_2025} opens text-prompted worlds to real-time exploration, GameNGen simulates an existing game in real time, and Oasis~\cite{oasis2024} and Matrix-Game 2.0~\cite{he2025matrixgame2} target real-time open deployment.
Yume~\cite{yume2025} and RELIC~\cite{hong2025relic} extend exploration to real-world scenes, with RELIC adding long-horizon memory for revisits.
Voice-controlled streaming characters such as Vidu S1 and Wan-Streamer (Section~\ref{sec:wm-realtime}) extend the input from game actions to speech, so response timing and turn-level coherence matter alongside visual continuity.
In driving, GAIA-2, Vista, and Cosmos-Drive-Dreams~\cite{cosmosdrivedreams2025} let users set trajectories, actions, or scene conditions and inspect the resulting futures.
With a fixed-control model, the user edits conditions between rollouts, whereas an interactive system accepts input within the stream.

\subsection{Benchmarks and Evaluation}
\label{sec:wm-benchmarks}

World-model evaluation follows the role of the prediction.
A rendered future needs visual and motion quality, and an interactive simulator adds action fidelity, persistent state, and bounded response time.
Planning, policy training, and policy evaluation further require the prediction to improve or faithfully estimate a downstream decision.
These requirements form the Interacting row of the evidence ladder in Table~\ref{tab:evidence-ladder}, from visual plausibility through controllable rollout to decision utility~\cite{yu2026worldmodeleval}.
Table~\ref{tab:worldmodel-benchmarks} indexes current benchmarks by the claim they test and the main limits on comparing their scores.
\begin{table}[!t]
\centering
\small
\caption{World-model benchmarks and protocols by target, with limits on interpreting and comparing scores.}
\label{tab:worldmodel-benchmarks}
\setlength{\tabcolsep}{3.5pt}
\renewcommand{\arraystretch}{1.10}
\surveytablecolors
\begin{tabularx}{\textwidth}{@{}>{\hsize=.70\hsize}Y >{\hsize=1.40\hsize}Y >{\hsize=.90\hsize}Y@{}}
\toprule
\surveytablehead
\tableheadcell{Evaluation target} & \tableheadcell{Benchmarks / protocols} & \tableheadcell{Interpretation limit} \\
\midrule
\textbf{Video plausibility} &
VBench~\cite{huang2023vbenchcomprehensivebenchmarksuite}, VBench++~\cite{huang2024vbenchpp}, VBench-2.0~\cite{zheng2025vbench20advancingvideogeneration} & No closed-loop action or hidden-state test \\
\addlinespace
\textbf{Action response} &
ACT-Bench~\cite{arai2024actbench}, iWorld-Bench~\cite{fang2026iworldbench}, MiraBench~\cite{yang2026mirabench} & Action spaces and control interfaces differ \\
\addlinespace
\textbf{State persistence} &
MIND~\cite{ye2026mind}, MemoBench~\cite{memobench2026}, LoopBench~\cite{wu2026worldtrace} & Static revisits alone do not test off-screen dynamics \\
\addlinespace
\textbf{Interactive latency} &
Waypoint-1.5~\cite{rajpal2026waypoint15} and system protocols & Display, latent, and control rates differ \\
\addlinespace
\textbf{Physical validity} &
WorldModelBench~\cite{li2025worldmodelbench}, Physics-IQ~\cite{motamed2025physicsiq}, WorldBench~\cite{worldbench2026}, CRONOS~\cite{cronos2026}, What-If World~\cite{whatifworld2026} & Intervention modalities differ \\
\addlinespace
\textbf{Scene / semantic consistency} &
EWMBench~\cite{ewmbench2025} & Visual similarity alone does not establish state correctness \\
\addlinespace
\textbf{Decision utility} &
WorldGym~\cite{quevedo2025worldgym}, WorldEval~\cite{li2025worldeval}, WMBench~\cite{gigaworldteam2026gigaworld1}, RoboWM-Bench~\cite{jiang2026robowmbench}, WorldArena~\cite{shang2026worldarena}, ARC-Bench~\cite{zhang2026arcbench} & Action/policy ranking and executed task success are distinct outcomes \\
\addlinespace
\textbf{Multiple dimensions} &
WorldScore~\cite{duan2025worldscore}, WorldSimBench~\cite{qin2024worldsimbench}, WorldMark~\cite{xu2026worldmark}, WBench~\cite{ying2026wbench}, WorldOlympiad~\cite{worldolympiad2026}, WorldRoamBench~\cite{worldroambench2026} & Aggregate scores can hide dimension-specific failures \\
\addlinespace
\textbf{Calibration} &
C$^3$ evaluation~\cite{mei2025c3}, PAWBench~\cite{pu2026pawbench}, CaliBench~\cite{sadeghi2026calibench} & Calibration targets and statistical tests differ \\
\bottomrule
\end{tabularx}
\end{table}

\subsubsection{Challenge-Aligned Benchmarks}

\leadlabel{Action following and controllability.}
Action benchmarks test whether changing a control changes the predicted future as intended.
ACT-Bench~\cite{arai2024actbench} measures instruction--execution consistency and trajectory error for driving maneuvers, and iWorld-Bench~\cite{fang2026iworldbench} separates trajectory following from visual quality and memory through a unified action layer.
In robotics, MiraBench~\cite{yang2026mirabench} tests action following, physical adherence, and optimism under actions that should fail, which matters because a simulator that turns failed actions into plausible successes misleads both policy training and evaluation.
Since a game key, a camera pose, a robot action chunk, and an ego trajectory need different error measures, reports should state the native action space, horizon, intervention frequency, and source of trajectory estimates.

\leadlabel{Long-horizon consistency and memory.}
Consistency asks how identity, geometry, and appearance degrade with duration even when content stays visible.
Measuring this degradation requires scores stratified by video length.
Memory tests instead remove content from view and check it later.
Revisit benchmarks such as LoopBench~\cite{wu2026worldtrace} test static recall, comparing returns with the model's own first visit.

However, a static cache passes revisit tests even when the world should have changed.
MBench~\cite{zhang2026mbench} separates entity, environment, and causal memory under occlusion and off-screen evolution, STEVO-Bench~\cite{ma2026stevobench} hides an ongoing process and checks whether it advanced, and MemoBench~\cite{memobench2026} requires an object that changed out of view to return in its updated state.
MIND~\cite{ye2026mind} and WorldRoamBench~\cite{worldroambench2026} test memory during closed-loop or open-world interaction, the latter scoring memory separately from action following, visual drift, and controllability-gated physics.
Memory Maze~\cite{pasukonis2022memorymaze} poses the same retention question for navigation agents.
These protocols score decoded video; latent-state models lack matching probes of drift and goal reachability.

\leadlabel{Real-time response.}
Latency is reported per system, usually as FPS or time to first response, and mixes model forward time with denoising steps, decoding, cache construction, input rate, and hardware.
A matched protocol would measure the delay from an action's arrival to the first affected frame, sustained throughput, and latency variance over long sessions, at stated resolution, precision, hardware, and action frequency.

\leadlabel{Physical, temporal, and causal validity.}
Physical evaluation ranges from prompt-based plausibility to comparison with recorded outcomes.
Prompt-based suites such as VideoPhy and VideoPhy-2~\cite{bansal2024videophy,bansal2025videophy2} are described in Section~\ref{sec:writing-eval}.
PhyGenBench~\cite{meng2024phygenbench} organizes prompts by physical principle, and WorldModelBench~\cite{li2025worldmodelbench} combines instruction following with checks for violations such as broken conservation.
Physics-IQ~\cite{motamed2025physicsiq} instead compares continuations of a real initial segment with recorded outcomes, and PhyWorldBench~\cite{gu2025phyworldbench} adds anti-physics prompts that test whether a model can follow instructions that deliberately violate physics while staying logically consistent.
Other suites probe specific factors.
WorldBench~\cite{worldbench2026} separates intuitive principles from quantities such as friction and viscosity, and CRONOS~\cite{cronos2026} varies viewpoint, scene, appearance, and category within controlled events.
What-If World~\cite{whatifworld2026} pairs prompts that change one physical detail and scores both the outcome difference and the preservation of the environment.
ChronoMagic-Bench~\cite{yuan2024chronomagicbenchbenchmarkmetamorphicevaluation} tests temporal direction through metamorphic and time-lapse processes such as growth or construction.
Prompt-level and rendered contrasts cover only part of the intervention protocol of Section~\ref{sec:wm-physics}, which ultimately needs logged actions and real outcomes.

\leadlabel{Calibration of predicted futures.}
Stochastic world models should also produce the right spread of futures.
PAWBench~\cite{pu2026pawbench} and CaliBench~\cite{sadeghi2026calibench} sample repeated futures from a fixed input and compare the outcome distribution with reference probabilities or sets of valid outcomes, reporting separately how often outcomes can be scored.
C$^3$~\cite{mei2025c3} calibrates confidence rather than outcome frequencies: it trains a probe that predicts whether a latent prediction is accurate and evaluates it against accuracy labels.

\subsubsection{Application-Aligned Benchmarks}

Application-aligned benchmarks measure a prediction's effect after another component consumes it, through policy-ranking agreement, optimization gains, planning success, executable action success, or data efficiency.
For policy evaluation, WorldGym~\cite{quevedo2025worldgym} and WorldEval~\cite{li2025worldeval} compare model-based success rates or rankings with real robot trials.
WMBench, introduced with GigaWorld-1~\cite{gigaworldteam2026gigaworld1}, studies evaluator design with an ordinal score that separates correct from incorrect task outcomes and high from low visual fidelity.
For planning, WorldPrediction~\cite{chen2025worldprediction} tests high-level state transitions and procedural ordering, using counterfactual distractors that remove low-level continuity shortcuts.
World-in-World~\cite{zhang2025worldinworld} places heterogeneous models behind a shared action interface and measures closed-loop task success.

Embodied benchmarks make predictions executable or check them against task requirements.
RoboWM-Bench~\cite{jiang2026robowmbench} converts generated manipulation videos into actions and executes them in a physics simulator, so visually plausible but unexecutable motion earns little credit.
EWMBench~\cite{ewmbench2025} separates scene consistency, motion correctness, and semantic alignment, showing that visually stable output can miss the task motion.
WorldLens~\cite{liang2025worldlens} and DrivingGen~\cite{zhou2026drivinggen} evaluate driving models, from generation and reconstruction to action following, trajectory plausibility, and downstream tasks.
WorldArena~\cite{shang2026worldarena} places perceptual quality alongside data-engine, policy-evaluation, and planning utility and finds that the two rankings need not agree.

Latent-state models need their own audits.
ARC-Bench~\cite{zhang2026arcbench} tests whether frozen latent-distance objectives rank candidate actions by their executed outcome costs, reporting rank agreement and regret while varying the replanning frequency, since frequent replanning can mask poor scoring.
PokeWorld~\cite{tan2026pokeworld} pushes visually identical objects with different mass, drag, and stiffness to ask whether these parameters can be decoded from the latent state and whether predicted action effects depend on them correctly.

\subsubsection{Foundational and Comprehensive Benchmarks}

Foundational video benchmarks set the visual and temporal quality floor on which control, memory, and utility tests build.
They include the generation suites VBench and VBench-2.0~\cite{huang2023vbenchcomprehensivebenchmarksuite,zheng2025vbench20advancingvideogeneration} (Section~\ref{sec:writing-eval}) and VBench++~\cite{huang2024vbenchpp}, which adds image-conditioned generation and trustworthiness.

Comprehensive world-model suites differ mainly in protocol.
WorldScore~\cite{duan2025worldscore} asks models to generate successive scenes along given camera trajectories and scores controllability, quality, and dynamics.
WorldSimBench~\cite{qin2024worldsimbench} pairs human-aligned perceptual assessment with a manipulation test that turns generated video into control signals.
WorldMark~\cite{xu2026worldmark} and WBench~\cite{ying2026wbench} evaluate interactive generation through standardized action sequences and multi-turn episodes, respectively.
WorldOlympiad~\cite{worldolympiad2026} scores physical faithfulness, geometric consistency, and interaction fidelity across games, robotics, and open-domain video.
Shared tooling helps hold data and controllers fixed across comparisons: the stable-worldmodel platform~\cite{stableworldmodel2026} provides common data interfaces, baselines, planners, and environments with controllable visual, geometric, and physical factors.

\begin{insightbox}{From predictive state to intervention and utility}
Evidence for a world model should match its transition protocol (Table~\ref{tab:evidence-ladder}).
Passive models show predictive structure, and controlled models must also change the world where an action acts and keep what it leaves untouched.
Interactive systems must preserve those consequences under feedback, and decision applications must turn them into better choices.
\end{insightbox}

\surveytheme{surveyNeutral}
\section{Open Challenges and Future Directions}
\label{sec:challenges}
\chapterlead{Eight open problems span reading, writing, sharing, and interacting; each needs new methods and better tests.}

Current models share recurring limits: encoders rarely maintain a predictive state, generators lose identity and physical constraints over time, unified models have yet to show controlled synergy, and interactive systems degrade as generated history accumulates.
These limits, together with gaps in evaluation, action supervision, generalization, uncertainty, and deployment, define eight challenges: long-horizon consistency, physical and causal grounding, measurable synergy, standardized evaluation, control from unlabeled video, generalization across scenes and embodiments, stochastic worlds with rare events, and trust and safety.
\subsection{Long-Horizon World Consistency}
\label{sec:challenge-long-horizon}

\begin{wrapfigure}{r}{0.63\textwidth}
  \centering
  \captionsetup{font=footnotesize,skip=3pt}
  \includegraphics[width=0.99\linewidth]{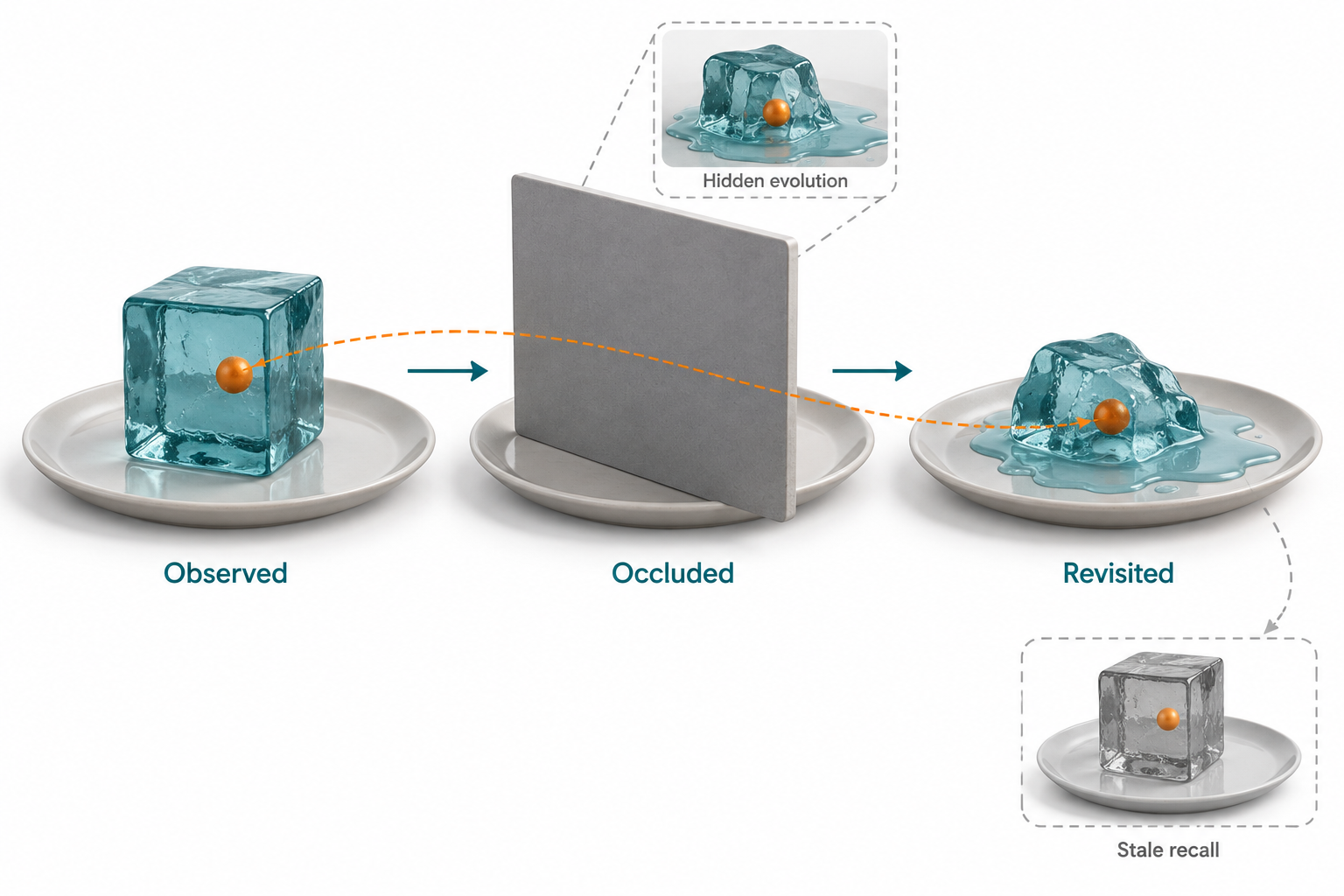}
  \caption{Hidden state evolves during occlusion; replaying an old observation can miss that change.}
  \label{fig:challenge-long-horizon}
\end{wrapfigure}

Long-horizon consistency requires a model to retrieve distant evidence, preserve state that is no longer visible, and render that state consistently in later predictions.
Long-video understanding benchmarks~\cite{wu2024longvideobench,wang2025lvbench,zhou2025mlvu,videoodyssey2026} mainly test the first of these (Section~\ref{sec:reading-tasks}).
Autoregressive generation and interaction add a further difficulty: generated errors enter the history that later predictions depend on.
Current methods balance recent and distant context~\cite{lu2024freelong,zhang2025framepack}, store retrievable observations~\cite{xiao2025worldmem,hong2025relic}, or compress history into recurrent or fixed-size memory~\cite{bian2026echoinfinity,wei2026gimworld} (Section~\ref{sec:wm-memory}).
However, most of them store what was visible, whereas a predictive state must also track what changed off-screen; only a few systems model such change explicitly~\cite{wang2026worlddirector,duan2026liveworld}.
For example, a hidden ice cube keeps its identity as it melts (Figure~\ref{fig:challenge-long-horizon}).
Replaying the stored observation returns a stale cube~\cite{memobench2026}.

One method direction is memory that updates a hidden state during occlusion, trained to predict how unseen entities evolve, instead of replaying stored frames.
Such memory complements rollout training: LIVE~\cite{liveworldmodel2026} targets error growth through forward--reverse cycle consistency without any external memory, so the two address different failures and can be combined.
Memory must also fit an interactive budget: uncapped archives grow with the rollout, and retention budgets~\cite{alradi2026keepsake} turn this growth into eviction decisions, so learning what to keep is itself a research problem.

Memory benchmarks~\cite{ye2026mind,zhang2026mbench,memobench2026,worldroambench2026} now test revisits, occlusion, and off-screen change (Section~\ref{sec:wm-benchmarks}).
The next step is to vary rollout length and revisit distance separately and report degradation curves instead of one terminal score, so that rendering drift and state loss can be told apart.
Latent-state models have no frames to score, so they need probes of goal reachability, feature stability, and decision quality as the horizon grows.
Consistency results should also report memory size, end-to-end latency, and action-update rate.

\Needspace{14\baselineskip}
\subsection{Physical and Causal Grounding}
\label{sec:challenge-physics}

\begin{wrapfigure}{r}{0.60\textwidth}
  \centering
  \captionsetup{font=footnotesize,skip=3pt}
  \includegraphics[width=0.99\linewidth]{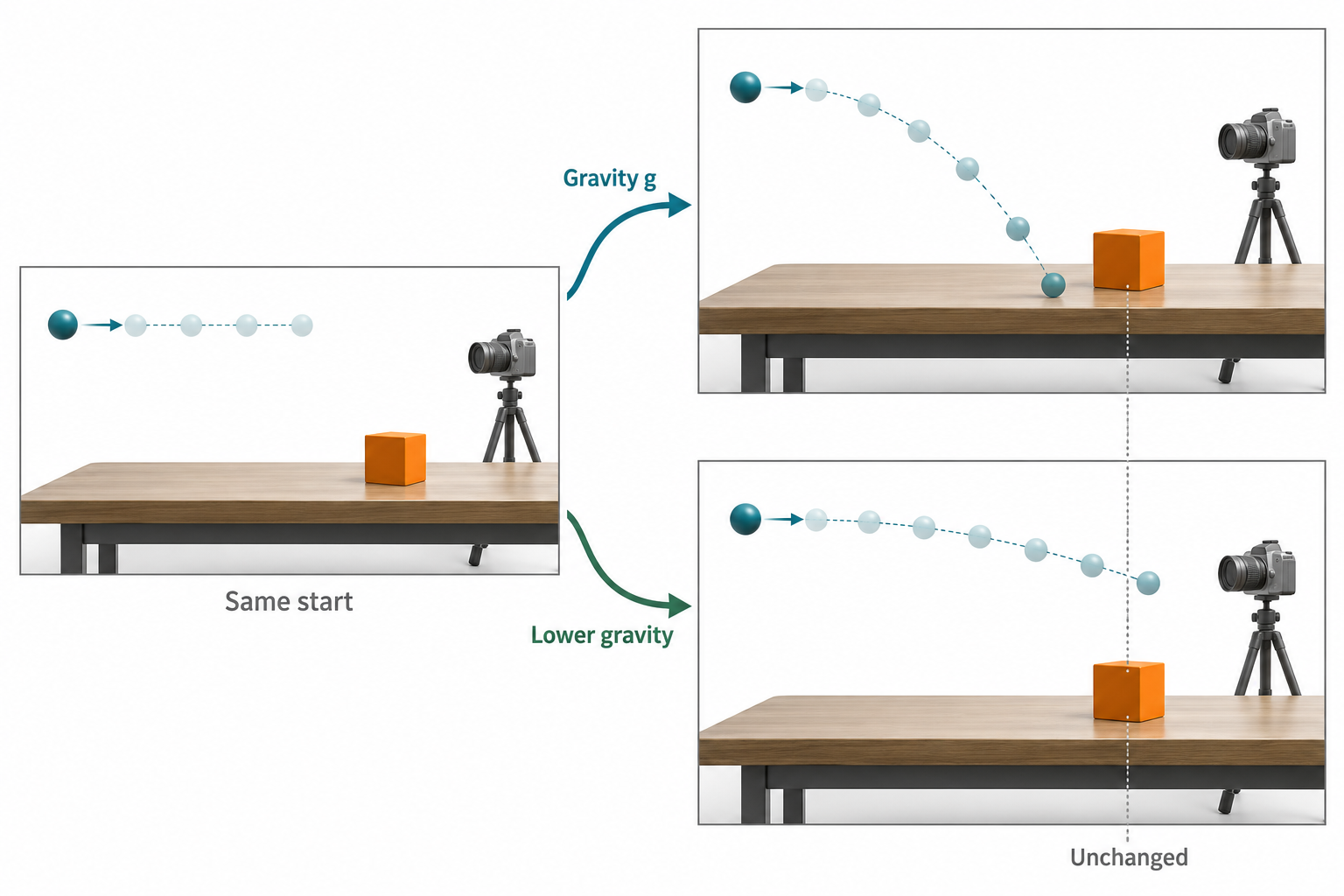}
  \caption{Changing gravity alters the trajectory while unrelated scene properties remain fixed.}
  \label{fig:challenge-physics}
  \vspace{-2mm}
\end{wrapfigure}
Physical grounding can be tested at three increasingly demanding levels.
Reading benchmarks~\cite{intphys2018,physion2021,clevrer2019} test whether a model recognizes persistence, stability, contact, and counterfactual outcomes in observed events.
Generation benchmarks~\cite{bansal2024videophy,bansal2025videophy2,meng2024phygenbench,li2025worldmodelbench,gu2025phyworldbench,yuan2024chronomagicbenchbenchmarkmetamorphicevaluation} ask whether a synthesized outcome follows physical commonsense, a named law, or the direction of an irreversible process (Section~\ref{sec:writing-eval}).
Action- and parameter-conditioned models face a stronger test: changing an action or physical parameter should change the affected outcome and leave unrelated state intact.
For example, a ball released with the same velocity under two gravity settings should follow different trajectories while the camera and background stay fixed (Figure~\ref{fig:challenge-physics}).
A realistic continuation can still respond wrongly to gravity or alter unrelated state, so realism is not enough~\cite{physeditworld2026,li2025worldmodelbench}.

Current methods add physics at different stages (Section~\ref{sec:wm-physics}): they align generator features with a video encoder~\cite{zhang2025videorepa}, simulate dynamics before rendering~\cite{liu2024physgen}, condition generation on a planned motion~\cite{yang2025vlipp}, reward geometric consistency~\cite{worldr1_2026}, or retrieve videos of similar phenomena~\cite{physrag2026}.
Each depends on an outside source: the teacher's coverage, the recovered state and material parameters, the generator's ability to follow the plan, or the quality of the feedback and retrieved evidence.
Paired rollouts that change one physical parameter~\cite{physeditworld2026}, production data with recorded branches~\cite{cgworld2026}, and coupled action-conditioned futures~\cite{he2026oneworld} provide the missing contrast between a change and its effect.
One direction combines these physics priors with intervention-paired data, with objectives that reward the intended change and penalize drift elsewhere.

Evaluation should also diagnose failures: an arrow-of-time error, a contact violation, and a wrong response to an action all make a clip implausible, but they implicate different model components and training signals.
Recent intervention benchmarks~\cite{worldbench2026,cronos2026,whatifworld2026} point toward a common protocol: hold the initial state fixed, change one action or physical parameter, and test the changed outcome, the invariance of unrelated state, the first violated constraint, and when it appears.
Applying the same interventions to reading questions, generated continuations, and closed-loop rollouts would make transfer across operations measurable.

\Needspace{14\baselineskip}
\subsection{Measurable Synergy in Unified Video Models}
\label{sec:challenge-synergy}

Unified video models span discrete-token, continuous, and modular designs~\cite{wang2024emu3,liu2025tuna,univideo2025,vega2026,hydrax2026}, and some route information from generation to understanding, from reasoning to generation, or into planning~\cite{qin2026univigu,pan2026omniweaving,lu2025uniugp} (Sections~\ref{sec:sharing-arch} and~\ref{sec:sharing-synergy}).
The open question is causal: how much of the gain comes from shared representations or information flow, and how much from additional data, parameters, or supervision?
Video evidence of transfer is so far partial or one-directional: video-diffusion features support recognition and geometry~\cite{gen4u2026,genception2026}, an MLLM's reasoning trace raises generation scores~\cite{pan2026omniweaving}, mixing data helps both tasks~\cite{fu2026lance}, a frozen vision-language critic improves editing~\cite{liu2025revise}, and predicted features affect actions~\cite{wang2026ewam} (Section~\ref{sec:sharing-learning}).
Breadth suites~\cite{wei2026univbenchunifiedevaluationvideo,li2026unimunifiedanytoanyinterleaved} catch regressions but not transfer, and image-domain synergy tests~\cite{shi2026realunifyunifiedmodelstruly,yang2026pseudounificationentropyprobingreveals} miss what video adds: transferred information must preserve temporal order, motion, identity, and evolving state.

\vspace{-2mm}
\begin{wrapfigure}{r}{0.57\textwidth}
  \centering
  \captionsetup{font=footnotesize,skip=3pt}
  \includegraphics[width=0.99\linewidth]{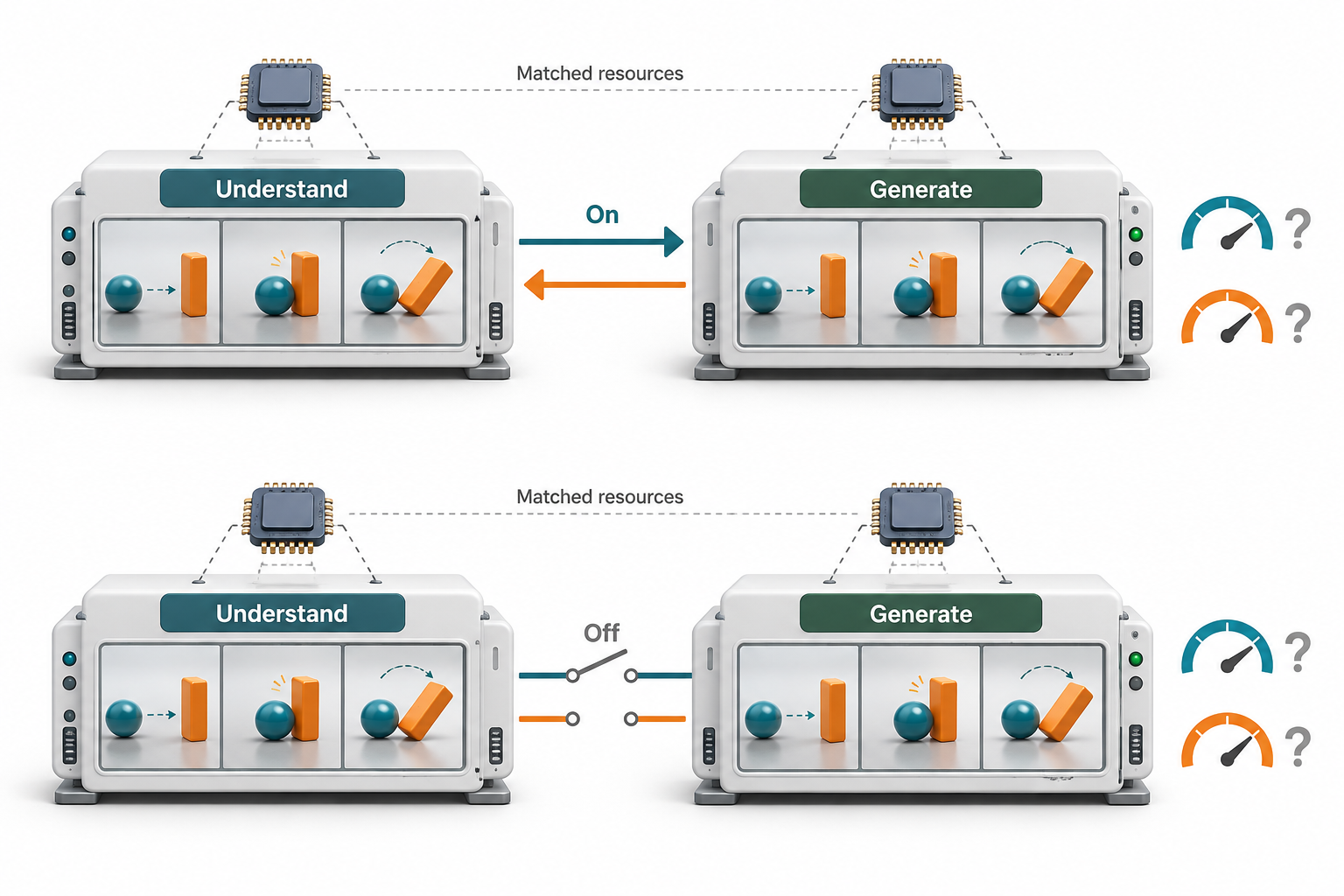}
  \caption{Two systems with matched resources differ only in whether information passes between understanding and generation.}
  \label{fig:challenge-synergy}
  \vspace{-2mm}
\end{wrapfigure}
On the method side, each transfer path can be built as a separate component that can be switched off, such as a temporal-analysis loss on the generator or generated intermediate states passed to a temporal reasoner.
Round-trip consistency between inferred trajectories and rendered video can then serve as a training signal as well as a test.

The test itself compares two systems with the same tasks, data, capacity, and compute that differ only in whether the cross-direction path is enabled (Figure~\ref{fig:challenge-synergy}).
Temporal analysis should then improve event order, motion, identity, or physical dynamics in generation, and generated intermediate states should improve temporal inference or planning under the same observation and parameter budget.
Round trips render an inferred trajectory or event graph and read it back, or analyze, revise, and regenerate a rollout.
Each task must define its invariants, because one description can match several valid videos, and the intermediate representation must not carry the answer verbatim~\cite{niu2025understandinginformgenerationunified}.

\Needspace{14\baselineskip}
\subsection{Standardized Evaluation of World Properties}
\label{sec:challenge-evaluation}

\begin{wrapfigure}{r}{0.50\textwidth}
  \centering
  \captionsetup{font=footnotesize,skip=3pt}
  \includegraphics[width=0.99\linewidth]{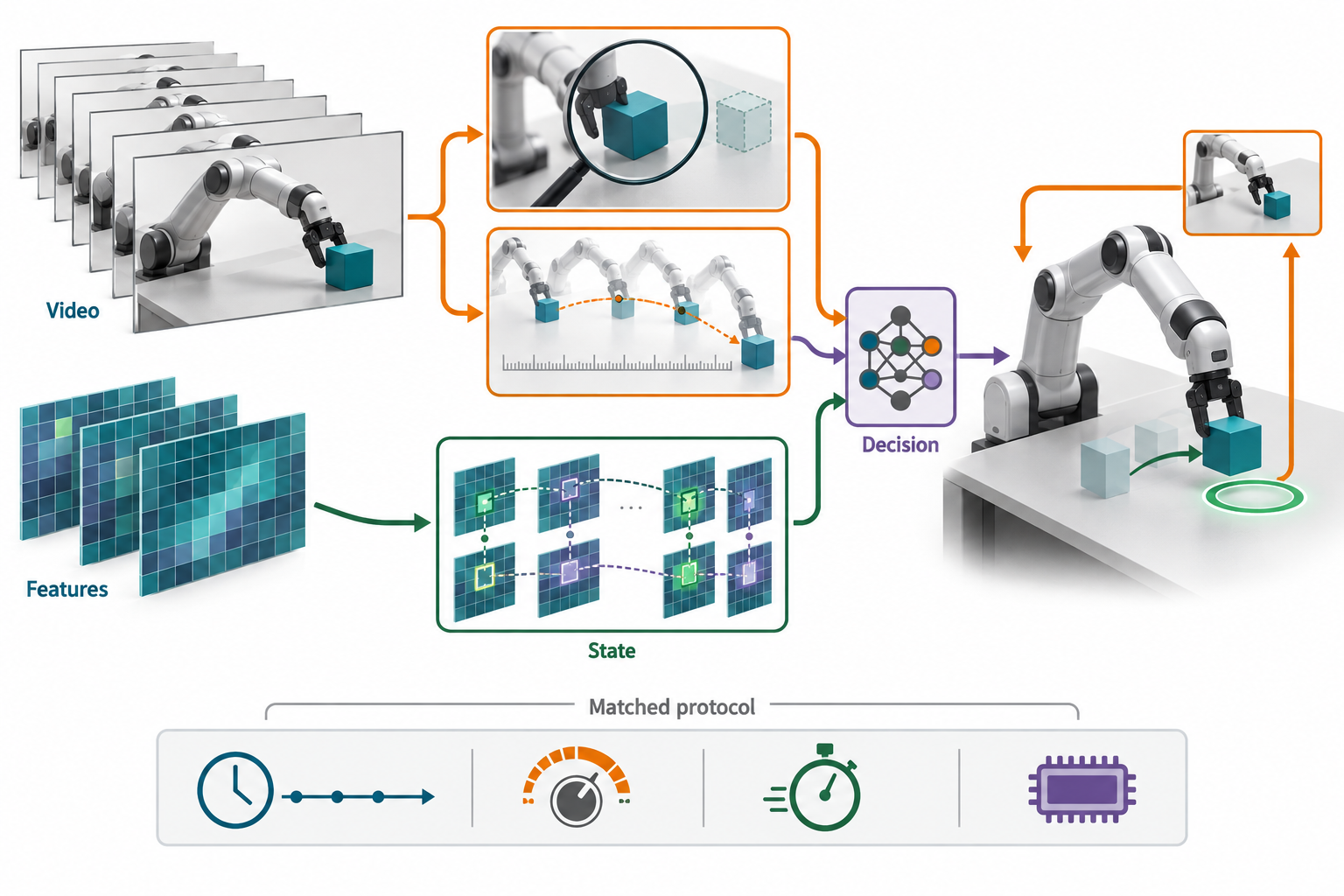}
  \caption{Video and feature predictions need different probes, but both face the same decision test.}
  \label{fig:challenge-evaluation}
  \vspace{-2mm}
\end{wrapfigure}
World-model evaluation now covers control, memory, physical validity, and decision utility as well as visual realism, but the resulting numbers are rarely comparable.
Suites for generated video, action response, and broader world properties~\cite{huang2023vbenchcomprehensivebenchmarksuite,zheng2025vbench20advancingvideogeneration,arai2024actbench,fang2026iworldbench,duan2025worldscore,xu2026worldmark,ying2026wbench,liang2025worldlens,worldroambench2026,worldolympiad2026} differ in prompts, control spaces, horizons, and judges (Section~\ref{sec:wm-benchmarks}), so scores do not transfer across suites.
For example, two predictors guiding the same robot may return video and features, respectively (Figure~\ref{fig:challenge-evaluation}).
Video allows visual and trajectory checks while features need state probes, and both need a decision test, because either can score well on its native metric yet fail at control~\cite{yu2026worldmodeleval,zhang2026arcbench}.

A reporting standard should first match the protocol to the model's prediction space and transition interface (Section~\ref{sec:wm-taxonomy}).
Interactive models need mid-rollout interventions and latency tests, action-conditioned models need faithful responses to fixed controls, and passive predictors need held-out transition tests.
Each result should state the horizon, action frequency, conditioning, and judge model, and claims should rise from visual to controllable-rollout to decision-utility evidence~\cite{yu2026worldmodeleval} (Table~\ref{tab:evidence-ladder}).
Real-time claims need control-to-display latency, sustained throughput, and latency variance, reported with hardware, resolution, denoising steps, and active context, because frames per second alone is ambiguous.

Shared infrastructure~\cite{stableworldmodel2026} standardizes data readers, planning solvers, and environment variations.
Fixed-candidate action ranking should accompany closed-loop success, because frequent replanning can hide a poor scorer~\cite{zhang2026arcbench}.
Integrity needs held-out environments and leakage-controlled synthetic tasks, because web-scale and synthetic training data make overlap hard to detect.
On the method side, two tools need development in their own right: learned judges, which can inherit their source models' blind spots and need validation against human ratings, and standard probes that read latent predictions without a pixel decoder.

\Needspace{14\baselineskip}
\subsection{Scaling Control from Unlabeled Video}
\label{sec:challenge-unlabeled-control}

\begin{wrapfigure}{r}{0.63\textwidth}
  \centering
  \captionsetup{font=footnotesize,skip=3pt}
  \includegraphics[width=0.99\linewidth]{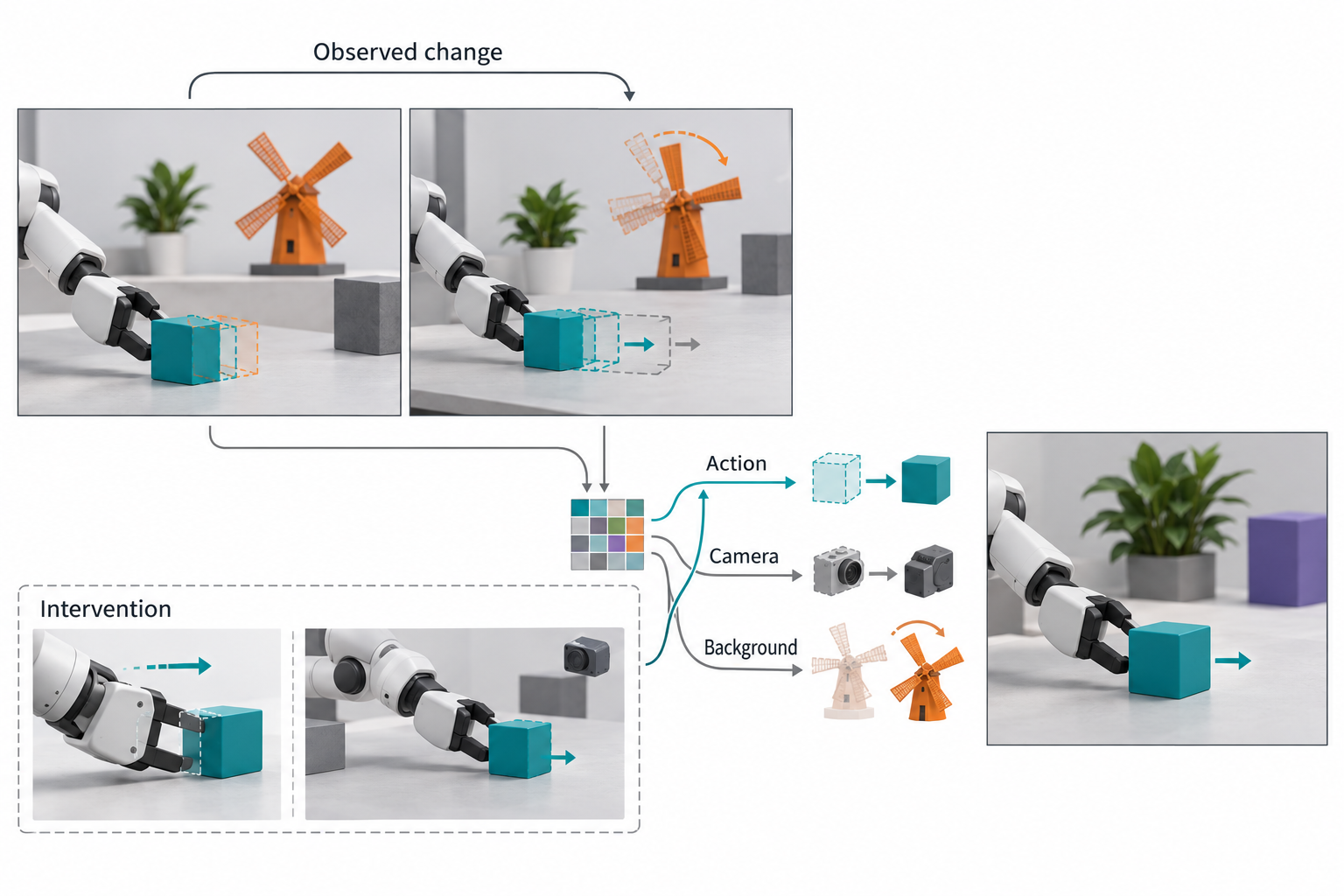}
  \caption{Action effects mix with camera and background motion.}
  \label{fig:challenge-unlabeled-control}
  \vspace{-2mm}
\end{wrapfigure}
Internet video covers objects, agents, and dynamics but usually lacks the action behind each transition.
Latent-action methods~\cite{schmidt2024lapo,bruce2024genie,ye2025lapa,chen2024moto} infer actions as a learned variable (Section~\ref{sec:wm-control}), enabling the use of far more video than labeled datasets, but scale alone does not specify what the variable represents.
For example, a gripper may push a block while the camera pans and a windmill turns (Figure~\ref{fig:challenge-unlabeled-control}).
Only the push is the agent's action, yet a latent code trained for reconstruction can absorb the camera and background motion while reproducing the clip.
Distractor experiments confirm this and show that limited action supervision can substantially improve the recovered factors~\cite{nikulin2025laom}.

Reconstruction accuracy and codebook usage are thus weak evidence of control: a useful latent action should predict controllable effects, stay stable under camera and background changes, and transfer through a simple mapping.
Several weak signals can push latent actions toward these properties.
A small labeled set can anchor latent dimensions to real controls, multi-view invariance can separate camera motion from scene change, temporal bottlenecks can discard details irrelevant to control, and paired interventions can mark which changes belong to the agent.
Continuous latent actions~\cite{gao2025adaworld}, human--robot shared representations~\cite{chen2024igor}, and egomotion controls from tracked pixel motion~\cite{sundar2026actionforcing} move in this direction, while state-annotated data~\cite{wildworld2026,cgworld2026} give stronger but narrower supervision.

Pseudo-action generation offers another source of training pairs: a video model produces a trajectory and an inverse or latent-action model labels its controls.
Such data helps only when the inferred action has the intended effect in a controlled environment or real system; otherwise the generator and the labeller reinforce each other's shortcuts.
Evaluation should keep a labeled subset with interventions and measure transfer under unseen cameras, backgrounds, scenes, and embodiments.

\Needspace{14\baselineskip}
\subsection{Generalization across Scenes and Embodiments}
\label{sec:challenge-generalization}

\vspace{-2mm}
\begin{wrapfigure}{r}{0.48\textwidth}
  \centering
  \captionsetup{font=footnotesize,skip=3pt}
  \includegraphics[width=0.99\linewidth]{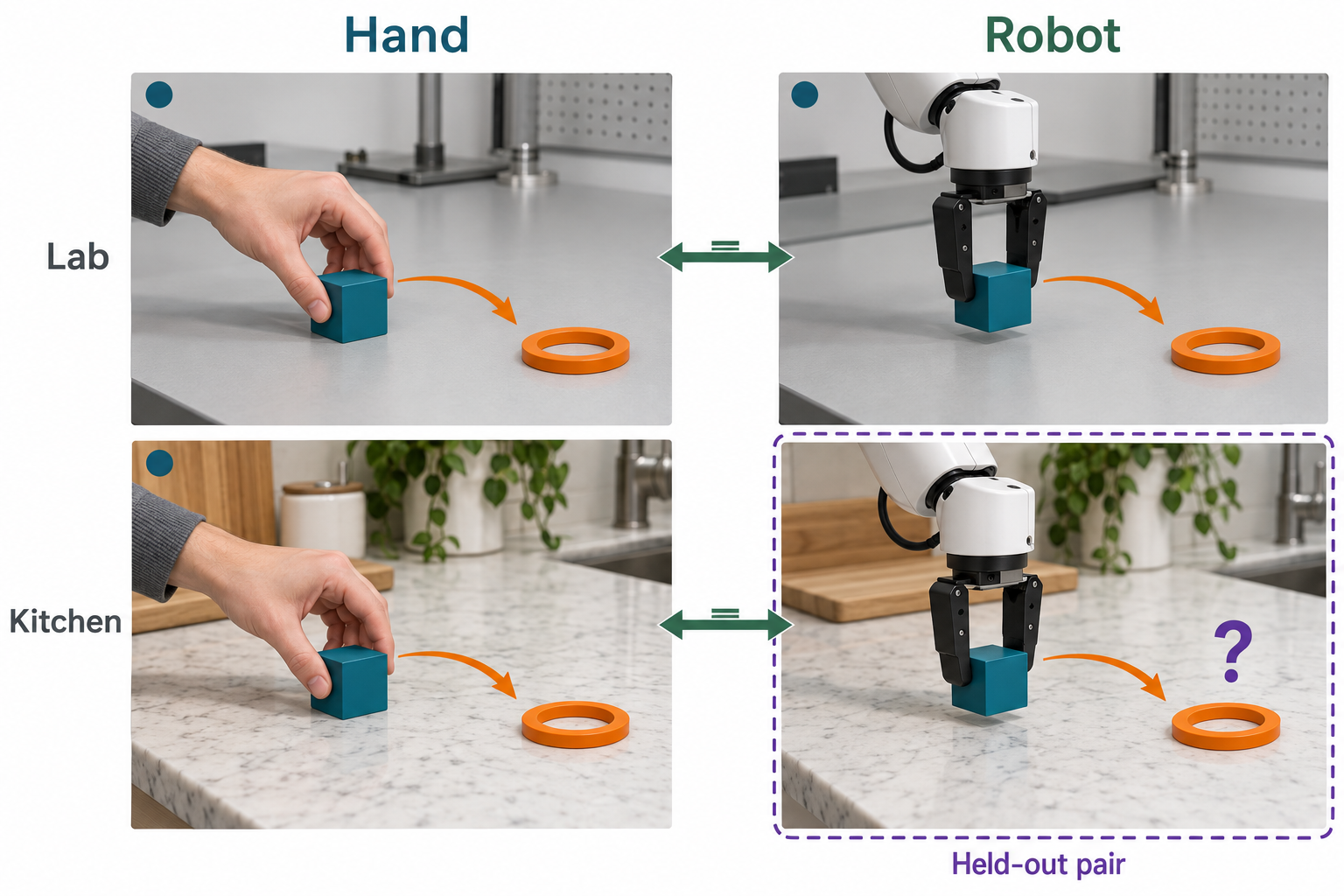}
  \caption{A held-out scene--body pair tests whether familiar factors compose while preserving action effects.}
  \label{fig:challenge-generalization}
  \vspace{-2mm}
\end{wrapfigure}
Scene generalization changes appearance, layout, objects, or local dynamics for the same body, as when a robot enters a new room.
Embodiment generalization changes the body, sensors, action frequency, degrees of freedom, or physical meaning of an action, as when skills move from human hands to a robot arm.

Zero-shot claims therefore depend on which factor is held out.
V-JEPA 2-AC~\cite{assran2025vjepa2} transfers to new scenes without target data, but on a robot type seen in post-training, not an unseen embodiment.
Other methods adapt latent actions with limited interaction~\cite{gao2025adaworld}, generate target-domain experience~\cite{jang2025dreamgen,gigaworld2025}, or broaden the pretraining prior~\cite{gamefactory2025}.
For example, training may cover a human hand in a lab and a kitchen but a robot only in the lab (Figure~\ref{fig:challenge-generalization}).
Deploying the robot in the kitchen tests an unseen scene--body pair, where the representation and control mapping must still move the block to the target.

Cross-embodiment transfer adds an alignment problem: the same semantic action may need different joints and timing on different bodies, while similar-looking motion can have different contact consequences.
Current methods connect human and robot behavior through shared image-goal units~\cite{chen2024igor} or map latent actions to robot commands with a small labeled set~\cite{ye2025lapa}.
A method direction is to train such mappings to preserve effects, with objectives defined on outcomes such as object displacement, contact, and task completion instead of raw control values or token similarity.
Evaluation should cross scenes, embodiments, sensors, and action spaces and test effect equivalence, controllability, and safety after transfer.
Reports should state whether results are zero-shot, calibrated, or fully fine-tuned, and how much target data they used.

\Needspace{14\baselineskip}
\subsection{Stochastic Worlds and Rare Events}
\label{sec:challenge-stochastic}

\begin{wrapfigure}{r}{0.63\textwidth}
  \centering
  \captionsetup{font=footnotesize,skip=3pt}
  \includegraphics[width=0.99\linewidth]{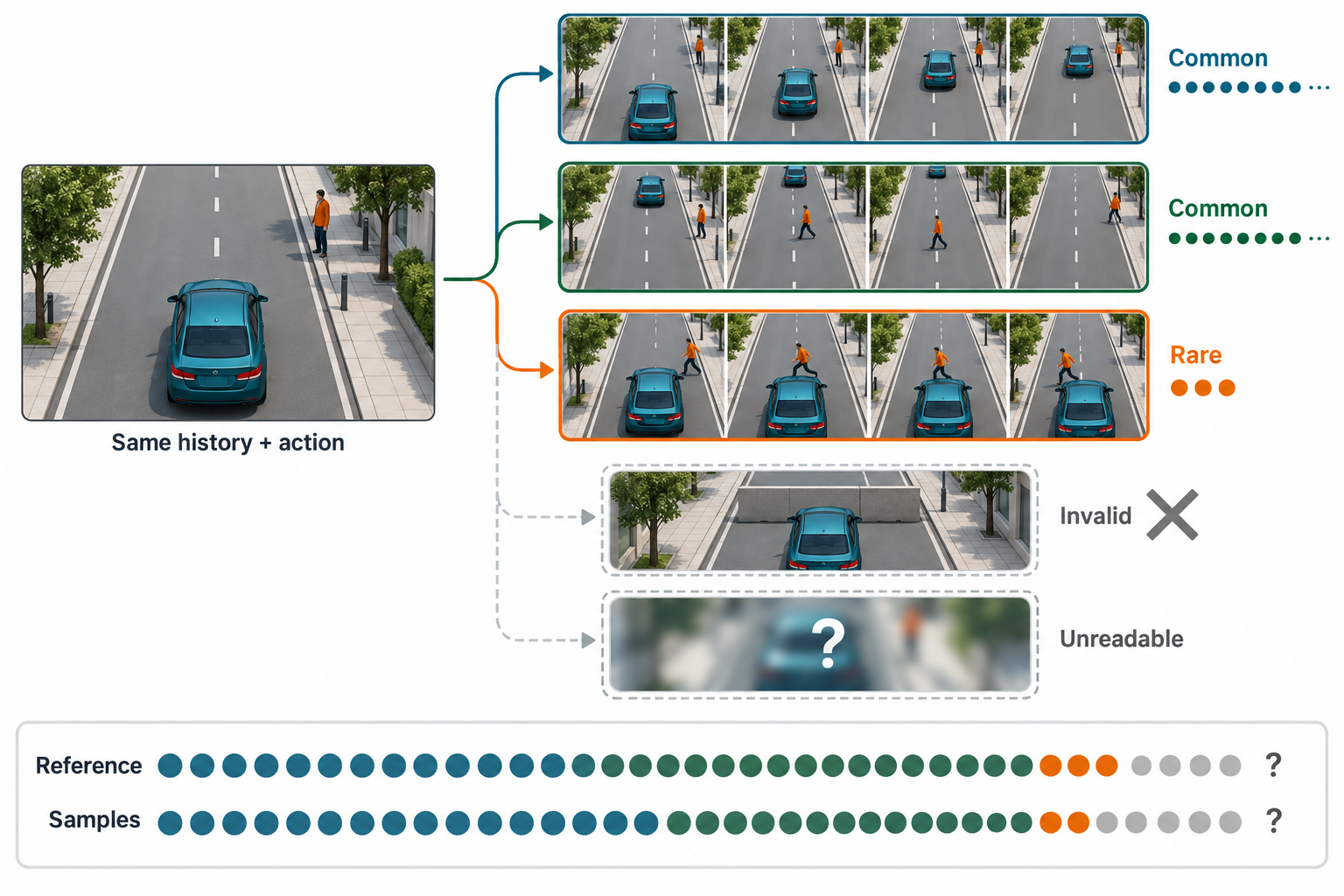}
  \caption{Valid futures differ in frequency, and invalid and unreadable samples must be counted separately.}
  \label{fig:challenge-stochastic}
  \vspace{-2mm}
\end{wrapfigure}

One history can admit several valid futures, because other agents may act differently and relevant physical state may be hidden.
A deterministic predictor cannot express these alternatives, and squared-error training can favor an average of incompatible outcomes.
Latent-variable predictors~\cite{babaeizadeh2018sv2p,lee2018savp} sample multiple futures, and controllable generators and data engines~\cite{russell2025gaia2,gao2024vista,nvidia2025cosmos,jang2025dreamgen,gigaworld2025} can synthesize rare scenarios for training (Section~\ref{sec:interacting}).
However, synthesis does not say how likely each outcome is: a prompted collision supplies a training case, whereas a calibrated collision risk supports a decision under uncertainty.
For example, given a fixed driving history and ego actions, a pedestrian may wait, cross after the car passes, or step into its path (Figure~\ref{fig:challenge-stochastic}).
A generator can render the rare crossing without assigning it a reliable probability.

One direction is to develop generators that assign calibrated outcome probabilities rather than merely produce diverse samples, using empirical event rates or justified simulator priors.
Generated rare cases must be checked for physical feasibility, causal relevance, and downstream utility before entering training data, since prompting fixes which failures appear but not their frequencies.

For a fixed history and action sequence, distributional evaluation assesses whether samples cover distinct valid outcomes, exclude invalid ones, and match observed or simulated outcome frequencies.
Diversity alone rewards arbitrary variation, and best-of-many scoring hides misplaced probability mass.
Reports should therefore use proper scoring rules, calibration curves, and mode coverage at a fixed sample budget, distinguish unreadable from invalid samples, and use sample sizes appropriate to event rarity.
Repeated-sampling benchmarks~\cite{pu2026pawbench,sadeghi2026calibench} and confidence calibration~\cite{mei2025c3} are a start (Section~\ref{sec:wm-benchmarks}), but extending outcome-frequency tests from endpoints in limited scenarios to long, action-conditioned trajectories is open.

\Needspace{14\baselineskip}
\subsection{Trust and Safety of Video World Models}
\label{sec:challenge-safety}

Beyond modeling the world faithfully, video world models must also be considered in terms of how their outputs shape real-world beliefs and decisions.

\begin{wrapfigure}{r}{0.63\textwidth}
  \centering
  \captionsetup{font=footnotesize,skip=3pt}
  \includegraphics[width=0.99\linewidth]{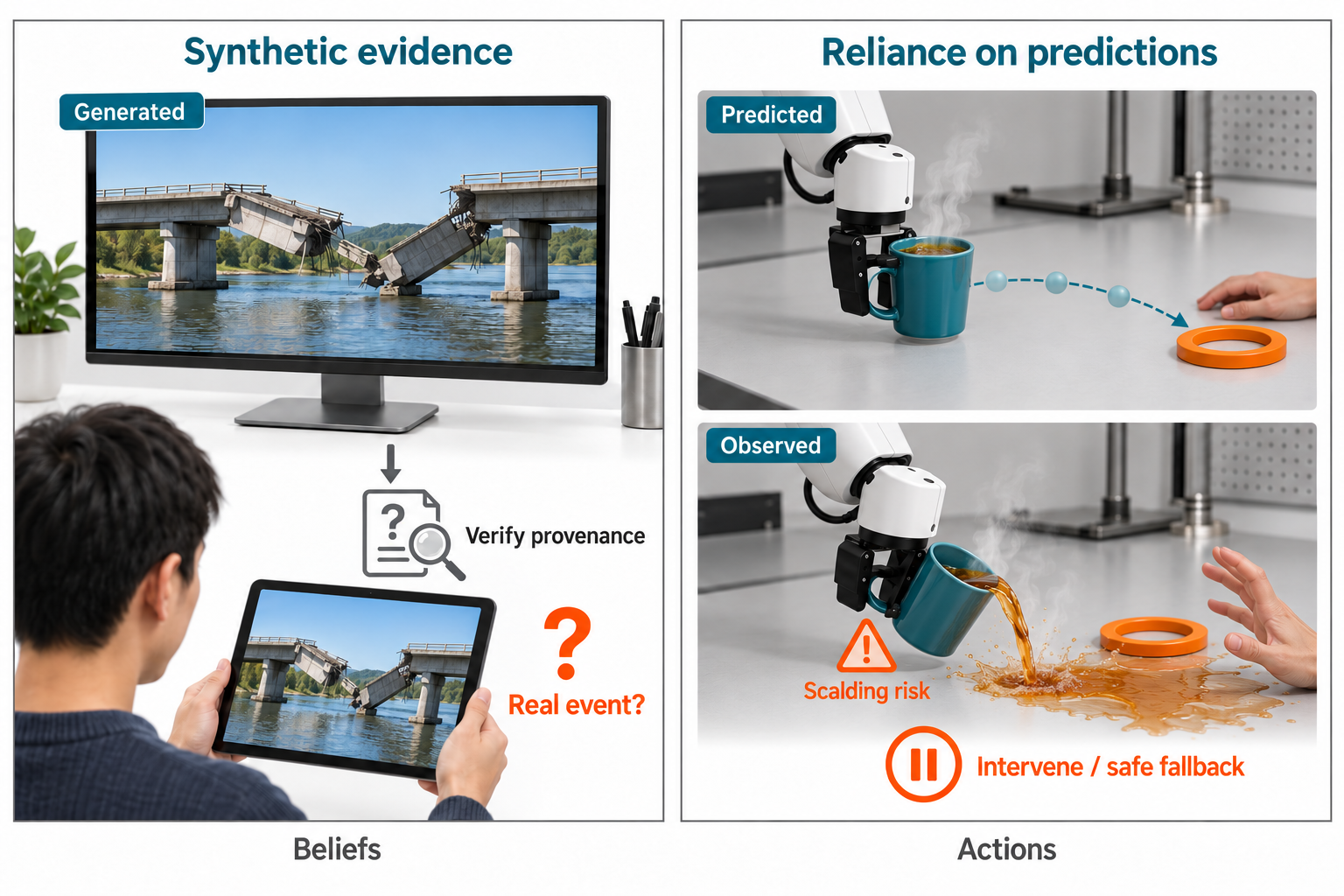}
  \caption{Synthetic video can shape beliefs, while predicted rollouts guide actions; trustworthy use requires provenance checks and intervention when predictions conflict with observations.}
  \label{fig:challenge-safety}
  \vspace{-2mm}
\end{wrapfigure}
One concern is the misuse of synthetic video. 
The world consistency that Sections~\ref{sec:writing} and~\ref{sec:interacting} pursue also removes the cues that reveal generated footage, and interactive generation makes fabrication real-time and conditional on a viewer's actions.
Three defenses exist, each with an open problem.
Passive detectors trained on one generator's artifacts transfer poorly to the next: on GenVideo, a million-scale benchmark that spans many generators, detectors lose much of their accuracy on held-out generators~\cite{chen2024demamba}.
Active watermarks embed a signal at generation time, in the latent decoder~\cite{fernandez2023stablesignature} or in the initial noise~\cite{wen2023treering} for images, and with spatial and temporal redundancy for video~\cite{fernandez2024videoseal,hu2025videoshield}; they must survive compression, re-encoding, and cropping, and, for world models, continuation or editing by another model that resamples the frames.
Provenance credentials sign content at capture or generation time~\cite{c2pa2024spec}, but they cover only signed pipelines and leave open-weight models unconstrained.
As with the protocols of Section~\ref{sec:challenge-evaluation}, detection and watermarking results should report robustness to unseen generators and to transformations rather than accuracy on one fixed set.

The second concern is whether capable video world models can be safely relied upon for consequential real-world decisions. 
Sections~\ref{sec:wm-imagination} and~\ref{sec:wm-policy-eval} illustrate how generated rollouts are used to train and evaluate policies, yet policies optimized within learned environments can exploit model errors~\cite{kurutach2018metrpo,janner2019mbpo}, particularly as their behavior shifts beyond the model's training distribution~\cite{liu2026worldvlaloop}. 
Improving simulation fidelity alone, however, does not guarantee safe decisions. 
Even accurate models may fail in rare, high-consequence situations (Section~\ref{sec:challenge-stochastic}), while predicting how the world evolves does not determine which actions are appropriate under real-world objectives, constraints, and effects on other agents. 
Moreover, reliability depends not only on the model but also on how its predictions are used. 
Uncertainty estimation and calibration~\cite{chua2018pets,mei2025c3} are useful only when they inform decisions to act, verify, or defer, while agreement between simulated and real-world policy outcomes~\cite{quevedo2025worldgym,li2025worldeval} provides assurance only under the evaluated conditions. 
The central challenge is therefore to establish justified reliance rather than simulation accuracy alone: defining operational boundaries, monitoring discrepancies between predicted and observed outcomes, and enabling intervention or fallback when those boundaries are exceeded. 

\begin{insightbox}{Scaling world models needs interventions, not just more clips}
Passive video expands coverage but leaves hidden state, physical causes, cross-direction transfer, action semantics, and outcome probabilities underdetermined.
Most of these challenges need data and protocols with interventions (known hidden changes, paired physical edits, matched ablations, labeled actions, held-out scene-body pairs, and observed event rates), and trusting a model's outputs needs interventions on the model itself: unseen generators, transformations, and adversarial policies that probe where it fails.
Observational scale shows what co-occurs; interventions show what changes what, and where a model breaks.
\end{insightbox}

\surveytheme{surveyNeutral}
\section{Conclusion}

This survey has treated video as an observed trace of a latent temporal world and used \readop{reading}, \writeop{writing}, \shareop{sharing}, and \interactop{interacting} to organize how foundation models connect observations to represented state.
Comprehension recovers what a trace reveals; generation constructs traces from conditions; unification shares information and machinery between the two; and world modeling asks how represented state changes under conditions or actions.
Within this framing, a video world model is defined by the transition it learns, which predicts future video or video-derived features and runs passively, under a fixed action sequence, or interactively (Section~\ref{sec:wm-taxonomy}).
The distinction between temporal coherence and world consistency (Definition~\ref{def:consistency}) separates the quality of rendered video from evidence that a model tracks the world state behind it.
The next phase demands stronger evidence (Table~\ref{tab:evidence-ladder}).
Beyond perceptual fidelity and local coherence, claims about the world require state that survives occlusion, revisits, and long horizons; controllable action effects; valid responses to interventions; and, for closed-loop use, latency within the control budget.
Stochastic futures also require coverage and calibration, unified models require matched evidence that information transfers between reading and writing, and systems built for training, planning, or evaluation must improve decisions in controlled environments or real tasks.
Meeting these requirements would move the field from rendering convincing video to modeling how the world changes.

\clearpage
\printbibliography

\end{document}